# Births are difficult to predict even with rich survey and full-population register data

**Authors:** Elizaveta Sivak[1,2*], Emily M. Cantrell[3], Thomas Emery[4], Javier Garcia-Bernardo[5], Flavio Hafner[6], Kasia Karpinska[4], Malte Lüken[6], Adrienne Mendrik[7], Joris Mulder[8], Hanzhang Ren[9], Varun Satish[3,10,11], Mark Verhagen[12], Angelica M. Maineri[4], Paulina Pankowska[13], Jasmin Abdel Ghany[12], Bruno Arpino[14,15], Giovanni Cassani[16], Julia Hellstrand[17], Katya Ivanova[18], Sanni Kuikka[19], Ana Macanovic[20], Charles Rahal[21], Felix C. Tropf[22], Roland J. Veen[23], Nicole Walasek[24], Daniël van Wijk[25], Kelsey Q. Wright[17], Emilio Zagheni[26], Henry Abbink[27], Emanuele Aliverti[14], Matteo Amestoy[28], Tilbe Atav[29], Nicola Barban[30], Sunnee Billingsley[31], Goan J. Booij[32], Louis Boucherie[33], Yael Broos[34], Li Ya Chang[35], Jamie C. Chiu[36], Chiara Ludovica Comolli[30], Boris Cule[37], Qixiang Fang[5], Dennis M. Feehan[38], Rachel Ganly[39], Erwin Gielens[18], Rolando M. Gonzales Martinez[40], Andrea Gradassi[41], Rosember Guerra-Urzola[42], Mario Guerra-Urzola[43], Stéphane Guerrier[44], Enamul Hassan[4,45], Vincent A. Haverhoek[46], Andrew T. Hendrickson[37], Amber Howard[47], Yuxuan Jin[48], Sayash Kapoor[49], Erik-Jan van Kesteren[5], Iris ten Klooster[50], Marie Labussiere[51], Lydia Liu[49], Tiffany T. Liu[3,10], Adam Maghout[5], Simone Meneghello[52], Lasse Mohr[33], Clara H. Mulder[40], Saul J. Newman[53], Jessica Nisén[54], Janis Norden[23], Mikkel Odgaard[55], Riccardo Omenti[30], Ozancan Ozdemir[23], Christina Pao[10,49], Paige Park[38], Gaia Penta[52], Juan C. Perdomo[56], Tanzir Pial[45], Alessio Piraccini[52], Federica Querin[30], Ziwei Rao[57], Christian Rellama[43], Adrien Remund[40], Frederieke Richert[23], Arnout van de Rijt[20], Mojtaba Rostami Kandroodi[37], Stijn J. Rotman[37], Lucas Sage[58], Germans Savcisens[59], Katrin Schwanitz[54,60], Steven Skiena[45], Alessandro Spata[61], Yannick Stadtfeld[62], Benedikt Stroebl[49], Gaetano Tedesco[14], Mathilde Theelen[63], Gianluca Tori[52], Abigail Tun-Mendicuti[40], Rishabh Tyagi[64], Keyon Vafa[65], Luiz Felipe Vecchietti[66], Linda Vecgaile[26], Willem R. J. Vermeulen[67], Maria-Pia Victoria Feser[30], Lionel A. Voirol[68], Thom B. Volker[5], Xinran Wang[69], Jiani Yan[70], Xinyi Zhao[39], Flora Zhou[4], Zuzana Zilincikova[40], Malvina Nissim[71], Matthew J. Salganik[3], Gert Stulp[1,2]

## Abstract

Major life events have proven difficult to predict. Does this reflect limits of theory, data, and algorithms, or the large role of chance? We examine one outcome – having a child within three years – through a near-ideal setting for prediction: a data challenge where 147 researchers predicted births for Dutch residents aged 18–45, using survey data and full-population registers. Methods ranged from logistic regression to a large language model and transformers. Predictions were moderately accurate (best F1: register 0.59, survey 0.76); advanced models did not outperform classical ones; and the larger registers did not beat the survey. Simulating the stochastic biology of conception and pregnancy, we estimated a predictive ceiling (survey F1 ≈ 0.86–0.94, register 0.88–0.96). Observed performance falls short of this ceiling, implicating imperfect data, methods, and unmodelled chance, while the ceiling itself shows that chance in reproduction alone sets a non-trivial limit on predicting individual lives.

## Introduction

Predicting life outcomes is increasingly recognised as both important and challenging. Accurate predictions can inform theory[1–4] and institutional decision-making[5–9]. Yet, while fields such as computer science[10,11], biology[12,13], or climate science[14,15] celebrate prominent cases of predictive success, social scientists emphasise the unpredictability of life events[16–21]. The difficulty in predicting life events may result from incomplete theories, imperfect measurement, limitations of the algorithms, or reflect the large role of chance in individuals' life-courses[22,23]. Another potential reason is that the conditions necessary for accurate predictions are not met, specifically, having large high-quality data[23–25].

We study the predictability of one life outcome – having a child within three years – under some of the most favourable conditions for accurate prediction currently available in the social sciences through 'Predicting Fertility in the Netherlands' (PreFer)[26], a large international and interdisciplinary data challenge[27]. Over 140 researchers from demography, sociology, data science, computer science, and other fields participated, applying traditional and state-of-the-art methods to predict this outcome. The diversity of approaches and competitive format make the results likely to reflect the best possible predictions for the given datasets, timeframe, and computational resources[19]. PreFer used full-population administrative records and high-quality nationally representative survey data from the Netherlands, covering many known predictors of fertility[28] as well as likely unknown ones. Together, these conditions offer a rigorous test of the current predictability of fertility behaviour.

Fertility outcomes – whether people have children, when, and how many – are a particularly good test case for understanding the predictability of life outcomes. First, they are among the most universal and important life outcomes, shaping individual lives[29–36] and societies[37,38]. Second, the predictability of individual-level fertility outcomes remains unexplored, as research has typically focused on identifying factors related to fertility[28,39,40] or on forecasting population-level fertility trends by extrapolating from past fertility patterns[41]. Third, births can be measured with low measurement error, are tracked in many datasets, and are relatively easily defined as a prediction target. Finally, reproductive physiology research and estimations of unplanned births enable us to quantify via simulation some of the variation in fertility outcomes that likely results from chance, thus setting an upper bound on predictive accuracy. By comparing the best predictive performance to this upper bound, we can assess how much room for improvement remains. All these reasons make fertility outcomes an ideal 'model organism'[42].

Administrative and social survey data offer distinct strengths and weaknesses in predicting individual-level fertility. Administrative data typically provides vastly larger sample sizes than surveys, enabling models to learn more effectively[22]. It also often has higher resolution and longer longitudinal coverage. Yet administrative records might lack key

information – particularly self-reported behaviours and subjective measures such as fertility intentions, parental support, or perceived gender equality in domestic duties, which have a prominent role in theories of fertility[28,43] – or this information may be harder to extract. Surveys typically capture such self-reported behaviours and subjective measures. However, surveys are typically constrained by relatively small sample sizes, substantial and selective non-response, and attrition, all of which can undermine predictive accuracy. Using both administrative and survey data allows us to test their relative advantages and assess the current predictability of fertility.

In PreFer, administrative data comes from Dutch registers – one of the world's largest and most detailed social datasets[44]. This data provides fine-grained longitudinal information, including socioeconomic, demographic, family, and residential characteristics, and social networks for the whole Dutch population[45,46]. The survey data comes from the LISS panel[47], a longitudinal nationally representative survey of Dutch households, covering similar topics in addition to health, values, personality, fertility intentions and other subjective factors like relationship quality. A key advantage of the LISS data is that it can be linked to register data in a secure environment. This allows us to combine these data sources to potentially enhance predictive performance.

The task of PreFer is to predict whether individuals had a new child between 2021 and 2023 using data up to and including 2020 for Dutch residents aged 18–45 in 2020. For both the survey and register datasets, we split households into training (70%) and holdout sets. The survey sample used in PreFer included 1,382 individuals with known fertility outcomes. The register-based data covered the full target population of about 6 million people (see Methods for details).

In the first phase of the data challenge, over 140 participants predicted fertility outcomes using the survey data. We received 69 valid submissions from 41 teams (details about participation in SI, section 1.2). In a second phase, a selected group of participants was given access to the highly restricted data from the Dutch registers. Facilitated by the Dutch national supercomputer, teams employed diverse approaches, including methods designed to leverage large, high-resolution datasets (overall 11 submissions from 12 teams; see details about computing environment in SI, section 1.3, and submitted methods in SI, section 3). These include a transformer model trained on individual life-courses represented in symbolic language[48], and a large language model fine-tuned on life-course data transformed into natural language[49,50].

To estimate the predictive upper bound, we assume three sources of randomness in human reproduction that are quantifiable from existing empirical research. First, for those trying to conceive, whether conception occurs in any given month is stochastic: individuals with identical characteristics will differ in how quickly they conceive, leading to different outcomes within three years. Additionally, some people might have a higher likelihood of conception (fecundability) than others. We treat variation in fecundability within age groups and age at sterility as random, which adds further random variation in the outcomes. Second, we treat fetal survival – whether a conception results in a live birth – as a chance event. Third, for those trying to avoid pregnancy, we treat contraceptive failure as random. To translate these sources of randomness into an upper bound, we simulate fertility outcomes for a synthetic population in which age, partnership status, and fertility intentions are the only determinants: each agent's outcome is determined by these factors and chance alone. We then fit and evaluate a model on these simulated outcomes using only those three variables. Its predictive error reflects only chance variation, making the resulting F1 score a ceiling on predictive performance under these assumptions (see Methods for details).

## Results

### Predictability is moderate even with very rich and large data

The best register-based model achieved an F1 score of 0.59, 95% CI [0.589, 0.593], and the best survey-based model an F1 score of 0.76, 95% CI [0.68, 0.83] (see Figures 1-2; see SI, section 3 for other metrics; CI calculated via bootstrapping of the holdout). The winning models in both parts of the challenge were established gradient-boosted decision-tree ensembles. Notably, two approaches designed to exploit the longitudinal structure of the register data – a transformer trained on life-course sequences and a fine-tuned large language model – did not outperform the tree-based ensemble (both F1 = 0.57 vs 0.59; see SI section 3 for descriptions of all approaches). Most submitted models outperformed a simple demographic baseline – a logistic regression including only a small set of variables linked to fertility: age (quadratic), gender, parity and age of the youngest child, partnership status, and the highest level of education. For both data sources, the best model's F1 score was three times as high as the demographic baseline (register-based: F1 = 0.19; survey-based: F1 = 0.25; see details and other metrics in SI, section 2.3).

However, in absolute terms, performance is moderate, particularly for identifying people who had a child within three years, which occurred in 15% of the register sample and 22% of the survey sample. The best register-based model correctly identified only 64% of people who had a child (and 91% of those who did not); only 55% of those predicted to have a child actually did so. The best survey-based model was more precise (91% of individuals who were predicted to have a child actually had a child), but achieved only slightly higher recall, correctly identifying 66% of people who had a child and 98% of those who did not.

Moreover, for the survey-based models, performance gains over a simple fertility intentions baseline – a logistic regression based on respondents' stated intentions to have children – were limited. This baseline relies on a single variable capturing whether individuals intend to have (more) children and the expected timing (see details in SI, section 2.3). Only 38% of models exceeded this baseline. The best model's F1 score (0.76) was only 19% higher than the fertility intentions baseline (F1 = 0.64). Fertility intentions information appears critical for the survey-based models: when we retrained two models (the best-performing model and one of the top data-driven models likely to discover alternative predictors) without fertility intentions, performance dropped substantially (see SI, section 2.4), suggesting that other variables cannot substitute for this information (at least for the given sample size).

| Task | Data | Participants | Models | Top result |
|---|---|---|---|---|
| predict who had child in '21-'23 | survey data '07-'20 N = 987 (train) | **147** participants in **41** teams submitted **69** models | logistic regression tree-based ensembles neural nets | F1 holdout (N = 395): 0.76 |
| Pre Fer 1\|0 Predicting Fertility data challenge | register data '95-'20 N = 4 million (train) | **45** participants in **12** teams submitted **11** models | logistic regression tree-based fine-tuned life2vec fine-tuned LLM | F1 holdout (N = 1 mil): 0.59 |

Figure 1. Overview of PreFer design and main findings.

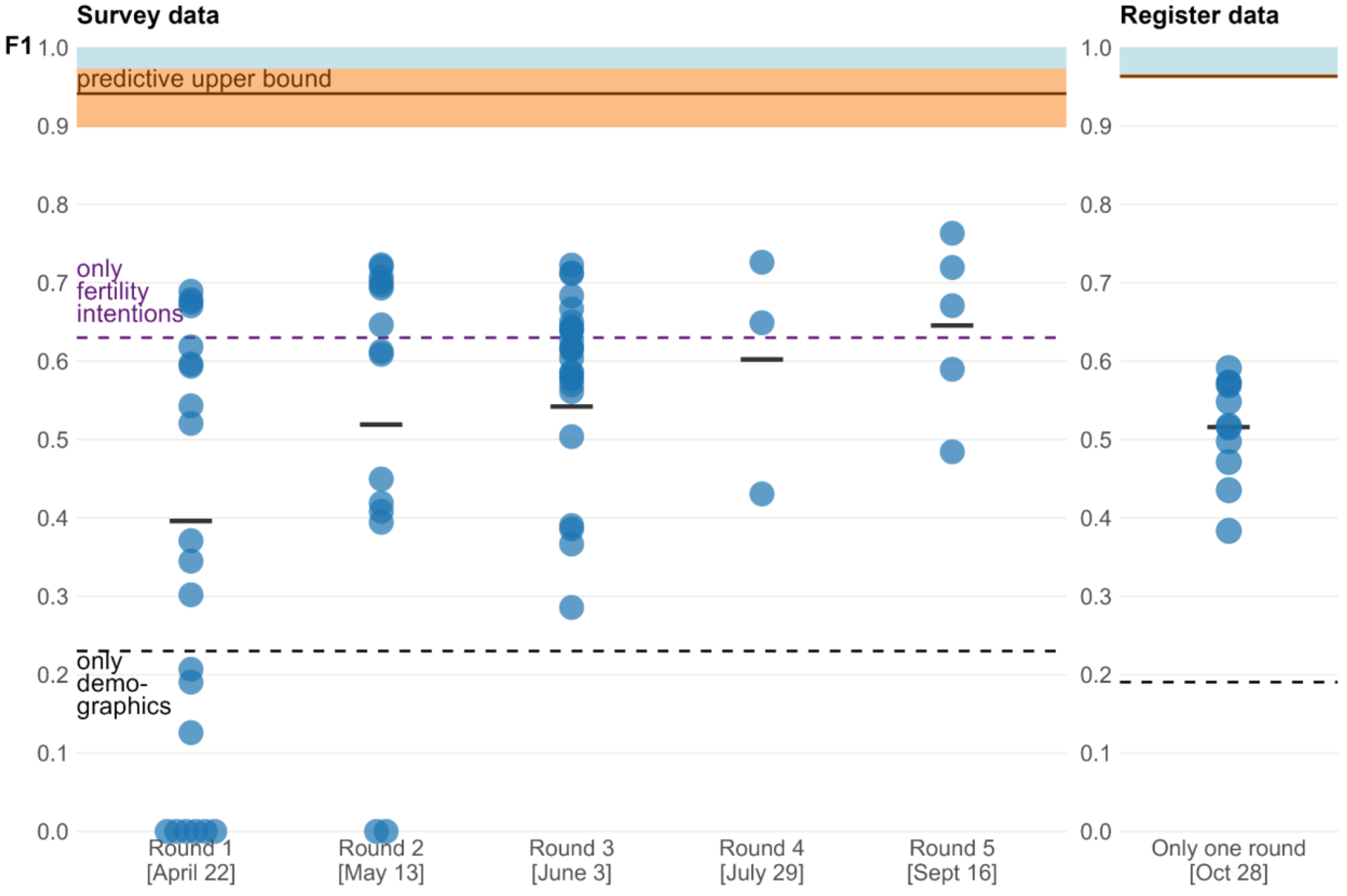

Figure 2. F1 scores of all PreFer submissions (see the list of submissions in SI, section 3). Each dot represents a submission; the short black line represents the mean F1 score in each round. Participants could submit a model up to 5 times in the survey-based part of the challenge and once in the register-based part (see Materials and Methods). Horizontal dashes are baselines (see explanation in Results). The brown line is a predictive upper bound; the orange area represents a 95% bootstrap confidence interval for this bound; the blue area is irreducible

predictive error (under specified assumptions) (see Materials and Methods). The predictive upper bound shown in this plot assumes a 1% contraceptive failure rate.

### Predictions from full-population register-based data do not outperform those from survey data

Notably, the best register-based model did not achieve a higher F1 score than the best survey-based model (0.59 vs 0.76), despite a much larger training sample size and longer longitudinal coverage. However, these scores are not directly comparable: the prevalence of people who had a child within three years is higher in the survey data (22%) than in the register data (15%) (see SI Section 2.5 for details), and F1 is sensitive to outcome prevalence. We addressed this in two ways: first, by retrieving missing outcomes from the register data for survey training cases, reducing prevalence in the training data to 15%, which resulted in F1 = 0.68, 95% CI [0.59, 0.76]; second, by additionally removing positive-outcome observations from the survey holdout to match register prevalence (F1 = 0.63, 95% CI [0.56, 0.69]; see SI Section 2.5 for details). Both adjustments lowered the survey-based score, but in neither case did it fall below that of the best register-based model (F1 = 0.59). The register-based model also depends on its scale to reach this level: trained on a smaller sample size to match the survey data, the F1 score of the top register-based model drops to 0.50 (see SI Section 2.9).

A more direct comparison between the best survey-based and register-based models is possible thanks to the linkage between the survey and register data, allowing us to evaluate both models on the same group – the survey holdout. The register-based model did not predict better on this shared holdout: its predictions had higher mean squared error than predictions of the best survey-based model (delta MSE = 0.03, 95% CI [0.01, 0.05], see details in SI, section 2.5).

### Predictive gap may in part reflect randomness in conception and pregnancy progression

In part, the moderate performance of even the best models can reflect randomness in fertility outcomes. We estimated a predictive upper bound to quantify how much of the error is irreducible given several assumed sources of randomness in conception and pregnancy progression. The magnitude of this bound depends on assumptions about births resulting from contraceptive failure (see Methods). In the Netherlands, the share of unplanned births is estimated at 17%[51–53], but not necessarily all of them result from a random contraceptive failure. Under a conservative estimate that only 1% of births result from contraceptive failure, the predictive upper bound reaches an F1 score of 0.963 (95% CI [0.961-0.966]) for the register data and 0.94 (95% CI [0.90-0.97]) for the survey data. Under a less conservative estimate (17%, i.e., all unplanned births are due to contraceptive failure), the upper bounds decrease to 0.881 (95% CI [0.877-0.886]) for register data and 0.86 (95% CI [0.81-0.92]) for survey data (see SI, section 2.6 for values across the full range).

Across the full range of assumptions, these assumed sources of randomness impose a non-trivial constraint on predictability. Yet, these sources of randomness alone cannot account for the gap between the best models and perfect prediction: a large share of that gap remains that reflects the limits of current data and methods or other sources of randomness. Closing this gap between the best survey-based model and this bound would require raising recall from 66% to between 83% and 99% – the recall achieved by the upper-bound model under our least and most conservative assumptions (see Table S5) – without a loss in precision. To put this in perspective, recall improved by just 8 percentage points (from 58% to 66%) across all rounds of the survey part of the challenge (if only submissions with high precision are taken into account).

**Combining models does not improve performance**

Predictive performance could potentially be improved through alternative modelling approaches. Several findings suggest this would be difficult, at least in a setting like PreFer. For the register-based models, very different modelling approaches converged on similar performance levels (see Figure 2 and the description of approaches in SI, section 3). Moreover, top-performing teams were mostly consistent in poorly predicted cases: 21% of people in the holdout set who had a child were misclassified by all top-5 models (Figure 3; details in SI, section 2.7).

We also explored whether strategies for combining the best-performing models improve performance. We tested three strategies: (1) aggregating predictions for the holdout set, (2) using predictions from different models as features, and (3) combining features from different models. Combining models can reduce error due to the wisdom of the crowd: either because models make different random errors, so averaging them can cancel out some of the errors[54] or because different submissions use different variables and methods, and one model may capture aspects of the underlying data-generating process that another misses. However, none of these strategies led to performance gains (see SI, section 2.8), even though some cases misclassified by one top model were correctly predicted by another.

For the survey data, a substantially better-performing approach is unlikely. Top-performing models again converged on similar performance and poorly predicted cases (Figure 3). Approaches to aggregating submissions (the same as for the administrative data) did not lead to gains in performance (see SI, section 2.8).

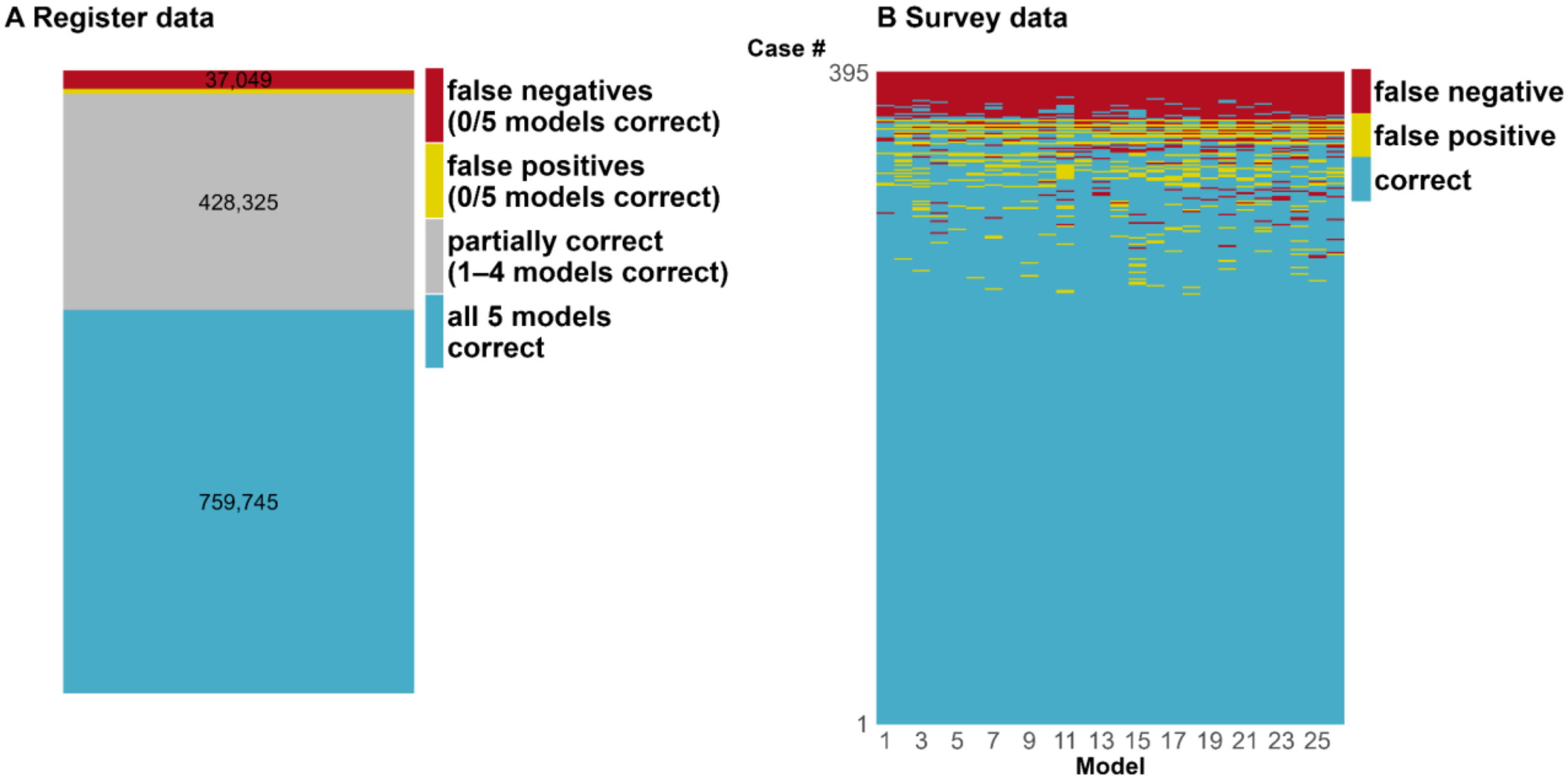

Figure 3. Cases that are consistently hard to predict across models. A Register data: a stacked bar shows the number of cases in the holdout set, grouped by predictions across the 5 top-performing models: false negative by all, false positive by all, inconsistently predicted, or correctly predicted by all. Count for false positives (n = 9,724) not shown due to small bar size. B Survey data: the heatmap displays individuals in the holdout sample (rows) and correctness of predictions by different models (columns); 26 models that outperformed the fertility intentions baseline were used. Patterns across the rows highlight which individuals are consistently difficult to predict across models. Plotting a heatmap for the register data is not possible since exporting individual-level values outside of the secure access environment is not allowed.

### Larger sample sizes alone are unlikely to substantially improve performance

We tested whether larger sample sizes could improve performance. For the register-based model, performance gains become marginal beyond approximately 100,000 training observations (Figure 4). For the survey-based model, we expanded the training sample by retrieving missing outcomes via register linkage (SI, section 2.9); gains similarly diminish beyond approximately 1,000 observations. These results suggest that although larger datasets can yield better predictions, major improvements from an increase in the sample size alone are unlikely for either source.

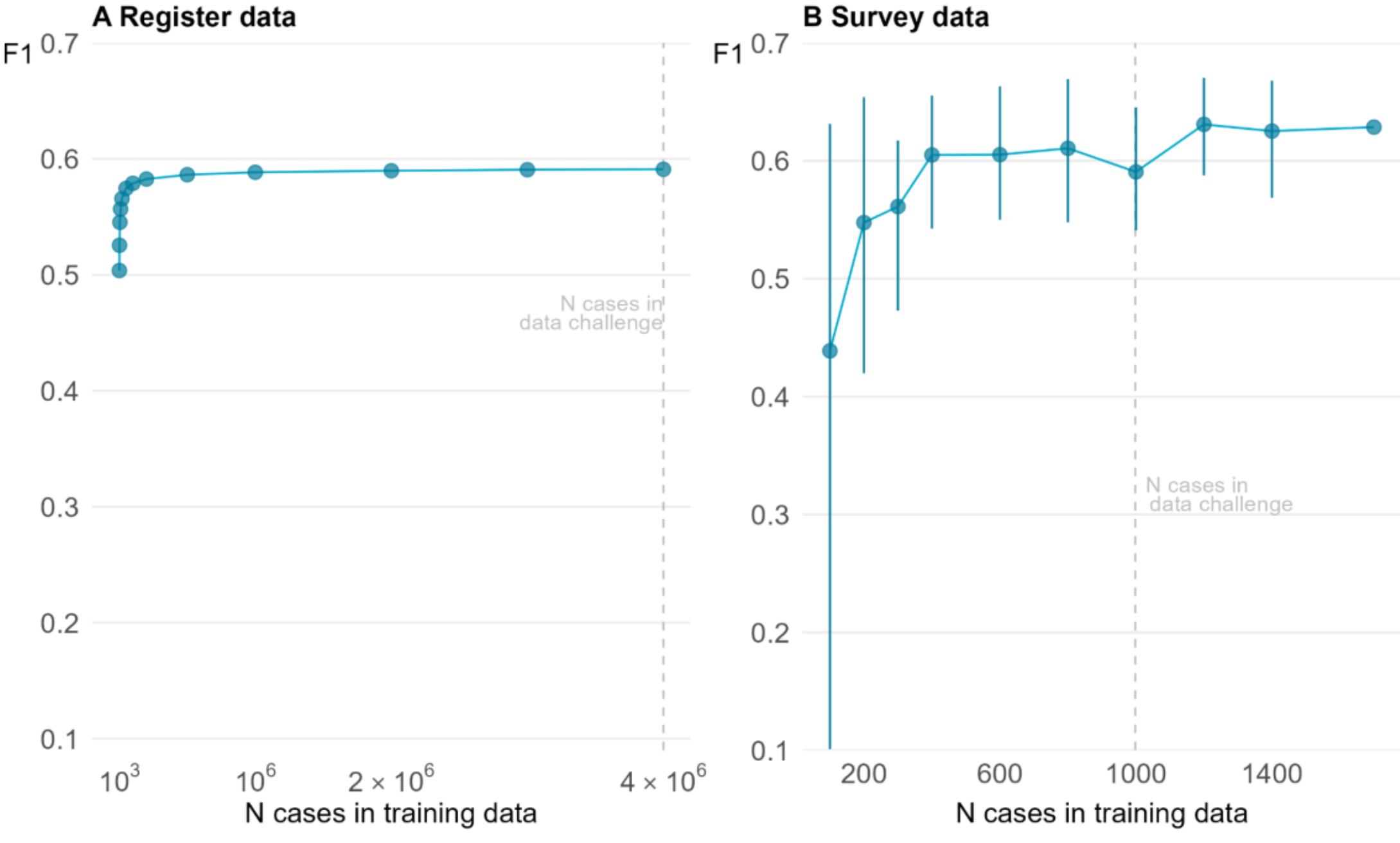

Figure 4. Learning curves for the best register-based and survey-based models. F1 scores are shown for models retrained on training samples of increasing size and evaluated on the original holdout sets. Points represent mean F1 across 30 random subsamples at each training size (except for 2M and 3M, where only one random sample is used). For the survey-based model (B), the training sample was expanded from 987 to 1,688 observations by retrieving outcomes from the register data in the secure environment for respondents who consented to linkage (see details in SI, section 2.9); error bars show 95% confidence intervals based on the 2.5th and 97.5th percentiles of these 30 values. Error bars are omitted from the register learning curve (A) as they overlap in the figure at small sample sizes and are negligible at large sample sizes; see SI Figure S13 for the plot with confidence intervals.

### Improvement from combining register and survey

Administrative data and survey data have different strengths, such as full-population coverage and high-resolution measurement for the former, and the inclusion of fertility intentions and other subjective measures for the latter. We tested whether combining them improves predictions using a straightforward approach: adding predicted probabilities from the best register-based model as an additional feature in the best survey-based model. As a baseline we used the score of the best survey-based model retrained on the cases where register predictions were available (F1 = 0.63, see details in SI, section 2.10). Adding register-based predictions raised F1 to 0.68. Experiments varying the register model's training sample, features, and overlap with survey training data point to two sources of improvement: additional variables and a larger sample size enabling better pattern learning (see details in SI, section 2.10).

## Discussion

Even with high-quality survey and full-population register data that capture many determinants of fertility, state-of-the-art statistical learning methods, and substantial computing resources, dozens of researchers could not very accurately predict who would have a child within three years. This result establishes realistic expectations for individual-level predictions of fertility and possibly for similar life-course outcomes.

Our setup allows us to explore what constrains predictive performance in PreFer. One commonly discussed constraint is limited sample size[22,25]. While modelling approaches beyond those applied in PreFer or an even richer set of features could benefit more from a larger sample[55,56], our findings suggest that insufficient sample size is not the most important reason for the difficulty of predicting births, and perhaps similar life outcomes. Increasing sample size for survey data via register linkage yielded only minor improvements, and reducing register data lowered performance only under substantial reductions. Similarly, the convergence of diverse modelling approaches and no gains from ensembling suggest algorithmic limitations were not the primary constraint in PreFer, but this conclusion may also not generalise to settings with more computational resources and data.

Another constraint is missing or underutilised information. For example, social interactions, (epi)genetic data[39,57,58], and health records[59,60] were absent in PreFer or inaccessible due to sensitivity. The gap between observed performance and the predictive upper bound suggests that missing features constrain both datasets. Avenues for improvement include incorporating health and genetics data[61], better harnessing high-resolution longitudinal data[49], or leveraging strengths of different data sources via transfer learning[62]. However, gains may remain limited given typically small effect sizes[21,39].

The final constraint is chance. It introduces stochastic variation, or aleatoric uncertainty that cannot be eliminated with better knowledge, data, or algorithms[23]. We were uniquely able to quantify part of this variation by simulating outcomes based on findings from reproductive medicine on the chance of conceiving a child and a successful pregnancy and accounting for unplanned births due to chance events. These stochastic processes in our simulation model impose a meaningful constraint on prediction.

Our estimate of this stochastic variation should be interpreted cautiously. Only part of it may be a genuine aleatoric uncertainty while the rest may be epistemic uncertainty, which better data and methods could, in principle, reduce[23]. Some variation in fecundability and fetal survival may be shaped by genetics and environmental exposure that could potentially be captured. More sophisticated knowledge of reproduction may similarly reduce what we now attribute to chance. Yet even with such advances, prediction would require data that will likely remain unavailable, such as cycle-specific biological factors and cellular-level events. In this sense, much of the variation we attribute to chance, even if not truly irreducible, is likely irreducible in practice.

At the same time, we likely underestimate stochastic variation because we have only quantified such variation arising from the physiology of couples trying to conceive a child and unplanned births. Chance influences health, employment, partnership formation, and other life domains affecting fertility. Stochastic variation due to chance events in these domains is harder to quantify. An additional source of uncertainty is the difficulty of estimating what fraction of births result from contraceptive failure[63], which leads to substantial variation in our estimates of the predictive upper bound. Nevertheless, even under conservative estimates, our results indicate that chance plays a substantial role in fertility outcomes. This highlights the importance of studying chance and luck, which are currently largely absent from fertility theories and from social science theories in general[64].

In PreFer, the best survey-based model outperformed the best register-based model despite the register's vastly larger sample size and detailed longitudinal coverage. The small gap suggests that registers cover many fertility determinants, yet points to information that administrative records do not fully capture. Much of the survey advantage comes from fertility intentions variables, contributing to debates about their reliability[65] and measurement[28,66]. Their key role in survey-based models suggests that fertility intentions summarise latent factors that are not captured in either survey or register. These might include perceived costs and rewards of childbearing[67] or other factors difficult to observe directly.

More broadly, PreFer provides a reusable benchmark for individual-level prediction. Because full-population registers largely remove sample-size constraints and are continuously updated, the same prediction task can be rerun on successive cohorts to track how predictability changes as data, methods, and social conditions evolve. This is especially valuable for fertility, whose trajectory in post-demographic-transition societies is uncertain[68]: repeating the challenge over time could show whether births become more or less predictable, and thus whether the mechanisms driving fertility behaviour are stable or shifting.

Predicting whether someone will have a child within three years is a specific prediction task and, at the same time, a good test case for how well individual lives can be predicted at all. Taken together, our results show that even under conditions about as favourable as the social sciences currently allow – full-population registers, rich survey data, state-of-the-art methods, and the combined effort of 147 researchers – the outcome proved only moderately predictable. Combining this data challenge with a simulation of reproductive biology let us separate the error that could potentially be reduced from the part that is likely irreducible, and show that chance in conception and pregnancy alone sets a non-trivial ceiling. Beyond fertility, our results have two broader implications: larger administrative data does not automatically result in better individual-level prediction, and chance deserves a more central place in social theory than it now holds. Because register-based challenges can be repeated on updated data, different cohorts, and with new methods, PreFer offers a moving benchmark against which future progress and its limits can be measured.

## Methods

### Register data

Register data comes from Dutch registers on population, taxes, education, employment, social benefits, housing, and neighbourhoods, managed by Statistics Netherlands. We selected all datasets potentially relevant to predicting fertility, except those containing highly sensitive personal information (medical prescriptions and treatments, healthcare costs). For most of the selected datasets, the coverage starts in 1994-1995 (see SI Section 1.1.1).

The target population is all Dutch residents aged 18–45 on December 31, 2020 (N = 5,878,530) (see SI Section 1.1.1 for details). We randomly split the target population into training (70%) and holdout (30%) sets, ensuring that entire households were assigned to the same set to reduce the risk of data leakage. If married or registered partners lived in separate households, they were also placed in the same set (further split and evaluation-set details in SI, section 1.1.1).

PreFer participants could use 2020 or earlier data for everyone, including holdout and out-of-target individuals. This was done to enable the construction of network characteristics (e.g. whether a person's siblings recently had a child), as network members of the training set individuals could appear in the holdout. The outcomes were withheld for all but the training set. In addition to the selected datasets, we provided the PreFer Base dataset with 2020 information with a subset of variables from the provided datasets for individuals and their married or registered partners (see SI Section 1.1.1 for details).

Despite the precautions about data leakage, one potential source of residual leakage was identified relating to Statistics Netherlands' methodology for identifying unregistered cohabiting partners; this is described and evaluated in SI section 2.11.

### Survey data

We used ten LISS Core study modules (comprising surveys on family, education, work, income, housing, assets, health, religion, social ties, personality, politics and values; fielded annually or biennially since 2007/08) and the monthly Background survey of household demographics (see details in SI Section 1.1.2). Our study sample is LISS members aged 18–45 in 2020 who participated in at least one Core study module in 2007-2020 (N = 6,877). Due to attrition, outcomes from 2021–2023 were available for only 1,382 respondents (see details in SI section 1.1.2).

We split the sample into training and holdout sets at the household level to limit data leakage, stratifying households by whether any member had a child to ensure both sets had similar outcome rates. This yielded a training set of N = 6,418 (outcomes known for 987) and a holdout set of 395 members with known outcomes (further details in SI, section 1.1.2).

Three LISS-based datasets were prepared for PreFer participants: a main training dataset combining Core studies modules (2007–2020) with selected Background survey variables, a supplementary dataset for out-of-target participants (under 18 or over 45 in 2020), and a background dataset with monthly household data (see SI Section 1.1.2 for details).

## Calculating the outcome

The outcome is defined as having at least one new biological or adopted child in 2021–2023. For the register data, we used the dataset Kindoudertab (listing legal parents) and Persoontab (birthdates) to identify whether each target group member had at least one child born between 2021–2023. For the survey data, we determined the outcome using survey responses from 2020 to 2023 on the number of children (biological and adopted) that a person had. If missing, we supplemented it with Background survey data on children's birthdays.

## Performance metric

We evaluated models with the F1 score for the positive class (having a new child) – the harmonic mean of precision and recall – which rewards a model for correctly identifying people who had a child while penalising it for false positives, important for imbalanced outcomes (other metrics in SI, section 3). The F1 score requires a classification threshold. Participants submitted binary predictions in the register phase and code that included the classification threshold in the survey phase, so the choice of threshold was part of their submission, and several teams in both phases tuned it to maximise F1 (see SI, section 3). The baselines and the predictive upper bound use a threshold of 0.5.

## Submission

The challenge ran in two overlapping phases. The first phase, based on the survey data, ran from 1 April to 16 September 2024. To enable reproducibility, submission in this phase was based on software containers (Docker); participants sent their code to be run on the holdout data (see details in SI, section 1.2). Participants could submit only five times to prevent overfitting to the holdout set. In total, 41 teams made 69 valid submissions (that ran successfully on the holdout data and weren't labelled as test submissions).

The second phase (register data), took place from 1 July to 25 October 2024. It began once results from the first three rounds of the first phase were available. Due to strict data access restrictions, participation was limited to five teams selected based on their interim results in phase 1 and the potential of their approaches for the register data; seven additional teams joined as part of the ODISSEI Summer School in Computational Social Science. Teams not selected continued working on the survey data. Overall, 12 teams participated and made 11 valid register-based submissions. Unlike the first phase, participants submitted predicted values for the holdout set along with the trained model and code for preprocessing and training. The remote secure environment through which participants could access the register data prevented an approach equivalent to that used for the survey data. Teams submitted only once because most teams used an evaluation subset of the training data to evaluate intermediate models.

## Predictive upper bound

We assume several sources of randomness in the reproductive process that we quantify via a simulation model[69,70]. Among people attempting to conceive, outcomes vary due to fecundability (monthly probability of conception given unprotected intercourse), its age-related decline until sterility, and fetal survival (probability that a conception results in live birth). Even among individuals with identical fecundability, the month of conception is stochastic – one person may conceive in the first month of trying, while another with identical characteristics only during the third year. This timing determines whether a birth occurs within the 3-year observation window. In the model, each agent is randomly assigned a fecundability at age 15 and an age of sterility, drawn from empirical distributions[70–72] that determine their age-specific fecundability trajectory. Fetal survival is treated as a stochastic event with rates that vary with age but are constant across agents of the same age. As a result, agents with identical attributes can experience different outcomes: some conceive immediately and carry to term, while others take longer, do not have a child within the 3-year window, or experience pregnancy loss.

A further source of randomness is contraceptive failure among people trying to avoid pregnancy who then carry the pregnancy to term. Minimising the chance of a birth would require abstinence, sterilisation, or the most effective methods of contraception; to model a realistic baseline, we instead assume typical population contraceptive use. Ideally, we would model monthly contraceptive failure rates by method, accounting for use patterns and responses to unplanned pregnancies (e.g. abortions). In the absence of such data, we use estimates of unplanned births. In the Netherlands, the share of unplanned births is estimated at 17%[51–53]. However, most likely, not all unplanned births result from contraceptive failure. Unplanned pregnancies also occur when no contraception is used, or it is used incorrectly[73], and because use and correctness of use vary by age, education, and income[74], some of these births are potentially predictable. In the LISS target group, 3.6% (12 of 337) of partnered women who reported not expecting another child nonetheless had one within three years. We conservatively set the proportion of births from contraceptive failure to 1%, but simulate several alternative proportions (see SI section 2.6).

We quantified a predictive upper bound as follows. We assumed a simplified scenario in which the only determinants of having a child within three years are age, partnership status, and fertility intentions; everyone has an unambiguous, unchanging fertility intention (shared within couples) and acts consistently on it. The upper bound thus reflects predictability when all relevant predictors are fully known, and the only source of predictive error is the random outcome variation described above.

From the LISS survey, we selected women in our target group with available partnership status and fertility intentions (N = 1,231). We then generated a synthetic population by drawing with replacement from these real respondents – matched either to the survey data (N = 1,382) or to a 500,000-person sample representing the register data (a smaller sample sufficed as simulation variance was minimal). For each agent, we simulated a fertility outcome from three selected determinants – age, partnership status, and fertility intentions (measured in 2020) – taken from their sampled real counterpart. Agents with a partner and an intention to have a child within 3 years attempted to conceive continuously and were stochastically assigned a conception and whether this conception developed into a live birth, based solely on age-specific probabilities for that agent. Agents without a partner were modelled as having no births, irrespective of their fertility intentions. Partnered agents without a positive fertility intention were also assigned no births, except births resulting from contraceptive failure that are randomly assigned to some of them. We derived the number of such births from the number of planned births (births to partnered agents with positive intentions) and an assumed share of births due to contraceptive failure. For a 1% share, births from contraceptive failure = planned births * (0.01/0.99).

We then fit a logistic regression with age, age$^2$, partnership status, and fertility intention on part of the simulated data and evaluated it on a holdout set using the F1 score. This score represents the theoretical maximum achievable with perfect knowledge of these predictors. Any residual error reflects only the chance variation introduced by the simulation – time-to-conception variability, pregnancy loss, and contraceptive failure. We repeated this procedure on 100 bootstrap samples to obtain confidence intervals. To verify that results were not sensitive to the choice of model, we fitted a random forest, which produced the same results (see SI, section 2.6).

We model fertility intentions as binary (to have or not have a child within three years), stable over the three-year window, with individuals acting accordingly. We acknowledge, however, that real-life reproductive decision-making is often more ambivalent[75]. We use the term “unplanned births” because this indicator is available for modelling births resulting from contraceptive failure, even though the planned/unplanned dichotomy is contested and not all pregnancies that occur due to contraceptive failure are regarded as unplanned by the individuals[76].

Certain aspects of human reproduction, affecting probabilities of conception and fetal survival, are inherently random. For example, during meiosis, random DNA recombination and de novo mutations produce unique gametes[77], many of which are not viable[78,79]. Stochastic genomic imprinting errors can also disrupt normal development[80]. Random X-chromosome inactivation in female embryos[81,82], when skewed, can contribute to miscarriage[83]. However, it is important to note that perhaps not all the variation we model is random. Some variation in fecundability and intrauterine mortality may reflect genetic differences[39,58,84] and environmental exposures[40,85]. Accounting for these factors would reduce the share of variation we currently treat as stochastic. This share would also be reduced with (even) better contraception methods. Our upper bound, therefore, reflects predictability given current biological knowledge and contraceptive technology.

We nonetheless regard this upper bound as conservative for four reasons. First, our model uses relatively high estimates of fecundability[72,86]; lower fecundability would lengthen times to conception and add unpredictable variation. Second, we assume that agents not wanting a child within three years never change this intention and all continuously use contraception, and that those without a partner and without a positive intention have zero chance of a birth. Third, we assume that all partnered agents wanting a child begin trying immediately and continuously with a consistently available partner who shares their intention. In reality, many may delay attempts, may not try continuously for three years on end when a partner may be temporarily unavailable or have different fertility intentions and stop trying at a certain point if attempts are unsuccessful. All this would reduce their likelihood of having a child within three years and make outcomes harder to predict. Fourth, we treat fecundability, age at sterility, and fetal survival as uncorrelated (we drew fecundability and age of sterility independently from empirical distributions, and applied age-specific fetal survival rates uniformly across all agents). If such correlations exist and are positive (e.g., lower fecundability associated with higher pregnancy loss), accounting for them would likely increase the unpredictable variation in our model rather than decrease it. This is because individuals would experience several reproductive challenges at the same time that would amplify differences in outcomes among those with identical observable characteristics.

## Ethical approval

This research was approved by the ethical committee of sociology at the University of Groningen (SOC-2425-S-0002).

## Data Availability

The LISS panel data is available at https://www.lissdata.nl/how-it-works-archive. Prepared PreFer survey datasets are available via SANE https://odissei-data.nl/facility/secure-analysis-environment-sane/. The register data is not publicly available; access is only granted for scientific purposes to vetted researchers affiliated with authorised institutions. Academic institutions can apply for authorisation through Statistics Netherlands. The PreFer Base dataset is available via the CBS Data Storage https://odissei-data.nl/facility/cbs-data-storage/.

## Code Availability

Scripts for creating PreFer datasets from LISS data are available on the project page in the LISS data archive at https://doi.org/10.57990/f3ge-3a61. The code to produce the register-based PreFer data (the outcome variable, Base dataset, and train-test split) is available at the same page in the LISS data archive. The scores and metadata about all PreFer submissions and code to reproduce the analysis presented in this paper are available at Dataverse https://dataverse.nl/previewurl.xhtml?token=12c08a33-c240-40f5-9c1f-977e6e16bd35.

## Acknowledgements

This work is supported by a VIDI grant (VI.Vidi.201.119) from the Netherlands Organization for Scientific Research (NWO) to GS. In this paper, we use non-public microdata from Statistics Netherlands (CBS). We also use data from the LISS (Longitudinal Internet studies for the Social Sciences) panel administered and managed by Centerdata (Tilburg University, The Netherlands) and from the LISS panel data linked to the CBS microdata by Statistics Netherlands. Funding for the panel's ongoing operations comes from the Domain Plan SSH and ODISSEI since 2019. The initial set-up of the LISS panel in 2007 was funded through the MESS project by the Netherlands Organization for Scientific Research (NWO). Access to the CBS microdata, the ODISSEI Benchmark Platform, the ODISSEI-SICSS Summer School, and the development of the LISS harmonised dataset are financed by the ODISSEI Roadmap Project financed by NWO, grant number 184.035.014. We thank SURF (www.surf.nl) for the support in using the National Supercomputer Snellius. This work was supported by the Netherlands eScience Center (see also SI, section 4).

## Affiliations:

[1]Department of Sociology, Faculty of Behavioural and Social Sciences, University of Groningen; Groningen, The Netherlands.

[2]Interuniversity Center for Social Science Theory and Methodology, University of Groningen; Groningen, The Netherlands.

[3]Department of Sociology, Princeton University; Princeton, NJ, USA.

[4]Erasmus School of Social and Behavioral Sciences, Erasmus University Rotterdam; Rotterdam, The Netherlands.

[5]Department of Methodology and Statistics, Faculty of Social and Behavioral Sciences, Utrecht University; Utrecht, The Netherlands.

[6]Netherlands eScience Center; Amsterdam, The Netherlands.

7EYRA; The Hague, The Netherlands.

[8]LISS panel - Quality & Innovation, Centerdata, Tilburg University; Tilburg, The Netherlands.

[9]Department of Sociology, Stanford University; Stanford, CA, USA.

[10]Office of Population Research, Princeton University; Princeton, NJ, USA.

[11]Center for Information Technology and Policy, Princeton University; Princeton, NJ, USA.

[12]Nuffield College, University of Oxford; Oxford, UK.

[13]Department of Sociology, Faculty of Social and Behavioral Sciences, Utrecht University; Utrecht, The Netherlands.

[14]Department of Statistical Sciences, University of Padua; Padua, Italy.

[15]Department of Philosophy, Sociology, Education and Applied Psychology, University of Padua; Padua, Italy.

[16]Research Center for Cognitive Science and Artificial Intelligence, Tilburg University; Tilburg, The Netherlands.

[17]Helsinki Institute for Demography and Population Health, University of Helsinki; Helsinki, Finland.

[18]Sociology, Tilburg University; Tilburg, The Netherlands.

[19]Department of Sociology, Demography Unit, Stockholm University; Stockholm, Sweden.

[20]Department of Political and Social Sciences, European University Institute; San Domenico di Fiesole, It aly.
[21]Oxford Population Health, University of Oxford; Oxford, UK.
[22]Social Research Institute, University College London; London, UK.
[23]Intelligent Systems Group, Bernoulli Institute for Mathematics, Computer Science and Artificial Intellig ence, University of Groningen; Groningen, The Netherlands.
[24]Evolutionary and Population Biology, University of Amsterdam; Amsterdam, The Netherlands.
[25]Data Infrastructure and Open Science, Netherlands Interdisciplinary Demographic Institute (NIDI-KNA W); The Hague, The Netherlands.
[26]Department of Digital and Computational Demography, Max Planck Institute for Demographic Researc h; Rostock, Germany.
[27]Research centre for education and labour market, Maastricht University; Maastricht, The Netherlands.
[28]Department of Epidemiology and Data Science, Vrije Universiteit; Amsterdam, The Netherlands.
[29]Applied Economics, Erasmus University Rotterdam; Rotterdam, The Netherlands.
[30]Department of Statistical Sciences, University of Bologna; Bologna, Italy.
[31]Department of Sociology, Stockholm University; Stockholm, Sweden.
[32]Research Institute of Child Development and Education, University of Amsterdam; Amsterdam, The Ne therlands.
[33]DTU Compute, Technical University of Denmark; Kongens Lyngby, Denmark.
[34]Socio-medical Sciences, Erasmus School of Health Policy and Management; Rotterdam, The Netherlan ds.
[35]Department of Cognitive Science and Artificial Intelligence, Tilburg University; Tilburg, The Netherlan ds.
[36]Department of Psychology, Princeton University; Princeton, NJ, USA.
[37]Department of Intelligent Systems, Tilburg University; Tilburg, The Netherlands.
[38]Demography, University of California, Berkeley; Berkeley, CA, USA.
[39]Leverhulme Centre for Demographic Science, University of Oxford; Oxford, UK.
[40]Population Research Centre, Faculty of Spatial Sciences, University of Groningen; Groningen, The Neth erlands.
[41]Psychology, University of Amsterdam; Amsterdam, The Netherlands.
[42]Department of Methodology and Statistics, Tilburg University; Tilburg, The Netherlands.
[43]Independent Researcher.
[44]Faculty of Science, University of Geneva; Geneva, Switzerland.
[45]Department of Computer Science, Stony Brook University; Stony Brook, NY, USA.
[46]Pharmacoepidemiology and Clinical Pharmacology, Utrecht University; Utrecht, The Netherlands.
[47]Human Geography, Planning and International Development, University of Amsterdam; Amsterdam, Th e Netherlands.
[48]Netherlands Interdisciplinary Demographic Institute (NIDI-KNAW); The Hague, The Netherlands.
[49]Computer Science, Princeton University; Princeton, NJ, USA.
[50]Faculty of Electrical Engineering, Mathematics and Computer Science (EEMCS), Faculty of Behaviour al, Management and Social Sciences (BMS), University of Twente; Enschede, The Netherlands.
[51]Sociology Department, University of Amsterdam; Amsterdam, The Netherlands.
[52]Independent Researcher; Padua, Italy.
[53]Centre for longitudinal studies, University College London; London, UK.

[54]INVEST Research Flagship Centre, University of Turku; Turku, Finland.
[55]Department of Computer Science, University of Copenhagen; Copenhagen, Denmark.
[56]Computer Science, Harvard University; Boston, MA, USA.
[57]University of Helsinki; Helsinki, Finland.
[58]Institute for Advanced Study in Toulouse, Toulouse School of Economics, Université Toulouse Capitole ; Toulouse, France.
[59]Network Science Institute & Khoury College of Computer Science, Northeastern University; Boston, M A, USA.
[60]Estonian Institute for Population Studies, School of Governance, Law and Society, Tallinn University; T allinn, Estonia.
[61]Independent Researcher; Florence, Italy.
[62]Faculty of Psychology, Dresden University of Technology; Dresden, Germany.
[63]School of Business and Economics, Maastricht University; Maastricht, The Netherlands.
[64]Labor Demography, Max Planck Institute for Demographic Research; Rostock, Germany.
[65]Harvard Data Science Initiative, Harvard University; Cambridge, MA, USA.
[66]Data Science Group, Institute for Basic Science (IBS); Daejeon, South Korea.
[67]Families and Generations, Netherlands Interdisciplinary Demographic Institute (NIDI-KNAW); The Ha gue, The Netherlands.
[68]Research Institute for Statistics and Information Science, University of Geneva; Geneva, Switzerland.
[69]Department of Psychology, Faculty of Behavioural and Social Sciences, University of Groningen; Groni ngen, The Netherlands.
[70]Sociology Department, University of Oxford; Oxford, UK.
[71]Center for Language and Cognition Groningen, Faculty of Arts, University of Groningen; Groningen, T he Netherlands.

*Corresponding author: Elizaveta Sivak, e.sivak@rug.nl

## Supplementary Information

### 1. PreFer design

#### 1.1 Data

##### 1.1.1 Register data

Brief descriptions of the datasets used in PreFer are provided in Table S1. The full codebooks (in Dutch) are available via the links to the CBS website (see links in Table S1). Descriptions and lists of variables are also available via the DOI links to the ODISSEI portal.

**Table S1** The list of datasets based on the Dutch registers available to the participants of the data challenge.

| **Name of the dataset** | **Description** | **DOI** |
|---|---|---|
| Gbapersoontab [CBS] | Personal characteristics of people in the population register [87]: age, gender, migration status of individuals and their parents. Coverage: from 1994 | https://doi.org/10.57934/0b01e4108071ba40 |
| Gbaburgerlijkestaat Bus [CBS] | Past and current civil status of persons included in the population register (single, married, registered partnership, divorced, etc.), with start date and end date of each status. Coverage: from 1995 | https://doi.org/10.57934/0b01e410803b37e0 |
| Gbaverbintenispartnerbus [CBS] | Partner's IDs to link partner's characteristics | https://doi.org/10.57934/0b01e410801f93bf |
| Gbamigratiegebeurtenisbus [CBS] | Migration dates. Coverage: from 1995 | https://doi.org/10.57934/0b01e4108021261a |
| Gbahuishoudensbus [CBS] | Household identifiers, composition, and start/end dates (due to member additions/removals, address changes). Coverage: from 1995 | https://doi.org/10.57934/0b01e410802125bb |
| Hoogsteopltab [CBS] | Highest achieved (meaning: with a diploma) and followed (meaning: without a diploma) level of education. Coverage: from 1999 | https://doi.org/10.57934/0b01e410801fd716 |
| Inpatab [CBS] | Personal income. Coverage: from 2011 | https://doi.org/10.57934/0b01e41080372fbd |
| Inhatab [CBS] | Household income. Coverage: from 2011 | https://doi.org/10.57934/0b01e41080371196 |
| Vehtab [CBS] | Household wealth. Coverage: from 2006 | https://doi.org/10.57934/0b01e4108037363f |
| Spolisbus [CBS] | Employment (excluding self-employed) and characteristics of jobs. Coverage: from 2010 | https://doi.org/10.57934/0b01e410804cb681 |
| Secmbus [CBS] | Personal socio-economic category (employed, self-employed, studying, etc.). Coverage: from 1999 | https://doi.org/10.57934/0b01e410803432a6 |
| Nabijheidkindopvtab [CBS] | Proximity to childcare (from objects, e.g. living places). Coverage: from 2012 | https://doi.org/10.57934/0b01e41080238887 |
| Gbaadresobjectbus [CBS] | Past and current addresses of people registered in the Netherlands: address ID, date of registration at this address, and end date of address (when a person moved out or died). Coverage: from 1994 | https://doi.org/10.57934/0b01e410802154d6 |
| Vbowoningtypetab [CBS] | Type of housing. Coverage: from 2020 | https://doi.org/10.57934/0b01e41080053c864 |

| Vslgwbtab [CBS] | Municipality and neighbourhood codes of residence objects to link external data about municipalities and neighbourhoods. Coverage: from 1995 | https://doi.org/10.57934/0b01e41080236a82 |
|---|---|---|
| **Networks:** | | |
| Burennetwerktab [CBS] | Neighbours network. Coverage: from 2009 | https://doi.org/10.57934/0b01e410807607b7 |
| Colleganetwerktab [CBS] | Colleagues network. Coverage: from 2009 | https://doi.org/10.57934/0b01e410807607b1 |
| Familienetwerktab [CBS] | Family network. Coverage: from 2009 | https://doi.org/10.57934/0b01e41080760802 |
| Huisgenotennetwerktab [CBS] | Household network. Coverage: from 2009 | https://doi.org/10.57934/0b01e4108076077c |
| Klasgenotennetwerktab [CBS] | Classmates network. Coverage: from 2009 | https://doi.org/10.57934/0b01e4108075f996 |
| **External data**: results of the Dutch elections | Results of the Dutch general elections (2017), provincial elections (2019), and municipal elections (2020) by municipality | DOI not available |

**Target population and train-holdout splits**

The target population was defined by selecting all individuals aged 18–45 at the end of 2020 (born between January 1, 1975, and December 31, 2002) who were registered at a residential address in the Netherlands on December 31, 2020 (initially, N = 6,092,379). "Residents" are individuals registered at a Dutch address who have stayed in the Netherlands for at least four months[87]. We then excluded 213,849 individuals (3.5%) who were no longer residents at the end of 2023, as their fertility outcomes for the 2021–2023 period could not be reliably determined. Temporary moves abroad did not lead to exclusion if individuals were again registered at the end of 2023, since births abroad are added to the register upon re-entry.

After the 70/30 training-holdout split, the holdout set was further divided into two parts: one-third for evaluating intermediate submissions and two-thirds for the final submissions (see Figure S1 for sample sizes). In addition, 10% of the training set was flagged as an evaluation set so that participants could test their models without replicating the splitting procedure themselves.

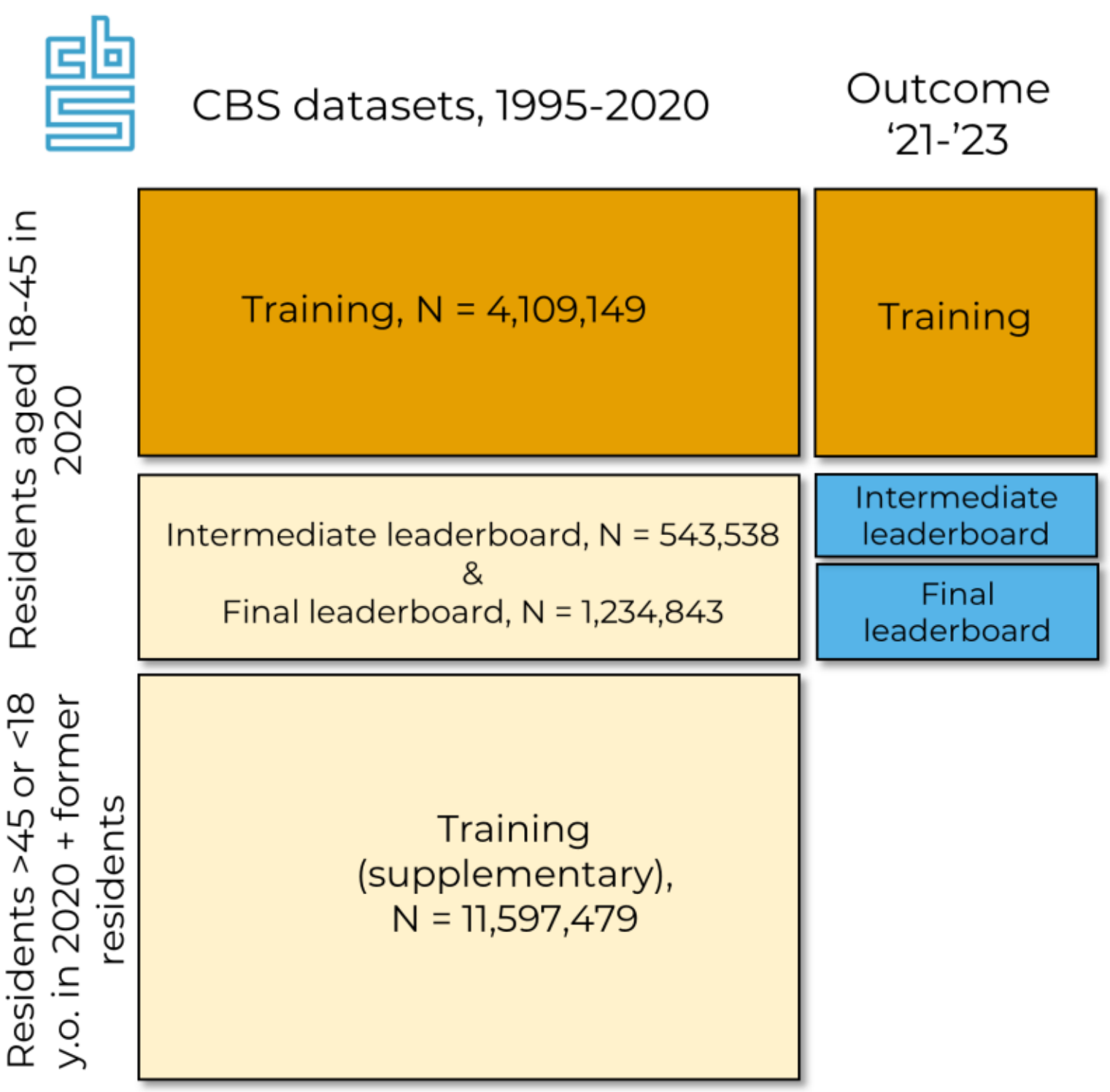

Figure S1. Register data used in PreFer. The data challenge participants could use the features up to 2020 and the outcome for people in the training set. Features about people in the holdout sets and supplementary training set were also available for training, e.g. for constructing network characteristics. The outcome for people in the holdout and supplementary groups was not available to the challenge participants to reduce the risk of data leakage.

**PreFer Base dataset**

To help PreFer participants familiarise themselves with register data and develop initial models without having to link many datasets, we prepared the Base dataset. It included 108 variables drawn from most of the datasets selected for PreFer, supplemented by variables we constructed from these datasets, such as the total number of children, the age of the youngest child, and linked partner information for those married or in registered partnership. This Base dataset was prepared for individuals in the target population and for 2020 Dutch residents outside of the target population (younger than 18 or older than 45 in 2020).

The variables were mostly from 2020, selected based on research on fertility determinants, and covered topics such as demographics of the person, their parents, and partner (if available); partnerships, education and employment of the person and partner, personal and household income, assets, household characteristics, and neighbourhood characteristics. The full codebook is available in the replication package and on the PreFer website https://preferdatachallenge.nl/posts/posts/2024-07-16-base-csb-dataset.html.

1.1.2 Survey data

Brief descriptions of the LISS datasets used in PreFer are provided in Table S2. The full codebooks (in Dutch and English) are available via the links in the table.

**Table S2** LISS panel studies that are included in the merged dataset.

| Name | Description | DOI |
|---|---|---|
| Background variables | Socio-demographic variables at the household level and individual level. Filled in by a contact person about all the household members when the household joins the panel and updated monthly. Includes age, gender, position within household, civil status, type of housing, urbanity, occupation, income. | https://doi.org/10.57990/qn3k-as78 |
| Core Study modules: | | |
| Health | Physical and mental health self-assessments and medication use, lifestyle habits. | https://doi.org/10.17026/dans-ze3-5uk9 |
| Religion and Ethnicity | Religious upbringing, religious affiliation, religiosity, religious orthodoxy. Nationality, origin, ethnic identification, language proficiency and use. | https://doi.org/10.17026/dans-xkw-t8dm |
| Social Integration and Leisure | Social contacts, core discussion network, loneliness. Leisure activities, voluntary work and informal care, social media usage. | https://doi.org/10.17026/dans-zaf-casa |
| Family and Household | Family structure, social support from family, parenting and children, domestic responsibilities, child education and childcare. | https://doi.org/10.17026/dans-xkd-5hp5 |
| Work and Schooling | Employment status and history, job satisfaction and conditions. Education, qualifications, and training. | https://doi.org/10.17026/dans-x26-tttv |
| Personality | Subjective well-being, personality traits. | https://doi.org/10.17026/dans-x5h-4cxd |
| Politics and Values | Political engagement, affiliation, and attitudes. Values and social attitudes. | https://doi.org/10.17026/dans-zms-r5rz |

| Economic Situation: Assets | Different kinds of assets, loans and debts. | https://doi.org/10.17026/dans-z2r-n69z |
|---|---|---|
| Economic Situation: Income | Different sources of income, subjective standard of living. | https://doi.org/10.17026/dans-24y-dkqk |
| Economic Situation: Housing | Housing characteristics, expenditures, satisfaction with housing. | https://doi.org/10.17026/dans-zgv-9qky |

**Train-test split of the LISS survey data**

To reduce the risk of data leakage, we randomly selected households rather than participants into the training or holdout data, meaning that all participants of one household are either in the training data or in the holdout data. To do that, we selected the households where the outcome was available for at least one household member and grouped these households into two groups: (1) where at least one person had a new child, (2) where no one had a new child. Then we randomly selected 30% households from each group. We assigned all participants who belong to these households to the holdout set and excluded people for whom the outcome is missing from the holdout set (64 individuals; resulting holdout sample size is 395). All participants from the remaining 70% of households (as well as participants from the households where the outcome was missing for all members) were assigned to the training set (N = 6,418, outcomes known for 987).

To verify whether the participants in the resulting training and holdout groups are similar, we compared the distributions of three variables in the holdout and training sets (excluding participants with a missing outcome): the outcome, age, and the number of waves of the Core Study modules a person participated in (operationalised as answering at least one question). Participants in the training and holdout data were very similar based on these variables: the outcome rate was 21.5% in training and 22.3% in holdout, mean age was 32.7 vs 31.9 years, the number of waves was 34.7 and 34.1 in the last 5 years and 21.2 vs 21.5 in the last 3 years, and the distribution of age and participated waves was similar across both sets, confirming that the split did not introduce systematic differences.

**LISS attrition rates**

The LISS panel started in 2007 when approximately 5000 households comprising 8000 individuals aged 16 years and older were recruited (about 6000 of them being 18–45 years old)[47]. The annual attrition rate is approximately 10%. To counteract this dropout, new panel members are recruited every two years based on the population registers (i.e., refreshment samples), maintaining the representativeness of the LISS panel[88]. Most of our target group – LISS members who participated in at least one Core study before 2020 and were aged 18–45 in 2020 (N = 6,877) – had dropped out by 2021–2023. Despite using both Core surveys and Background variables, we could only calculate the outcome for 1,382 respondents.

**LISS-based datasets prepared for PreFer**

We prepared several LISS-based datasets for PreFer participants:

1. PreFer LISS Main Training Dataset, which merges all Core study modules from 2007–2020. This dataset includes all members of the study sample selected for training (N = 6,418 with outcome data available for 987 of them) and more than 30 thousand variables. The outcome is provided in a separate file. Several variables from the monthly Background survey were also added: for variables that change over time, such as income and partnership status, one value per respondent per year was included; for more stable variables (e.g. gender), a single value was added. All original monthly values remain available in the LISS Background Training Dataset.
2. PreFer LISS Supplementary Training Dataset, which includes the same variables as the LISS main training dataset but covers only respondents outside the target population – those younger than 18 or older than 45 in 2020 (N = 10,644; the outcome is available for 2,806 of them). The outcome for this group is also provided in a separate file. Individuals who belong to the same households as people in the holdout set (161 people) were excluded from this group to reduce the risk of data leakage and do not appear in any PreFer datasets. Some PreFer participants used the supplementary training dataset to increase the training sample size[89].
3. PreFer LISS Background Training Dataset – provides monthly data on approximately 30 variables from the LISS Background survey (2007–2020) for all LISS panel participants in the training set and their household members (12,854 individuals from 4,950 different households).

The same training, outcome, and background files, but for the holdout sample, were available only for the organisers to evaluate the submissions (for sample sizes, see Figure S2).

In all prepared PreFer datasets, original IDs were replaced with pseudonymised IDs generated by our team to prevent linking of outcome-window data (2021–2023) available for download from the LISS data archive.

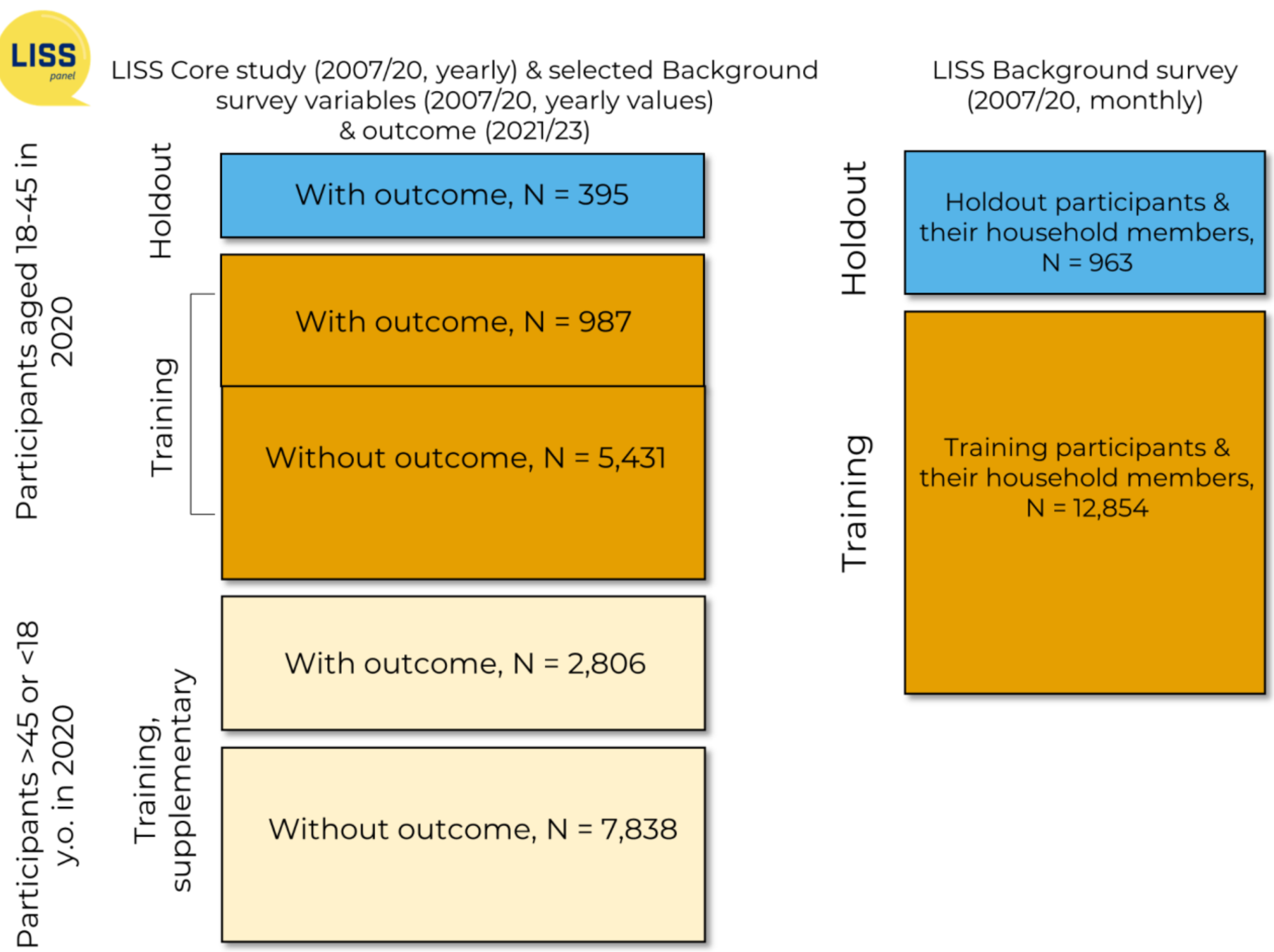

Figure S2. LISS data used in PreFer. Training and supplementary training datasets (brown and yellow) with the outcome were available for the data challenge participants. Holdout sets (blue) were available only for the organisers for the evaluation.

## 1.2 Participation

We recruited participants for the PreFer data challenge through professional networks, conference presentations, mailing lists, and online communities. In total, 314 individuals from diverse backgrounds (mainly demography, sociology, computational social science, data science, and computer science) registered. Seven additional teams (17 people) joined as part of the ODISSEI Summer School in Computational Social Science.

In the first phase, 153 participants from 43 teams made at least one submission (89 submissions in total). Of these participants, 147 from 41 teams produced at least one valid submission, defined as a submission that ran successfully on the holdout data and was not labelled as a test submission. These 41 teams produced 69 valid submissions in total in the first phase.

In the second phase, 12 teams worked with the more restricted register data; they produced 11 valid submissions; 2 other submissions did not run correctly on the holdout data.

One of the teams was unable to complete their approach within the challenge timeline, finalising their model after the deadline. Overall, these teams included 45 people; however, due to data access rules and costs of access, not all team members had access to the register data. In total, 26 people out of 45 had access to the register data.

To foster reproducibility, in the first phase (survey data), participants submitted a GitHub repository containing their model, preprocessing and prediction code, a file specifying required software (packages), a setup script (Dockerfile), and a description of their approach. Participants used a template GitHub repository to prepare these scripts; https://github.com/eyra/fertility-prediction-challenge, a copy of which is archived in the replication package at Dataverse. Their code was run inside a Docker container built from these files, generating predictions on the holdout data. Before submission, automated GitHub checks tested whether the code could run successfully on synthetic data matching the structure of the holdout set. This system ensured that the code is run in the same environment as the participants used themselves and that it is reproducible. Submissions were made via the Next platform http://next.eyra.co/. The submission platform supported submissions in R and Python. After each round of submission, participants could see their score on the holdout data on an anonymised leaderboard, which displayed the scores of all submissions without revealing which team made which submission – only their own score was identified. This allowed participants to see their relative ranking without revealing how other teams' approaches were performing.

## 1.3 Computing environment

### 1.3.1 In PreFer

In the survey-based part of PreFer, data challenge participants could run their computations on computing infrastructure of their choice (for example, their local machines or on remote high-performance computing resources). Only CPU-based computations in R or Python were allowed due to the specifications of the submission platform. The packages used for creating each submission are listed in the submissions' GitHub repositories. Links to all the repositories are available on the PreFer website and in the Dataverse.

In the register-based part of PreFer, selected teams worked with register data in a secure environment: a shared virtual desktop server, called remote access environment or RA, and/or a high-performance computing system – the ODISSEI Secure Supercomputer (OSSC) within the Dutch national supercomputer Snellius[90]. Both environments are very controlled and regulated to maximise security. They limit external connections: there is no internet access; transferring results and code in and out of the environment is regulated and requires approval; exporting individual-level data out of the environment is not allowed;

access is only possible from within the countries in the European Economic Area and a small number of other countries[1].

In RA, only CPUs are available: with 48 GB RAM (default) or 128 GB RAM (if requested). All teams used this environment; access to the RA was available at least for one researcher per team. OSSC provided an alternative with more computing resources on both CPUs and GPUs. Three teams used OSSC because their methods required GPUs: team Cruijff and team SBU EUI (see details about their approaches in SI Section 3); a third team that used GPUs did not produce a valid submission mainly due to complications of working with OSSC. Eight other teams used OSSC to benefit from additional CPUs for speeding up data preprocessing, hyperparameter tuning, or training.

There were several complications in using OSSC, and especially GPUs. For example, time on OSSC is limited and must be requested 5 days in advance. Computations on OSSC require job scheduling through a workload manager rather than interactive execution. Even within a single GPU node, using multiple GPUs requires additional configuration. These constraints contributed to one team being unable to produce a valid submission and affected methodological choices made by teams. One team chose not to develop a graph neural network; instead, they extracted features manually from the network data (teamSGBD). Another team reduced the amount of data that they used because of these limitations; they decided to use CPU in the RA to preprocess the data[49,50].

We focus our description on GPU resources and their usage, as time on GPU nodes was more limited in PreFer than on CPUs (and is typically more limited in general). During PreFer, OSSC offered NVIDIA A100 GPUs (40 GB memory each, with 4 GPUs per node) managed through the SLURM workload scheduler. Time on GPUs was limited to approximately 1,200 GPU-node hours in total. One team (team Cruijff) used GPUs on OSSC for fine-tuning a Llama language model (Llama 3.1 8B Instruct) on preprocessed data (register data converted into 'books of life'[49]) and performing inference[50]. Their computations involved between one and five GPU nodes, in parallel, depending on the task. In total, approximately 1,200 GPU-node hours were used across all stages of the workflow. The second team used OSSC after the PreFer deadline, when additional GPU time was available, to develop a foundational model trained from scratch on the register data turned into text (see details in [48]) (around 4,000 GPU-node hours).

Teams used Python or R for their submissions. Packages and versions of the default installed Python environment are available on the PreFer website https://preferdatachallenge.nl/. Some teams used other Python environments (details available in the code repositories in the replication package at Dataverse).

[1] See more details about the rules here https://www.cbs.nl/en-gb/our-services/customised-services-microdata/microdata-conducting-your-own-research

1.3.2 For producing this paper

The results in this paper were generated by code written in R 4.3.3 and Python 3.10 using packages listed in the scripts in the replication package at Dataverse.

## 2. Analysis of the predictive performance

### 2.1 Calculating confidence intervals

We estimated confidence intervals by bootstrapping the holdout set: we sampled individuals with replacement from the holdout set 1,000 times and calculated a performance metric each time. We then used the 2.5th and 97.5th percentiles of the resulting distribution as the confidence interval. Exceptions are noted where a different procedure was used.

### 2.2 Alternative train-test splits

To test the robustness of the performance of the best survey-based model and to verify that results were not an artefact of overfitting to a particular holdout set, we retrained and reevaluated this model using 25 alternative training-holdout splits by randomly selecting households into the holdout set using different random seeds. Performance on our original split was representative, falling near the middle of the distribution (Figure S3). We did not conduct this check for the best register-based model, as its holdout set of more than one million observations is large enough that performance estimates are not meaningfully affected by the choice of split.

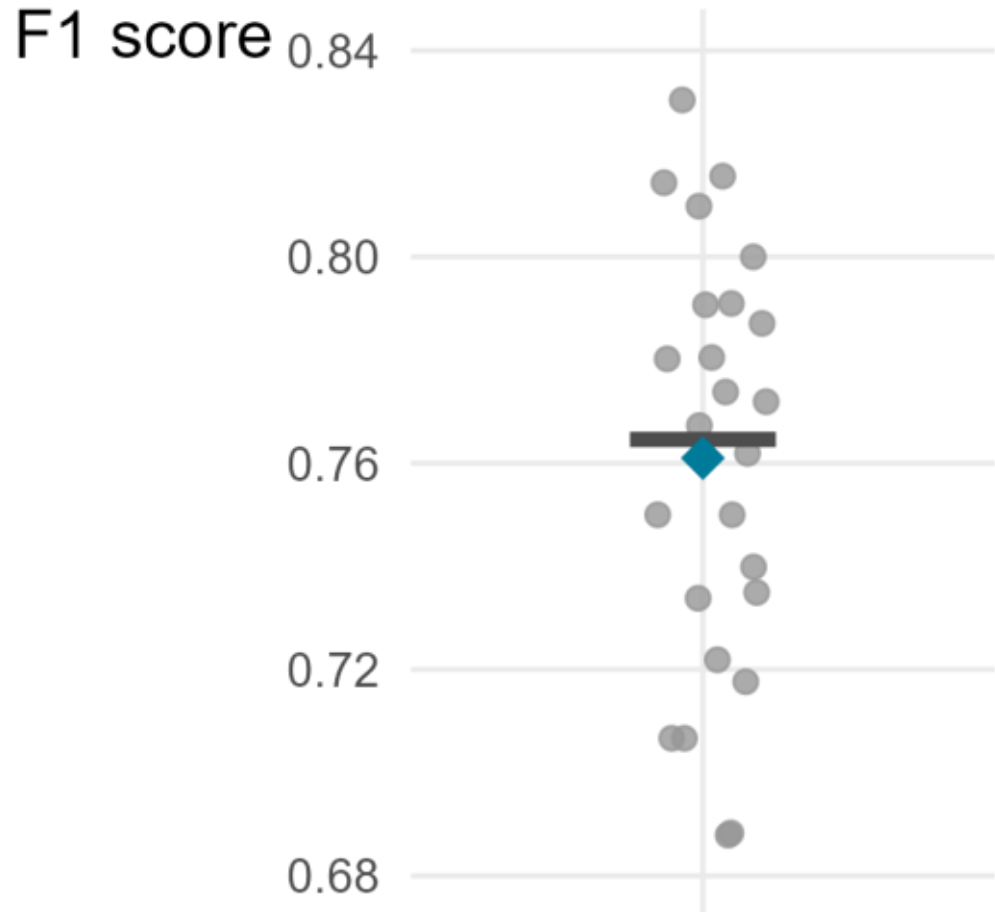

Figure S3. F1 scores of the best survey-based model, retrained using 25 alternative training-holdout splits. The grey line represents the median F1 score across these splits. The blue diamond represents the score obtained from the original split.

## 2.3 Baselines

We use two baselines. The demographic baseline uses logistic regression with six predictors from 2020: age and age$^2$, gender, partnership status, highest level of education, and a composite categorical variable combining information on whether the person has children and the age of the youngest child. The composite variable uses category 0 to distinguish childless individuals, categories 1–7 for individuals with the youngest child aged from under 1 to under 7 years, and the final category 8 for individuals with the youngest child aged 7 or older. For the survey demographic baseline, missing values in partnership status and education were imputed using the mode and missing values in the number of children – using the median. There were no missing values (for individuals with outcome available) in gender and age. Two rare categories in partnership status (separated and widowed) were combined into a single category; this resulted in four final categories: married, never married, divorced, and separated or widowed. See the performance of this baseline for the survey and register data in Table S3.

The fertility intentions baseline uses logistic regression with a single categorical variable derived from two survey questions from 2020: "Do you think you will have more children in the future?" (yes/no/don't know) and "Within how many years do you hope to have your [next/first] child?" (answers include the number of years or 0 if currently pregnant). Based on these two questions, we created a categorical variable with 9 categories: 0 – currently pregnant/partner is pregnant, 1/2/3 – hope to have a child within 1/2/3 years, 4 – within 4-6 years, 5 – within 7-9 years, 6 – within 10 years or later, 7 – does not expect to have children, 8 – does not know if expect children or the answer to the first question is missing. These categories were combined because in 2016-2019, the category 'don't know' was coded as a missing value (i.e. category 'don't know' is present in the 2020 data but absent in previous years). Combining these two categories ensures that the fertility intentions variable can be constructed in the same way for all years.

We also tested an additional fertility intention baseline that followed respondents' fertility intentions literally: 1 if they hoped to have a child within three years, 0 otherwise (again, in a logistic regression model). This model performed slightly worse than the 9-category fertility intention baseline due to lower precision (see Table S3). The reason is that the proportion of those who actually had a child within three years (realisation rate) is substantially lower among people who hoped to have a child within three years than among those with a 1-2-year timeline (see Figure S4). The 9-category model learned this heterogeneity in realisation rates from the data.

There are several potential reasons why realisation rates are lower for people with a 3-year timeline than for those with a 1-2-year timeline. People whose timing is further away might not start trying to conceive immediately, for example, because they underestimate how long it may take to conceive, or they may reconsider their plans before actively trying. Alternatively, they may currently lack necessary conditions (e.g., a partner) and hope these will materialise within three years, but for some, this doesn't happen in time.

Table S3 presents different metrics for the baseline models. $R^2$ (equation 1) ranges from 0 (meaning that the model performed no better than simply using the mean outcome of the training set as predicted probability for all holdout observations) to 1, which indicates perfect predictive accuracy.

Equation 1:

$$R^2_{\text{Holdout}} = 1 - \frac{\sum_{i \in \text{Holdout}} (y_i - \hat{y}_i)^2}{\sum_{i \in \text{Holdout}} (y_i - \bar{y}_{\text{Training}})^2}$$

**Table S3.** Performance of the baseline models

| **Model** | **F1 score** | **Precision** | **Recall** | **MCC** | **$R^2$** | **Log loss** | **Accuracy** | **AUC** |
|---|---|---|---|---|---|---|---|---|
| Demographic baseline (register data) | 0.19 | 0.49 | 0.12 | 0.19 | 0.15 | 0.34 | 0.85 | 0.79 |
| Demographic baseline (survey data) | 0.25 | 0.39 | 0.18 | 0.14 | 0.01 | 0.58 | 0.75 | 0.69 |
| Fertility intentions main baseline (9-category; survey data) | 0.64 | 0.76 | 0.55 | 0.56 | 0.38 | 0.38 | 0.86 | 0.85 |
| Fertility intentions additional baseline (3-year cutoff; survey data) | 0.62 | 0.66 | 0.58 | 0.52 | 0.26 | 0.42 | 0.84 | 0.75 |

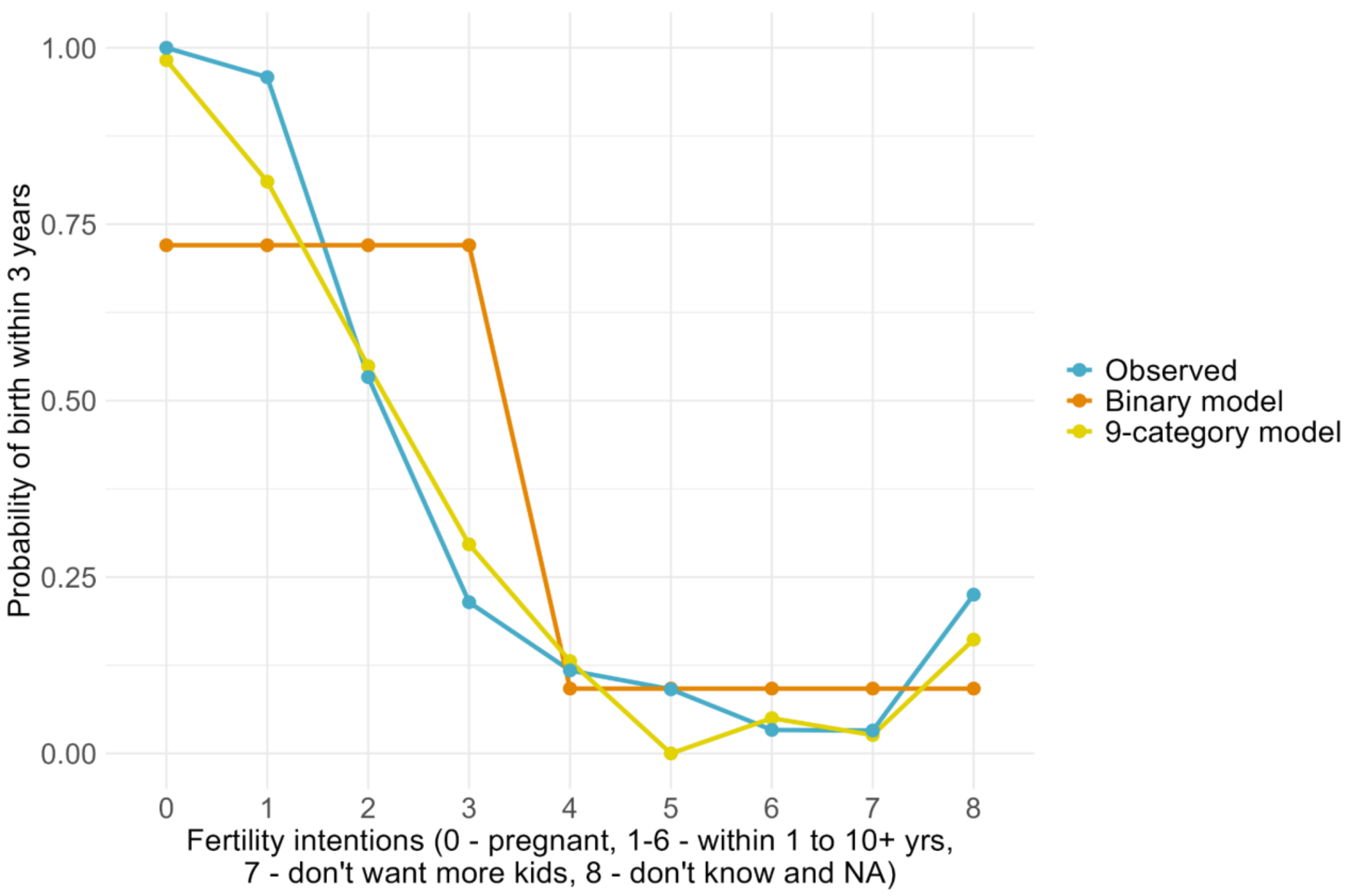

**Figure S4.** Observed and predicted probability of having a child within 3 years (2021–2023) by fertility intentions in 2020.

### 2.4 Excluding fertility intentions from survey-based models

Most of the submitted survey-based models (83% of 69 valid submissions) included variables on fertility intentions. To test the extent to which survey-based models rely on fertility intentions variables, we conducted an ablation study on two models: the best-performing model (Stork Oracle; XGBoost with theory-based selection of 86 variables) and the data-driven model that is likely to discover alternative predictors (teamSGBD; lightGBM with all the available variables from 2017 to 2020). When retrained without all fertility intentions variables, both models showed substantial performance drops. The best model's F1 score decreased from 0.76 to 0.65 (15% relative decrease), while the data-driven model dropped from 0.71 to 0.54 (24% relative decrease). This suggests that even sophisticated machine learning algorithms with access to numerous demographic, socioeconomic, and relationship variables cannot fully compensate for the absence of direct fertility intentions measures. This shows the central role of fertility intentions in these survey-based models.

**Table S4.** Scores of the two models with and without fertility intentions

| Model | F1 score (with fertility intentions) | 95% CI | F1 score (without fertility intentions) | 95% CI |
|---|---|---|---|---|
| Best model (Stork Oracle) | 0.76 | [0.68, 0.83] | 0.65 | [0.54, 0.73] |
| Data-driven model (teamSGBD) | 0.71 | [0.62, 0.78] | 0.54 | [0.43, 0.63] |

## 2.5 Accounting for different base rates in the survey and register data

The share of individuals who had a child within three years was 21.5% in the survey training data, 22.3% in the survey holdout data, and 15% in the register data (both in the training set and in the holdout set). This discrepancy arises because among the survey participants with incomplete responses on the number of children in 2020-2023, we could only determine the outcome for those who had a new child during the years for which information was available, while missing data prevented certainty about non-births, leaving those without a new child underrepresented.

To account for this difference, we did the following. We linked the survey data to the register data in the secure access environment and retrieved register-based outcomes for survey training set participants with missing outcomes, expanding the training set from 987 to 4,388 observations with known outcomes. For many of these people, however, the features from 2018–2020 were missing. This would make the comparison between the model retrained on the larger sample size and the model trained on the original training data unfair. We restricted this expanded sample to participants who had completed the 2020 Family and Household survey to ensure comparable feature availability across the original and expanded training sets. Original training data (participants with available outcomes) also consisted mostly of people who participated in the Family and Household 2020 survey, because that survey was required to calculate the outcome. The outcome prevalence in this updated training set was 15%, and the number of observations with known outcomes was 1,688.

We also retrieved register-based outcomes for the holdout set, but this only reduced prevalence from 22% to 20% – still considerably higher than in the register holdout (15%). We therefore evaluated the retrained survey-based model in two ways and compared its performance to the best register-based model in each case.

In the first evaluation, we evaluated the retrained model on the original survey holdout (prevalence 22%); that resulted in F1 = 0.68, 95% CI [0.59, 0.76]. We additionally compared the mean squared error (MSE) of predictions from the retrained survey-based model and the best register-based model on this holdout. The register-based model had a higher MSE (MSE register = 0.12, MSE survey = 0.09, difference = 0.03, 95% CI [0.01, 0.05]; see Figure S5 for distribution of difference in MSE).

In the second evaluation, we randomly removed positive-outcome observations from the survey holdout to match the 15% prevalence of the register holdout. We repeated this across 1000 random subsamples and evaluated the retrained model on each, which resulted in F1 = 0.63, 95% CI [0.56, 0.69]. The difference in MSE between register-based and survey-based predictions remained positive across subsamples (mean difference = 0.02, 95% CI [0.01, 0.03]), consistent with the first evaluation.

There are several possible reasons for the better performance of the survey-based model on the survey holdout. First, the survey may contain more predictive information, particularly fertility intentions. Second, this result may partly reflect the fact that the survey training and holdout sets are drawn from the same sample, which, while in general representative, may underrepresent certain subgroups. The register-based model, trained on the full population, is less tailored to this particular sample and may therefore be at a slight disadvantage when evaluated on the survey holdout.

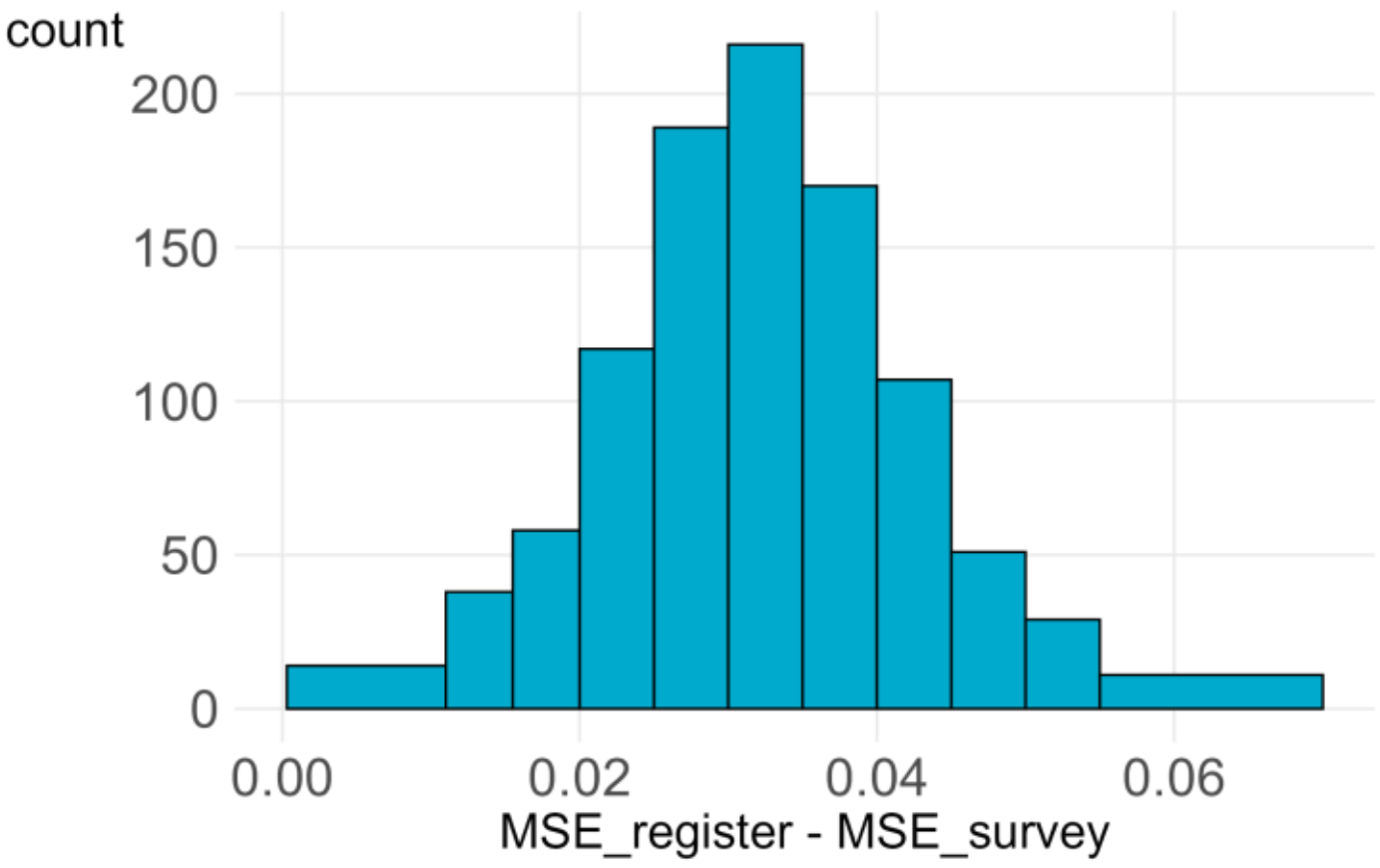

**Figure S5.** Difference in MSE of predictions for the survey holdout set of the best register-based model and the best survey-based model. Two bins are wider than the rest because exporting results from the CBS secure environment for groups smaller than 10 people is not allowed.

### 2.6 Predictive upper bound: estimates for different assumptions

We run simulations described in Methods (see 'Predictive upper bound') across a range of assumptions. In the first condition, we assume that stochastic processes affect only agents who are trying to conceive a child and not those who are currently pregnant or do not want a child (i.e. we assume for all agents who are pregnant that the pregnancy results in a live birth and there are no births due to contraceptive failure). The second condition differs only in that stochastic intrauterine mortality is applied to currently pregnant agents. All other conditions additionally allow for births due to contraceptive failure, with the share varying incrementally from 1% (conservative estimate) to 17% – the estimated overall share of unplanned births in the Netherlands. For each set of assumptions, we estimate a predictive upper bound following the procedure described in Methods. We repeated this procedure 100 times on bootstrapped samples to obtain confidence intervals for the F1 score.

We also tried an alternative modelling approach (random forest instead of logistic regression) to verify that results were not sensitive to the choice of model. We used the same data on the same predictors as in logistic regression: age, partnership status, and fertility intentions. We tune hyperparameters (number of trees, number of variables randomly sampled at each split, and minimum node size) via a grid search and using 10-fold cross-validation on the training set. We then use the best combination in the final model that is refit on the full training set.

Figure S6 shows F1 scores with 95% confidence intervals of the predictive upper bound for different assumptions and for the two modelling approaches (logistic regression and random forest); the results for different models are almost the same.

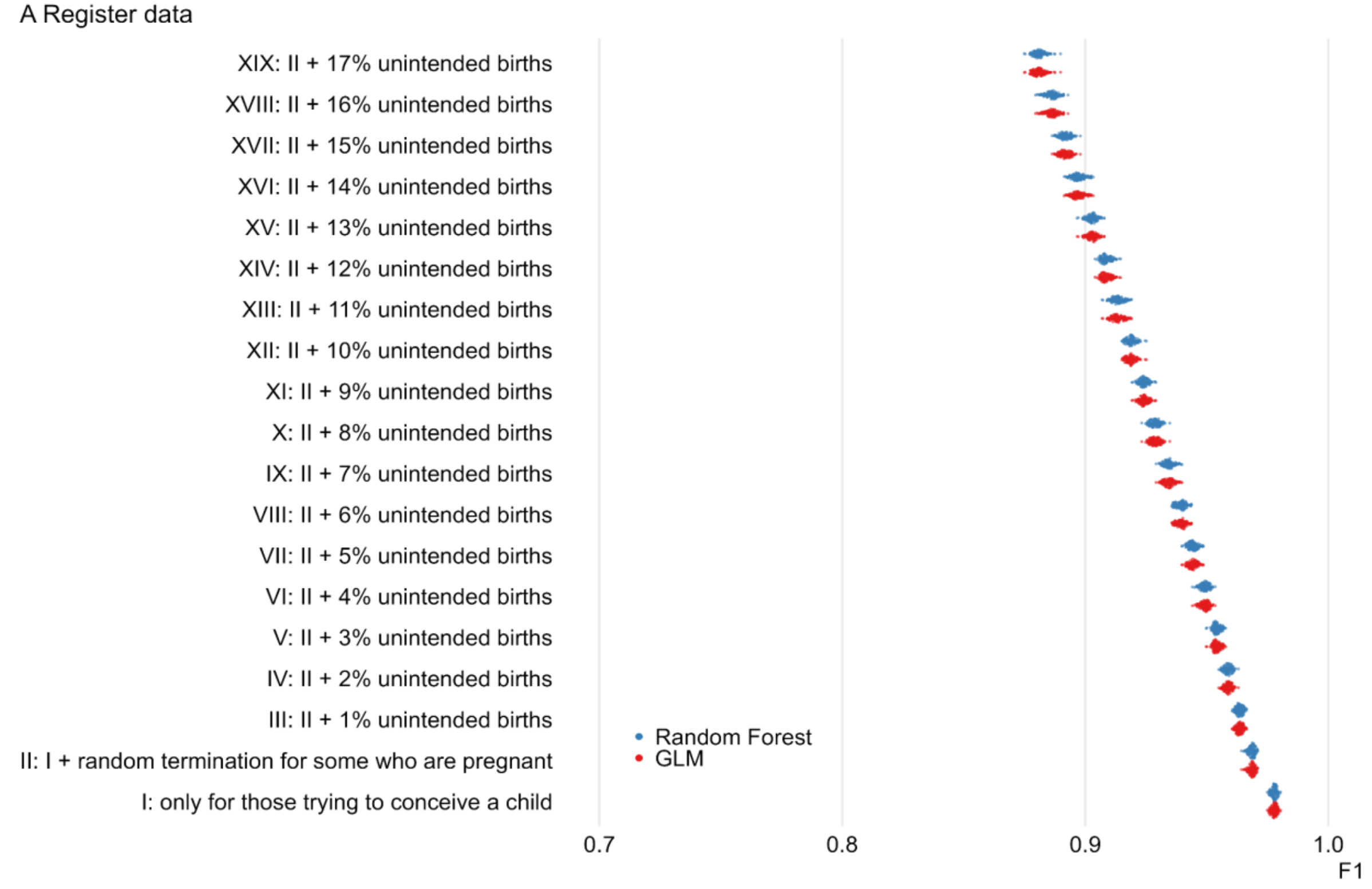

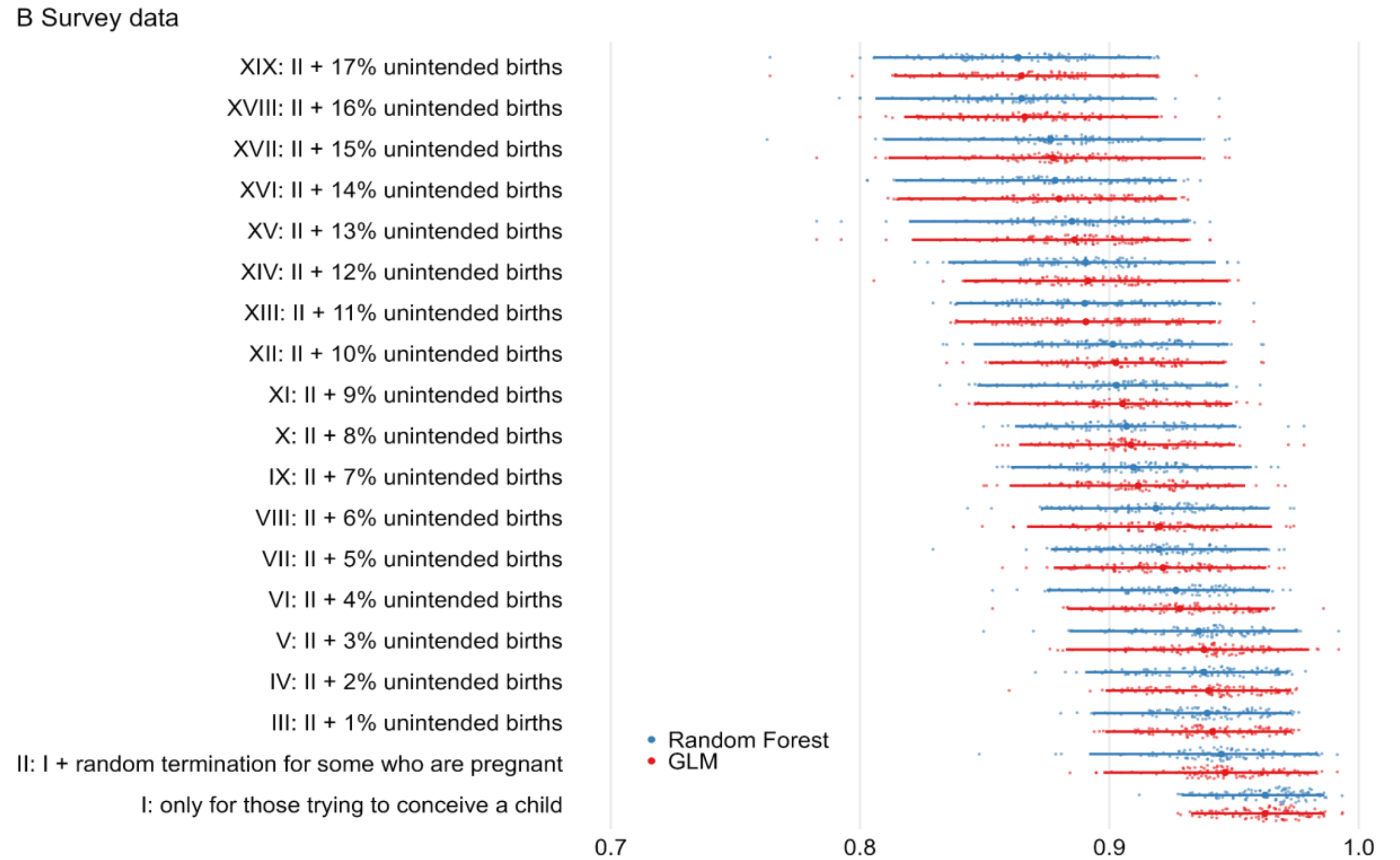

**Figure S6.** Predictive upper bound calculated for different assumptions and proportions of unplanned births (from 0 to 17%); A) for the register data, B) for the survey data. Each row of points represents a distribution of F1 scores. Black horizontal lines show 95% confidence intervals based on these 100 bootstrap values.

Table S5 shows F1, precision, and recall for selected assumptions for the survey data, and F1 only for the register data (we did not calculate precision and recall for the register-based upper bound); values are means across 100 bootstrap samples. For the survey data, the upper bound's precision is stable across assumptions (0.93 in the first two conditions, 0.91 thereafter) while its recall varies: from 0.99 in the first and second condition, which assume no births from contraceptive failure, to 0.98 under a 1% share of such births and 0.83 under a 17% share. This happens because births from contraceptive failure are unpredictable by construction, so adding them adds false negatives without adding false positives.

**Table S5.** Predictive upper bound under selected assumptions about births due to contraceptive failure

| Condition | Survey F1 | Survey recall | Survey precision | Register F1 |
|---|---|---|---|---|
| No intrauterine mortality for pregnant agents, no births from contraceptive failure | 0.96 | 0.99 | 0.93 | 0.98 |

| Random intrauterine mortality applied for pregnant agents, no births from contraceptive failure | 0.95 | 0.99 | 0.90 | 0.97 |
|---|---|---|---|---|
| 1% of births due to contraceptive failure | 0.94 | 0.98 | 0.91 | 0.96 |
| 17% of births due to contraceptive failure | 0.86 | 0.83 | 0.91 | 0.88 |

Figure S7 illustrates for different assumptions, how the gap between the best survey-based model's performance and perfect prediction (F1 = 1) decomposes into two parts: (1) the part attributable to sources of randomness captured in the predictive upper bound, i.e., uncertainty arising from randomness in conception and pregnancy progression that is irreducible under the chosen assumptions, and (2) the remaining part attributable to the limits of current data and methods (potentially reducible uncertainty) and other sources of randomness not accounted for in our simulation (irreducible uncertainty).

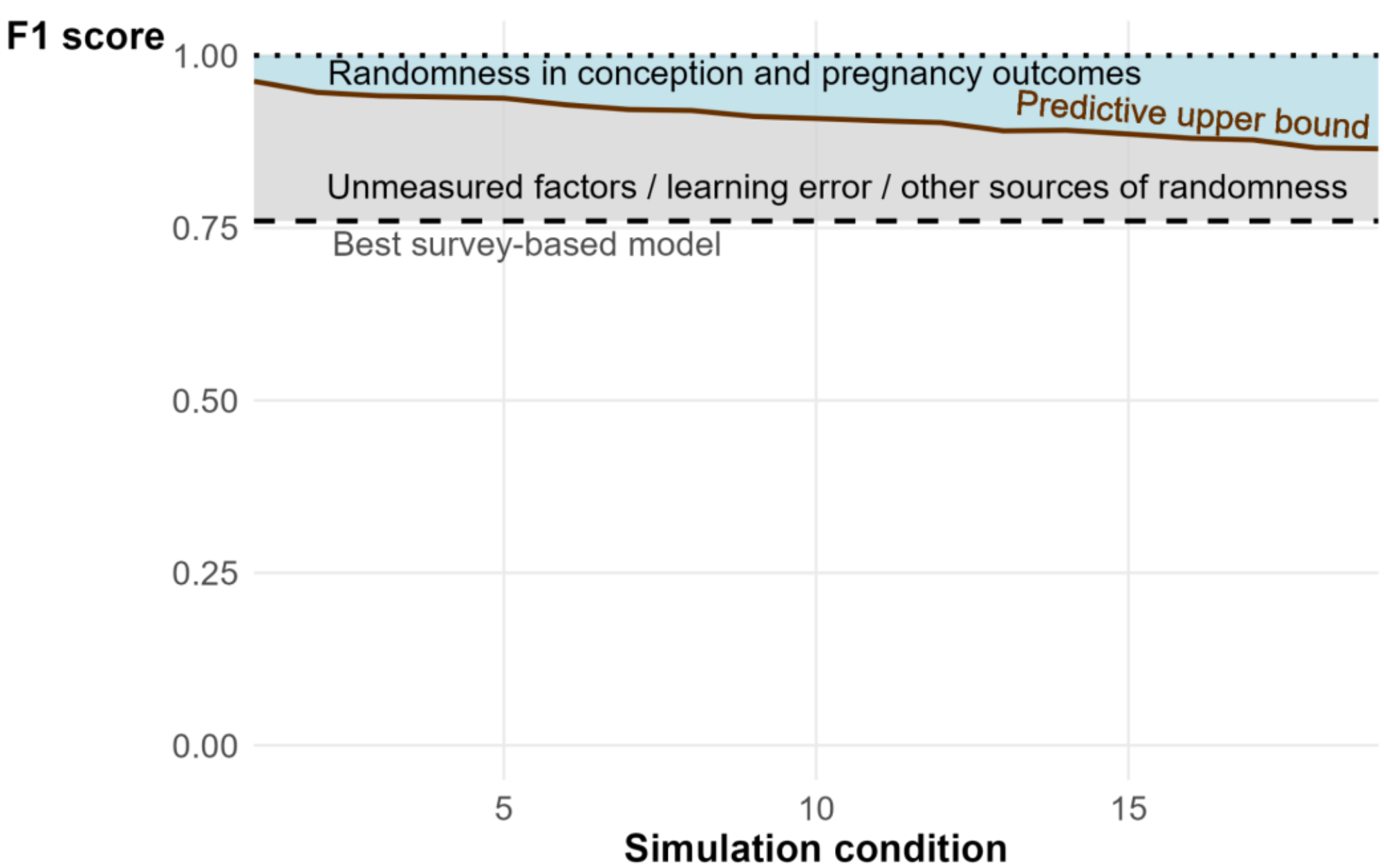

**Figure S7.** Decomposition of the gap between the best survey-based model and perfect predictive performance (F1 = 1), across 19 simulation conditions. Condition 1 assumes no intrauterine mortality for currently pregnant agents and no unplanned births; condition 2 adds stochastic intrauterine mortality; conditions 3-19 additionally vary the share of births due to contraceptive failure from 1% to 17%. The brown line shows the predictive upper bound estimated under each set of assumptions. The blue area shows the part of the gap attributable to sources of randomness captured in the predictive upper bound (irreducible under the chosen assumptions). The grey area represents the gap due to limits in current data and methods (potentially reducible) and other sources of randomness not accounted for in the predictive upper bound.

## 2.7. Hard-to-predict cases

The main text (Figure 3) shows that 21% of people in the register holdout set who had a child were misclassified by all top-5 register-based models. Here in Figure S8, we show the proportion of consistently misclassified cases for both groups: people who had a new child in 2021–2023 and did not have a child in 2021–2023. Top-performing models misclassify a smaller proportion of negative-outcome cases (1%), which is likely due to class imbalance.

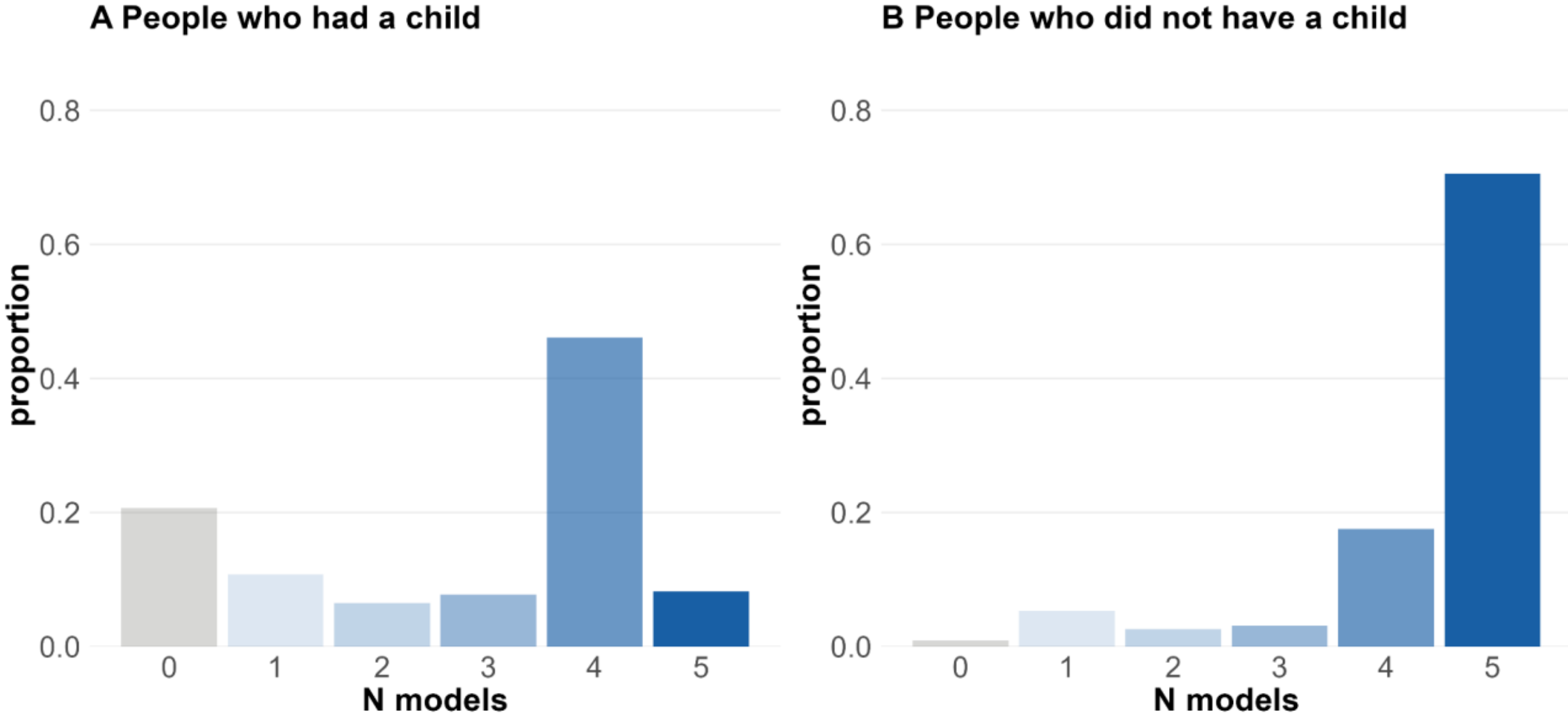

**Figure S8.** Proportion of people in the register data holdout set who were predicted correctly by the 5 top-performing models: A) among those who had a child in 2021–2023, B) among those who did not have a child in 2021–2023.

There are many reasons why some cases are hard to predict: from limited observations and less prevalent outcomes in certain subgroups, to poorer measurement quality (for instance, for migrants, key variables such as education might be more often missing), to the fact that some groups may be disproportionately affected by random chance. Timing of the outcome may also play a role. Some individuals may be hard to predict not because of their inherent characteristics, but because of where they were in their trajectory toward having a(nother) child at the start of the outcome window. If someone was still far from having a child at the beginning of 2021, their characteristics measured in 2020 would likely carry less predictive signal than for someone who was already close to that transition.

Figure S9 illustrates this with recall of the best register-based model calculated separately for individuals in the holdout set who had a child in each month of the outcome period (January 2021 – December 2023). Recall decreases throughout the outcome period, with the steepest decline occurring in the first months, showing that later births are harder to anticipate from pre-2021 data alone.

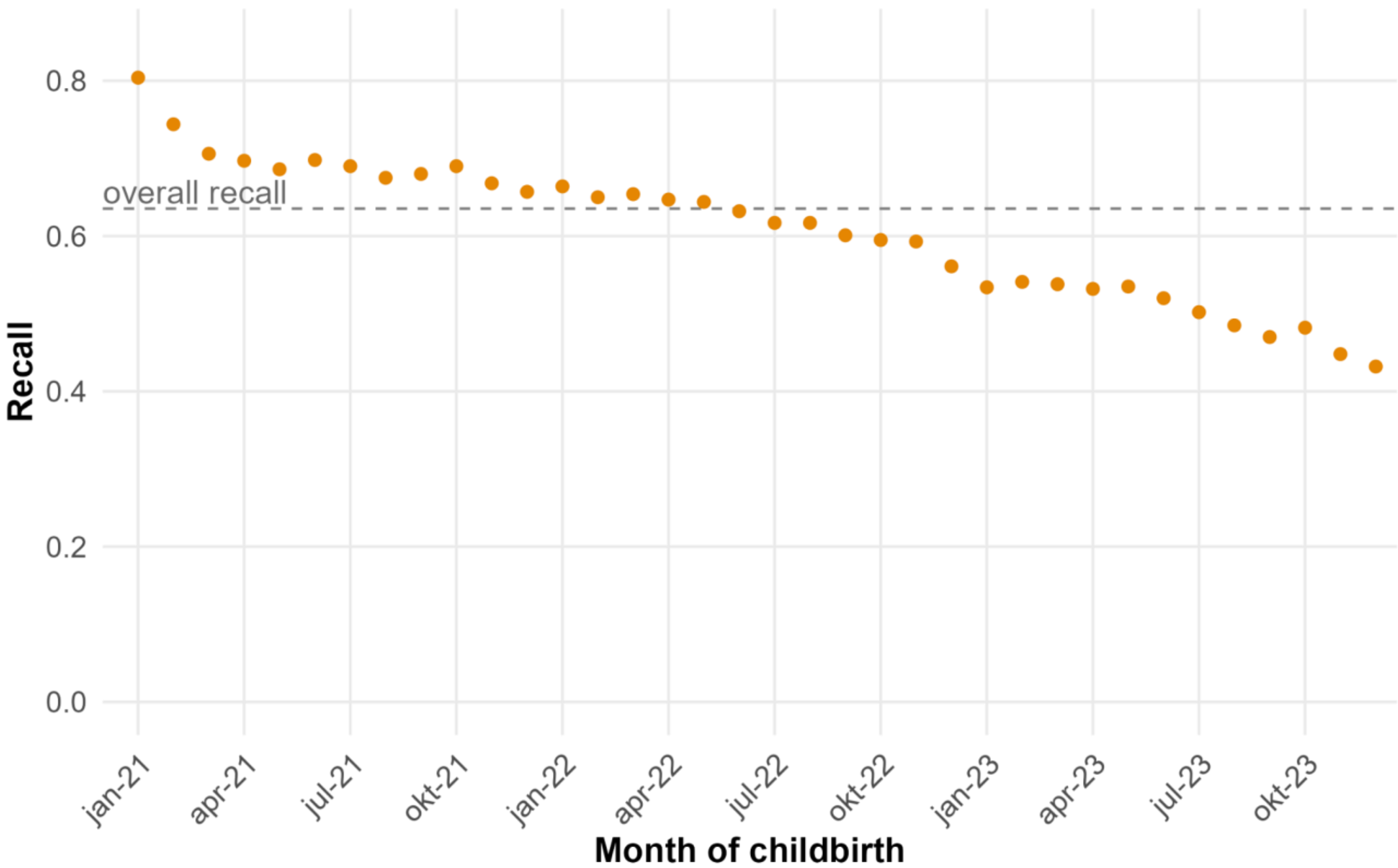

**Figure S9.** Recall by month of childbirth. Each point shows the recall of the best register-based model for individuals who had a child in that month (January 2021 – December 2023). The dashed line indicates the overall recall of this model across the full outcome period.

## 2.8 Combining submissions

We tested three strategies of combining submissions. First, we aggregated predicted classes for the holdout set by averaging predictions across models and applying a classification threshold. We tried aggregating only the top models (10 for the survey data and 5 for the register data) or all available models for each data source (for survey: excluding 9 models where there were missing values in predictions), and also different thresholds, but performance was not improved (see Figure S10).

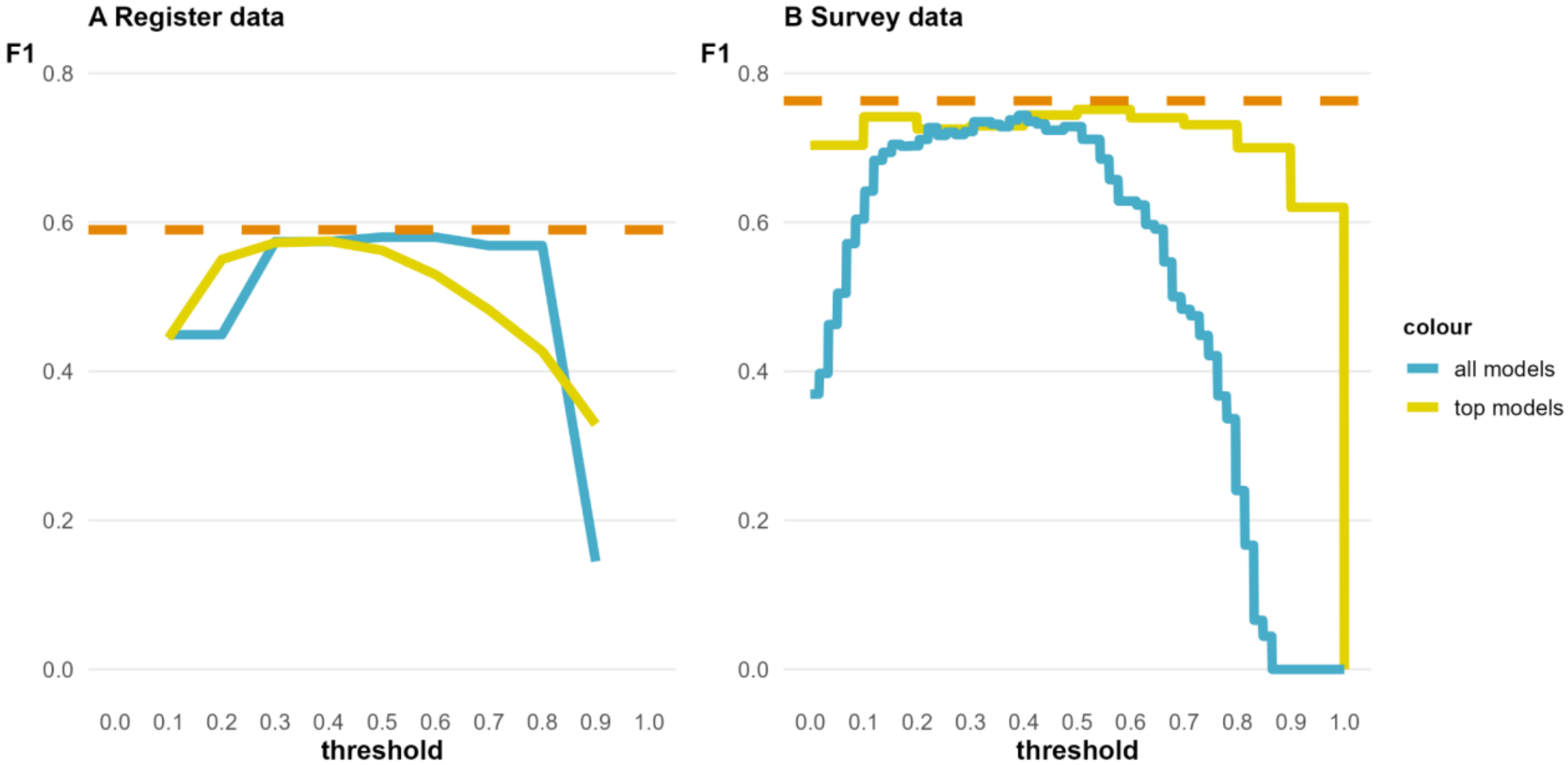

**Figure S10.** F1 scores for aggregated predictions on the holdout set. Aggregated predictions obtained by averaging predictions from different models and applying a classification threshold. A: register data. B: survey data. The dashed line indicates the performance of the best model for each data source. The blue line represents the aggregation of the top five models, and the yellow line represents the aggregation of all available models for each data source.

Second, we used predicted classes from multiple models as features to predict the outcome. To do so, we generated predictions for both the holdout and training sets for survey-based models, and for the training set only for register-based models (as holdout predictions were submitted directly). We were unable to generate predictions for 1 of 69 survey-based models due to a reproducibility issue, and for 2 of 11 register-based models due to difficulties of running them on the secure supercomputer. We also excluded 9 survey-based models because their predictions included missing values; in total, we used predictions from 59 survey-based models. We then trained three algorithms on these predictions as features (for both data sources): elastic net, LightGBM, and LightGBM on PCA-transformed inputs. Hyperparameters were tuned on the full training set for survey data and on a subset for register data (N = 30,000), after which final models were trained on the full training set using the selected hyperparameters. This strategy did not improve predictive performance (see Figure S11).

We repeated this approach using predicted probabilities instead of class labels, as probabilities preserve more information. To obtain predicted probabilities, we slightly modified the submissions so that they output probabilities for the positive class rather than classes. For an additional 5 survey-based submissions, predicted probabilities could not be generated because the models did not support probability outputs in a straightforward way: for instance, some used hard voting, some did not clearly distinguish positive from negative class probabilities, and one produced a constant output of 0 for all cases. We did not include predictions from these models. As with predicted classes, using predicted probabilities did not improve predictive performance (Table S6).

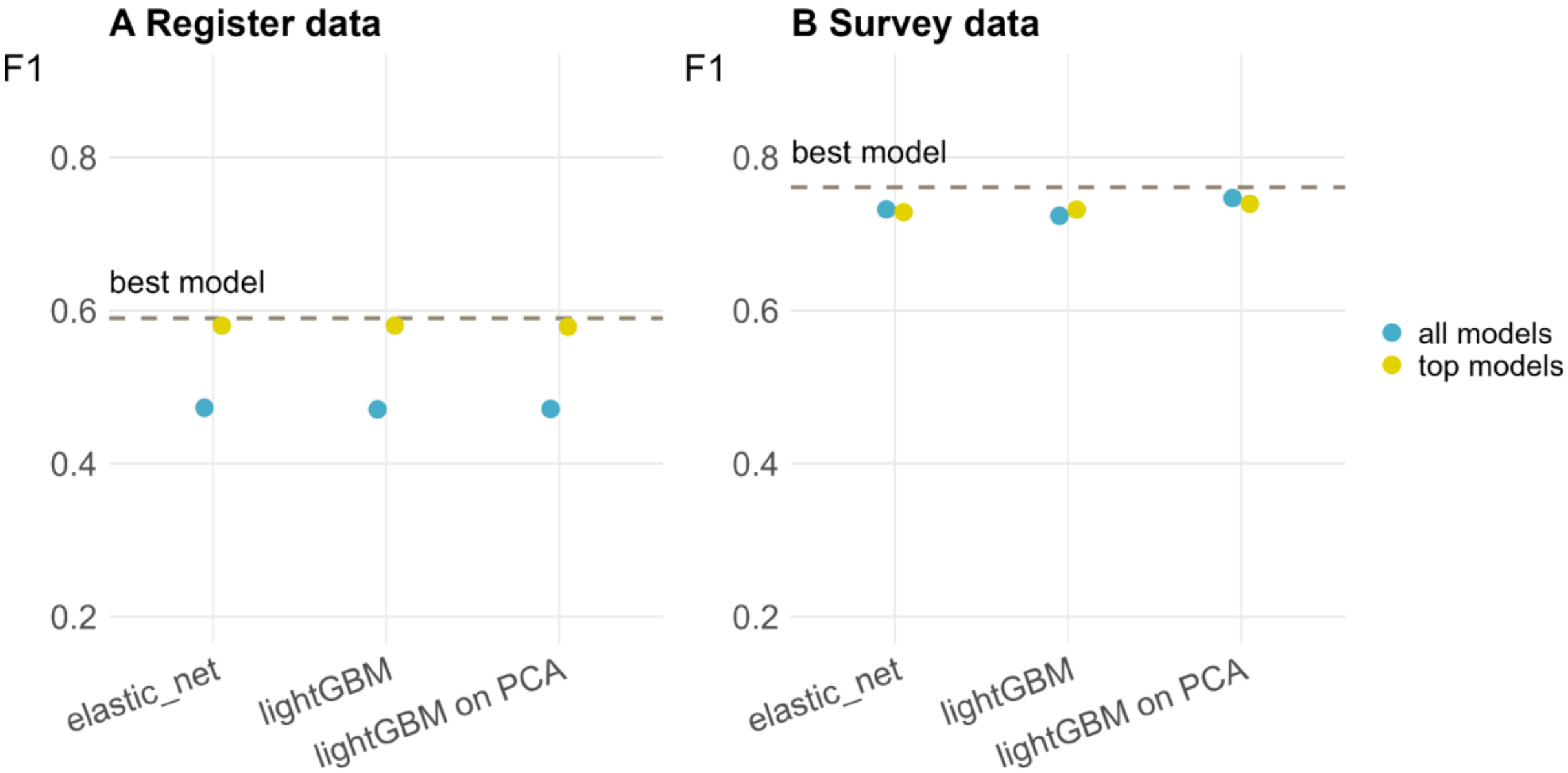

**Figure S11.** Using predicted classes for the training data from different models as features to predict the outcome. Blue dots: results when only top-performing models are used (5 for register, 10 for survey)

**Table S6.** Performance of the ensemble models that use predicted probabilities as features

| Model | Models used | Data | F1 score |
|---|---|---|---|
| Elastic net | All (9) | register | 0.59 |
| lightGBM | All (9) | register | 0.59 |
| lightGBM on PCA (5 components) | All (9) | register | 0.50 |
| Elastic net | Top 5 | register | 0.59 |
| lightGBM | Top 5 | register | 0.59 |
| lightGBM on PCA (3 components) | Top 5 | register | 0.59 |
| Elastic net | All (54) | survey | 0.74 |
| lightGBM | All (54) | survey | 0.73 |
| lightGBM on PCA (5 components) | All (54) | survey | 0.68 |
| Elastic net | Top 10 | survey | 0.71 |
| lightGBM | Top 10 | survey | 0.67 |
| lightGBM on PCA (5 components) | Top 10 | survey | 0.72 |

Lastly, we combined variables from several top-performing models. We retrained the best register- or survey-based model cumulatively, adding blocks of features (grouped by topic) one at a time: first, blocks from the model itself to test if removing features improves performance, and then additional blocks from other models. Once added, no blocks were removed regardless of performance. Each time, we evaluated the model on the original holdout set.

For the register-based experiment, we added features from two models – the 4th and 7th best. Models ranked 2nd-3rd were foundation models and did not use engineered features; models ranked 5th-6th had similar performance to the 7th but fewer features that were not already present in the best register-based model, so the 7th was preferred as the more informative source. The best register-based model (CatBoost) was trained with default hyperparameters, which we also used when retraining with different feature sets.

For the survey-based experiment, additional features were drawn from three models. Models using a purely data-driven approach with thousands of input variables were skipped: since these models select variables internally, the only feasible way to extract features would be via feature importance, which is not justified, as importance scores from one model may not transfer to a different feature set. We therefore drew features only from models with small feature sets. We first added features from the 3rd-best model (since it uses only a few handcrafted features), then variables from the 2nd-best model (which uses several hundred pre-selected variables), and finally handcrafted features from the 7th model. Only variables not already present in the best model were included in each case. The best survey-based model (XGBoost) uses hyperparameter tuning; we retuned hyperparameters at each step using the original grid.

Neither removing features from the best models nor adding features from other models improved performance (see Figure S12).

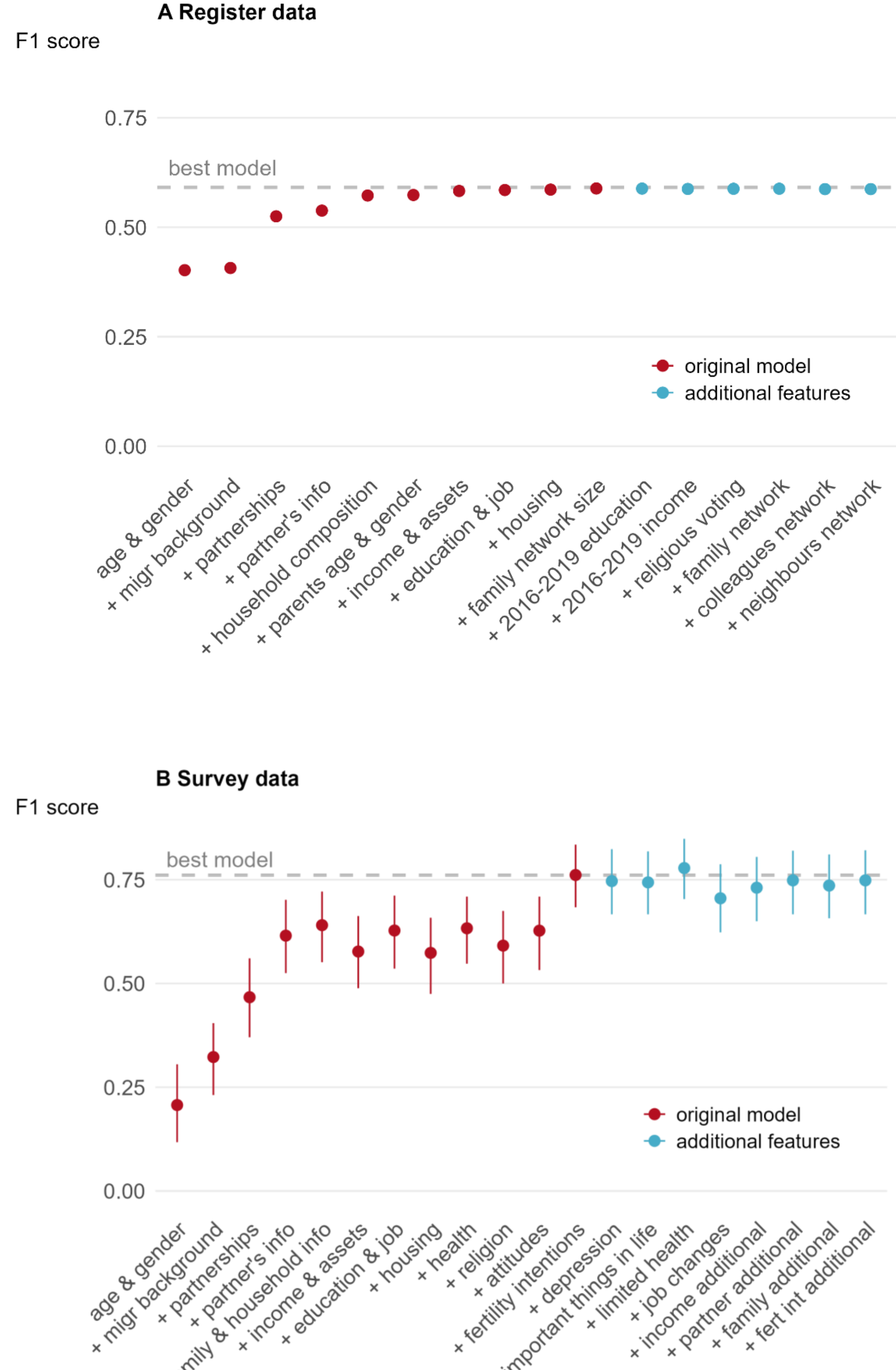

**Figure S12.** Combining variables from different submissions. Error bars show 95% confidence intervals. Error bars for the register are not visible because they are very narrow.

## 2.9 Learning curves

For the register data, we constructed the learning curve by drawing random subsamples of varying sizes from the training data and evaluating the retrained best model on the original holdout set, using 30 random subsamples per training size (except for training sizes 2M and 3M, where only one random sample was used). We retrained only the single best-performing model rather than all submitted models. The best register-based model was trained with default hyperparameters, which we also used when retraining with different training sample sizes. Figure S13 shows the learning curve with a log-scaled x-axis to better show performance and 95% confidence intervals at smaller training sample sizes.

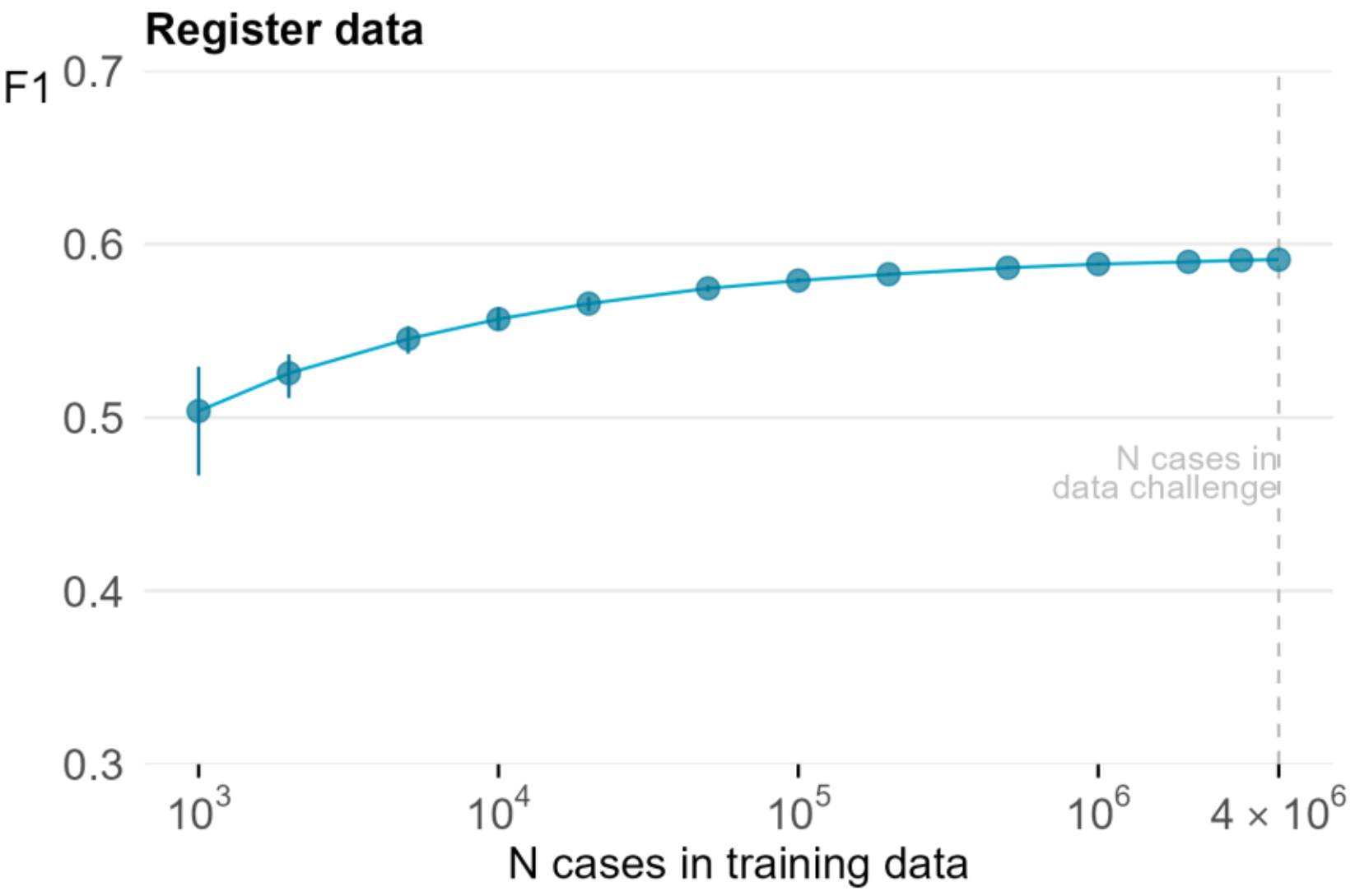

**Figure S13.** Learning curve for the best register-based model with a log-scaled x-axis. Same as Figure 4 (A) in the main text, but with a log-scaled x-axis to better show performance at smaller training sample sizes. Points represent mean F1 across 30 random subsamples at each training size (except for training sizes 2M and 3M, where only one random sample was used). Error bars show 95% confidence intervals based on the 2.5th and 97.5th percentiles of these 30 values. For 2M, 3M, and the full sample, confidence intervals are not shown.

For the survey data, we first expanded the training sample by linking survey responses to register data in the secure environment, retrieving outcomes for respondents who consented to linkage and for whom the outcome was missing in the survey data. This expanded the training sample from 987 to 1,688 cases (see details in SI section 2.5).

Note that the winning survey-based model further augmented the training set via a three-year time shift[89] (see details in SI section 3), which we exclude here, as most of those observations represent the same respondents measured at an earlier point rather than additional independent cases. As with the register data, we retrained only the single best-performing model, using 30 random subsamples per training size. Since the best survey-based model included hyperparameter tuning, we performed tuning once per training size and applied the resulting hyperparameters to all 30 subsamples at that size.

## 2.10 Combining survey and register data

As noted in the main text, adding predicted probabilities from the best register-based model as a feature in the best survey-based model raised the F1 score from 0.63 to 0.68. The baseline (F1 = 0.63) is the survey-based model trained on the updated training set (with outcomes retrieved from register data for those with missing outcomes) without time-shifted data (see details in SI section 3). We did not use the full training sample for this best survey-based model (i.e. with time-shifted data) because for the time-shifted data, predicted probabilities from the register-based model are not available: that would require retraining the register-based model on the time-shifted data as well. The enhanced model (F1 = 0.68) adds predicted probabilities from the best register-based model as an additional feature. These probabilities are available for 94% of our survey dataset (among those with available outcome).

The improvement could happen for several reasons: 1) patterns between variables (present in both survey and register data) were captured better by the register-based model due to a larger sample size of the register data, 2) because the register-based model includes features missing from the survey-based model; 3) because of the more accurate measurement of some variables in register data, 4) because of the error cancellation: when the two models are trained on different individuals, their prediction errors are partially independent, and averaging across them reduces overall error. We designed a series of experiments to isolate the contribution of each mechanism. We systematically varied three aspects of the register-based model: 1) whether it's trained on the same people as the survey-based model, 2) training sample size, and 3) feature set. For this analysis, we retrained the register-based model, excluding individuals in the survey holdout set to reduce the risk of data leakage.

First, we trained a register-based model using only overlap features (features present in both survey and register data) on the same individuals as in the survey training sample (for whom register data can be linked, N = 1,598). Second, we trained register-based models on subsamples of varying sizes (from 1,600 to the full training set; we trained 20 models for each training size), again keeping only features overlapping with the survey. Each time, we added the predictions to the best survey-based model.

Adding predictions from a register model trained on the same individuals as the survey training set reduced performance to F1 = 0.60 (Figure S14). Adding predictions from a register model trained on a different random subsample of comparable size (N = 1,600, drawn from outside the survey training set) yielded F1 = 0.63 (on average across 20 subsamples) – no improvement compared to the baseline (model without added predictions). Larger subsamples (N = 50,000 and N = 500,000) similarly showed no improvement. Only when using the full register training sample (~4M observations, excluding the survey holdout) did performance improve to F1 = 0.65. The final gain to F1 = 0.68 came from adding predictions from a register model trained on the full sample with the complete feature set (rather than only features overlapping with the survey). These results indicate that the performance improvement stems from two sources: (1) additional register variables present in the best register-based model but absent in the best survey-based model, and (2) very large sample size enabling more accurate learning of patterns present in both datasets.

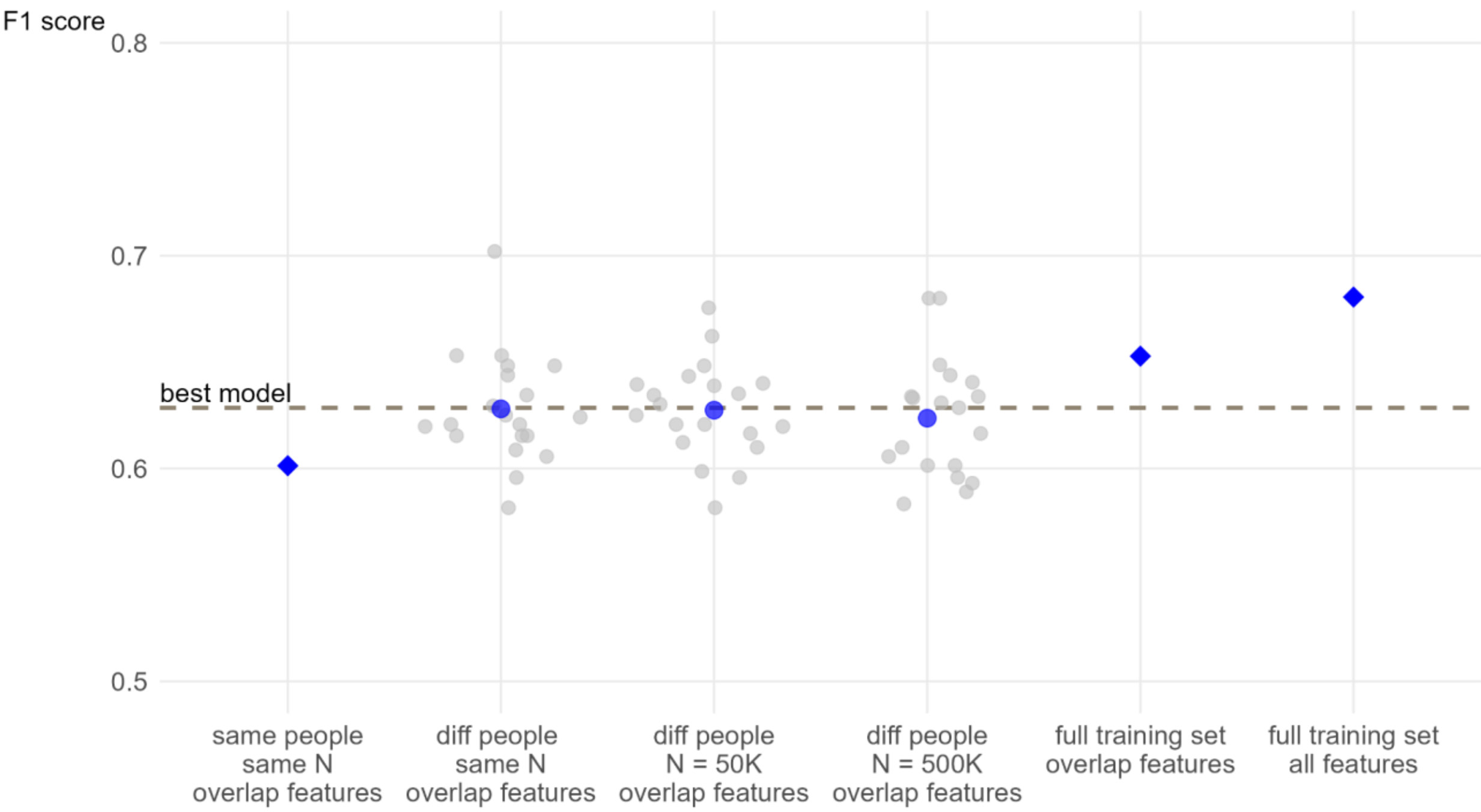

**Figure S14.** Decomposing sources of performance improvement from adding register-based model predictions to the survey-based model. Each dot represents the F1 score of the survey-based model augmented with predictions from a register-based model trained on a specific configuration. Multiple dots at the same sample size reflect models trained on different random subsamples (N = 20 per size). Dots are shown separately for models trained on overlap features only versus the full feature set, and for models trained on the same individuals as the survey training sample versus different individuals. A horizontal dashed line indicates the baseline F1 score of the survey-based model without added register predictions.

## 2.11 Potential data leakage

There is a potential source of data leakage in the register data arising from Statistics Netherlands' methodology for identifying unregistered cohabiting partners (see details about the methodology in the documentation of the Samenwonerbus dataset[91]). Pairs who had a child together during the outcome period (2021–2023) are systematically more likely to be identified as unregistered cohabiting partners in the data from previous years. This means that household type in 2020 may have been partly inferred from future births for a small subset of the holdout set – introducing a leakage of outcome-related information into the 2020 predictors. This could affect models that used household type as a predictor (9 out of 11), including one team that used it to merge in additional partner-level information.

However, the leakage is likely minimal for two reasons. First, the version of the household dataset that included the variable about household type (Gbahuishoudensbus) used in PreFer was created at the beginning of the outcome period (in April 2021), meaning only people who had a child in the first few months of 2021 could be affected. Of those, only a small fraction involved unmarried cohabiting partners rather than married couples or single parents. In total, 7,699 holdout observations fall into this potentially affected group, representing 0.6% of all holdout observations and 4% of all observations with a positive outcome in the holdout set.

Second, we tested whether performance drops when this group (individuals who had a child between January and April 2021 and were in an unmarried partnership) is excluded from the holdout set. For the best-performing model, the drop is small (F1: 0.59 → 0.58). For most other models that used household type, the drop was similarly small except for two models (0.50 → 0.36 and 0.57 → 0.54), see Table S7. For one model, performance slightly increased. The two models that did not use household type showed no decrease in performance.

However, interpreting these drops is not straightforward. Partnered individuals might be more predictable because they have more stable household characteristics and partner features available in the 2020 data. So, excluding a part of them may reduce scores for reasons unrelated to leakage. The fact that the two "clean" models show no change does not rule this out, as those models may simply perform worse on the partnered subgroup for independent reasons. It remains unclear whether the observed performance drops reflect genuine leakage or the removal of an inherently more predictable subgroup. Most importantly, where drops do occur, they are small for the majority of models. We therefore conclude that this potential leakage did not substantially affect our results.

**Table S7.** Model performance on the full holdout set and the holdout set excluding observations potentially affected by unregistered partnership data leakage

| Team | Score on the full holdout set | Score on the holdout set, excluding cases potentially affected by a data leakage |
|---|---|---|
| Stork Oracle | 0.59 | 0.58 |
| Cruijff | 0.57 | 0.54 |
| SBU EUI | 0.57 | 0.56 |
| teamSGBD | 0.57 | 0.56 |
| ConceptionCoders | 0.55 | 0.54 |
| DEPLearning | 0.52 | 0.51 |
| JustThe2OfUs | 0.52 | 0.51 |
| Cruijff & Stork Oracle | 0.50 | 0.36 |

| Coding babies | 0.38 | 0.39 |
|---|---|---|
| *Models that did not use the household type variable:* | | |
| data storks | 0.47 | 0.47 |
| PhDecline | 0.44 | 0.44 |

## 3. Submissions: scores and brief description

In the table S8, we present several performance metrics for the best register- and survey-based models in addition to F1 scores presented in the main text. A table with all the scores for all submissions is provided in the replication package. $R^2$ is defined as in SI, section 2.3.

**Table S8.** Performance metrics for the best register- and survey-based models

| Data source | Team | F1 | Precision | Recall | MCC | $R^2$ | Log loss | Accuracy | AUC |
|---|---|---|---|---|---|---|---|---|---|
| register | Stork Oracle | 0.59 | 0.55 | 0.64 | 0.52 | 0.35 | 0.27 | 0.87 | 0.89 |
| survey | Stork Oracle | 0.76 | 0.91 | 0.66 | 0.72 | 0.53 | 0.29 | 0.91 | 0.89 |

Tables S9 and S10 provide descriptions of all valid submissions, defined as those that ran successfully on the holdout data (i.e., produced scores) and were not labelled as test submissions by the team. Table S9 (register-based models): teams did not provide descriptions; we wrote descriptions ourselves based on the code and papers describing these approaches. Submissions are sorted by F1 score.

**Table S9.** Descriptions of the register-based submissions

| Team | F1 | Description |
|---|---|---|
| Stork Oracle | 0.59 | CatBoost trained on about 100 variables, selected manually based on ideas about what would be important for predicting fertility.<br><br>Model: CatBoost model (selected for effective at handling categorical features, which are prevalent in the register data). Default |
| | | hyperparameters were used after an exploration of tuning options, where no meaningful performance gains were observed.<br><br>Features: PreFer Base file is used, supplemented by several additional features from the register datasets. These additional features cover family network data and data about cohabiting partners, including partners who are not married or registered.<br><br>Data preparation: minimal preprocessing; relying on CatBoost's built-in capacity to detect patterns without extensive preprocessing and feature engineering. No missing value imputation, no one-hot encoding, and no variable transformations as none of them are required by CatBoost. Predicted probabilities were converted to binary labels using a threshold tuned to maximise F1. See details about this approach in [55]. |
| Cruijff | 0.57 | An open-weight large language model fine-tuned on "books of life": natural-language life histories constructed from the tabular register data, which mirror traditional biographical accounts. The core idea is to leverage an LLM to identify patterns across millions of individual lives while preserving the temporal, network, and multi-level complexity of the register data, which might be lost to a large extent when life courses are reduced to feature sets. A publicly available pre-trained LLM (Llama 3.1 8B-Instruct) was fine-tuned rather than trained from scratch, in order to build on a model that had already acquired contextual information about social life from internet-scale data.<br><br>The submitted model used "enriched" books, which combined register-based life histories (demographic information, the most recent household spell, and the last ten residential spells) with the predicted probability from a gradient-boosted decision tree trained on the PreFer Base dataset, included in the text as a single short phrase. This was done to increase the density of information per token under the challenge's compute constraints. The team reports that the performance of this submission is attributable primarily to the embedded gradient-boosted tree rather than to what the language model added on top of it, and that a version using books without the embedded prediction performed worse on their internal evaluation set. Books were restricted to approximately 1,000 tokens (roughly three pages of text) due to memory and processing time limitations. Predicted probabilities were converted to binary labels using a threshold tuned to maximise F1. See details about this approach in [49,50]. |
| SBU EUI | 0.57 | A transformer-based foundation model pre-trained from scratch on a large number of life sequences and fine-tuned for fertility prediction (approach similar to [20]).<br><br>Tabular registry data is represented as text-like sequences written in symbolic language; each individual life course is a "sentence" where individual life events and their attributes are represented by units similar to words (tokens) and ordered chronologically. This representation is designed to leverage the power of transformer-based models, originally developed for NLP.<br><br>The model has two components. The foundation model consists of stacked transformer layers and is trained to capture complex interdependencies, timing effects, and path dependencies within life-course sequences, analogous to how language models learn grammar and semantics. The 8M-parameter foundation model is pre-trained from scratch on approximately 19 million life sequences, learning general patterns of how life events relate in terms of categories, co-occurrences, and timing. Life sequences are constructed using data on all available domains – demographics, education, household, income, employment, partnerships, divorce – for the focal individual as well as, where applicable, their parents and partner; sequences also include counts of network ties to family, household members, neighbours, schoolmates, and colleagues.<br><br>The second component is a classifier head placed on top of the foundation model that outputs probabilities of having a child for each individual. During fine-tuning, both the classifier head and the foundation model are trained jointly on the downstream task, with loss backpropagated through the full model. This allows the model to leverage its general knowledge of life-course patterns while refining its internal representations toward signals most predictive of fertility outcomes. See details about this approach in [48]. |
| teamSGBD | 0.57 | LightGBM-based approach with a focus on social network data.<br><br>Several hundred additional variables were constructed that captured the same information as in the PreFer Base file, but for 3 different personal networks. Neighbour network: a random sample of 20 people living within 200 meters of an individual's address. Family network: familial relations typically of the same age as the individual, specifically full siblings, half-siblings, step-siblings, and cousins. Colleague network: up to 100 colleagues who lived closest to the individual; to ensure relevance, these results were filtered to exclude any colleagues older than 45. PreFer Base file data was transformed |
| | | into features for these networks by calculating the average feature value of the people within each network. For categorical features, this resulted in a fraction representing the proportion of people in that network belonging to a specific category. Additionally, income trends and educational changes were created and used as features.<br><br>Predicted probabilities were converted to binary labels using a threshold tuned to maximise F1. See details about this approach in [92]. |
| Conception Coders | 0.55 | Ensemble voting classifier with random forest, logistic regression, HistGradientBoosting, AdaBoost, and KNN. Variables from the PreFer Base file and several additional variables are used: demographics (age, gender, migration background), parents' migration background, household type, number of children and age of the youngest child, partnerships, socioeconomic category, employment characteristics, longitudinal income, assets, house ownership, education, household characteristics, partners' education and employment, childcare proximity, municipality-level data about voting (to infer political values/religiosity); number of siblings, parents, and children of siblings. Missing values were replaced with a new category. Oversampling of the minority class (those who had a child in 2021-2023) to match the majority class size. |
| DEPLearning | 0.52 | XGBoost with hyperparameter tuning. Variables: demographics (age, gender, migration background), partnerships, household type, income, assets, employment characteristics, socioeconomic category, education, municipal election results, municipality-level statistics (such as population density, proportion of residents aged 15-44); network features – number of siblings, siblings' children under 3 years old, and their total number of children, colleagues' children under 3 years old. Missing values are handled using KNN imputation (numerical columns) and a new category for categorical and ordinal columns. Oversampling of the minority class (those who had a child in 2021-2023) to match the majority class size. |
| JustThe2OfUs | 0.52 | Random forest with tuned hyperparameters. Variables: demographics (gender, age, migration background), household type, number of children and age of the youngest child, years spent in household type with/without children, number of persons in household, assets, partnerships, employment, income, education, partners' age and gender and country of birth, municipality-level data about voting; network features – number of children of siblings under 4/12/18 years, sibling count, parental age at first birth. Missing values are handled |
| | | differently depending on the variable; in some numeric variables, by creating a new category; in partner variables, by creating new categories, representing either no partner for not married/partnered, or another category if information is genuinely missing. Class imbalance is handled via 'class_weight' parameter in random forest. |
| Cruijff & Stork Oracle | 0.50 | Predicted probabilities generated by the Stork Oracle's model are added to the textual "books of life" (see the Cruijff model above). |
| data storks | 0.47 | Random forest. Variables: demographics, household type, financial assets, house ownership, education, number of children and age of the youngest child, partnerships, partner age and gender and employment, municipal elections data. Missing data imputation with either mode (categorical) or mean (for numeric). Oversampling the positive class to handle class imbalance. |
| PhDecline | 0.44 | Gradient boosting classifier. Variables: demographics, parents' migration background, income, number of children and age of the youngest child, partnerships, employment, partner age and gender and employment. Mean imputation of missing values. |
| Coding babies | 0.38 | Penalised logistic regression with hyperparameter tuning. Variables: demographics, household type, education, economic and financial independence, employment characteristics, income; network features – how many children under 3 years old and under 12 years old are present among persons' siblings, colleagues, colleagues of siblings, siblings of colleagues. KNN imputation for missing values in numerical features, creating a new category for missing values in categorical features. |

Table S10 (survey-based): submissions are sorted by team ranking (best-performing team first) and by round within each team. Descriptions are based on participants' original text, shortened for space. Original descriptions are available in the replication package (in the folders with original repositories).

In some cases, teams described their overall approach or how it evolved across rounds in the final submission rather than describing each round separately. We provide these descriptions for the round where they were submitted and did not split them across the rows of the table. If no description was provided for a submission, we reconstructed the approach from the submitted code and description of the overall approach (if available). All reconstructed descriptions or additions to the original descriptions made by us are in italics. Where an earlier-round description largely duplicates a later one, we retain only the most recent full description and summarise only the differences for the earlier round.

**Table S10.** Descriptions of the survey-based submissions.

| Team | Round | F1 | Description |
|---|---|---|---|
| Stork Oracle | 5 | 0.76 | XGBoost with the following strategies: (1) Tripled training sample size with "time-shifted" data, (2) Partner matching and merging in data from the partner's survey for households where both partners participated, (3) Combined data from related features into "scales". Details:<br>(1) We calculated whether suitably aged people in the training and supplementary data had children between 2018 and 2020, thus creating additional outcome data. For these rows of additional outcome data, we recoded features from year t-minus-1, year t-minus-2, etc., to have the same name as the equivalent features at year t-minus-1, year t-minus-2, etc. in the original data. For example, in a time-shifted row, cf17j128 is renamed as cf20m128 in order to correspond with data 3 years later. To help account for the temporal distribution shift, we include a feature that is an indicator of whether the row comes from the time-shifted data or the original data.<br>(2) We merge the partner's fertility intentions, number of children, age, income, year of birth of the most recent child, and gynaecologist visit.<br><br>(3) We generate "scales" in the feature data by averaging related features together. Our scales are: Feelings toward current child, gendered religiosity, attitudes about traditional fertility, attitudes about traditional motherhood, attitudes about traditional fatherhood, attitudes about traditional marriage, attitudes toward working mothers, and sexism.<br><br>We choose the hyperparameters for our XGBoost model via grid-search hyperparameter tuning with 5-fold cross-validation.<br><br>*In addition to scales, variables about a person include Handpicked variables about demographics, partnerships, family and household information, income and assets, education and job, housing health, religion, attitudes (scales), and fertility intentions. See [89] for further details about this overall approach.* |
| Stork Oracle | 3 | 0.72 | *Same approach as Round 5, but the year the most recent child was born is not included, and fewer linked partner variables are used (only fertility intentions and number of children). Some variables are used, which are later dropped. There are also differences in the hyperparameter tuning, e.g. less tree counts, fewer deeper tree options, and less focus on lower learning rates.* |
| Stork Oracle | 2 | 0.70 | XGBoost model with 66 hand-picked variables, which are converted to 43 predictors. |

| | | | *Same approach as Round 3, but there is no linked partner data, and some other variables are not included (e.g. fertility intentions from years before 2020, satisfaction with relationships/family life).* |
|---|---|---|---|
| Stork Oracle | 1 | 0.68 | XGBoost with 66 hand-picked variables, which are converted to 36 predictors. *Same approach as Round 2, but there is no time-shifted data, and the preprocessing of some variables is different. There are also differences in the hyperparameter tuning, e.g., a larger number of trees are tried, and there are no slow learning rates.* |
| LYC-MRK | 5 | 0.67 | Xgboost with feature selection based on literature review (325 features originally) + dimensionality reduction based on feature importance from the Random Forest Classifier. During all submissions, the reduction of features showed an improvement in our performance. Therefore, feature reduction was applied in all five submissions, from around 200-300 features in the first two submissions to 180 and then around 70 in the last two submissions (*mainly by dropping measurements of the variables from 2008-2017, not by dropping topics*). Variables included in the last submission, most from 2018–2020: demographics, fertility intentions, partnerships, partners' age and gender, number of children, household composition, satisfaction with relationships/family life, division of household labour, income, education, housing. We utilised the synthetic minority over-sampling technique (SMOTE) method to rebalance the original training set. Missing values have been substituted with the numerical value of ’0’. |
| LYC-MRK | 4 | 0.73 | *Same approach as above, but variables from earlier years are added about fertility intentions and family. Some demographic background features are dropped. A simpler model configuration is used (fewer trees and smaller tree depth).* |
| LYC-MRK | 3 | 0.65 | *Similar approach as above, but random forest with default hyperparameters is used instead of Xgboost; more variables from earlier years are included (but same broad topics).* |
| LYC-MRK | 2 | 0.65 | *Similar approach as above, but XGBoost is used, with minimal hyperparameters; more variables from earlier years are included (but same broad topics).* |
| LYC-MRK | 1 | 0.67 | *Similar to above, but a slightly different set of features from the same topics.* |
| PASPR EFER | 3 | 0.68 | Our final submission is based on an ensemble method using various Random Forest classifiers grown on different sets of variables. The variable selection is a combination of theory- and data-driven considerations. We started with a review of the literature on the (standard and newer) correlates of fertility |
| | | | behaviours. We then carefully screened Prefer codebooks to identify these relevant variables. In a few cases, we were not able to find relevant information (e.g., on fecundity issues, contraceptive behaviours). This preliminary theory-driven selection of variables was further refined using an additional Lasso regression to identify a broader set of useful predictors and confirm the necessity of including the previously identified variables. Final selection includes fertility intentions, physical and mental health, demographics, satisfaction with achievements in life, education, partnerships, satisfaction with relationships, number of children, income, traditional values index, division of caregiving responsibilities, employment, gynaecologist visits, political attitudes, and social media use. In several cases, we combined information from several variables (e.g. fertility intention variables or items on mental well-being/depression) into scales. In the previous submission, only a few theory-based variables were used<br>In the case of missing values, we have considered forced imputation from past values of the same variables only if this made sense (e.g. for (almost) time-invariant factors). For the remaining missing values, we have created specific categories. We did not consider multiple imputation or alternative approaches due to their complexities within the context of a data challenge. In addition, in some cases, we noted that the missing category was predictive of the outcome. Two separate Random Forest workflows are tuned independently; one uses only the three fertility intention variables, the other uses all remaining predictors except intentions<br>Finally, to address class imbalance, we tuned the classification threshold on the validation set in order to yield the best value in terms of F1 score. |
| PASPREFER | 2 | 0.72 | Random Forest. Predictors: age, gender, fertility_intention, health, education, partnership, N_children, income, occupation, migration_bg. |
| PASPREFER | 1 | 0.52 | Logistic regression with age, gender and intention as predictors. |
| yann1cks | 5 | 0.72 | The submission uses an ensemble of Gradient Boosting Machines (LightGBM, XGBoost and Sklearn's Histogram-Based-Boosting). The classifiers are only trained on the individuals with an available outcome variable. Variables are selected based the Feature Importance of simple LightGBM and XGBoost models trained repeatedly on a subset of the data. The household variable from the background dataset is used to conduct grouped Train-Test-Splits or Cross-Validation to avoid data leakage per household. Moreover, I tried to preprocess the features based on the definitions in the codebook and certain heuristics, i.e. all personality variables are defined as continuous variables, missing value indicators are removed, years are reformatted to ages (or time-differences), and categorical variables are defined as the respective pandas dtypes. During preprocessing, I also removed the free-form answers. The feature set used is very large (over 1000 variables), because I have seen minor |

| | | | improvements in prediction quality, but I lacked the time to identify the relevant variables. My goal was to take a data-driven approach to feature selection to identify currently unknown correlates of fertility, but inspecting the selected variables, this was not successful. I cannot rule out overfitting, so the large number of variables most likely degrades the performance on the holdout data. The training and hyperparameter optimisation were done with the Microsoft FLAML library. This library offers so-called flamlized versions of common Machine Learning Classifiers (e.g. LightGBM), which enable zero-shot Hyperparameter Tuning. These Hyperparameters are selected based on characteristics of the dataset, so no expensive Optimization is needed while iterating on the ideal model. Moreover, the library offers an easy-to-use way to optimize Hyperparameters utilizing the information described above.<br>I tried using Semi-Supervised with SelfTraining or TriTraining (with Disagreement) to utilise the large amounts of missing data, but was not able so reach a better F1-score. Moreover, I tried time-shifting the data, which works better, but it contradicts my goal of a fully data-driven approach. |
|---|---|---|---|
| Team Erazmus | 2 | 0.72 | Our submission used a gradient boosted random forest for classification. As input, we used a subset of the 200 best-performing features. These features were identified by calculating the information gain of each one when compared against a proxy gold standard (the answer to the survey question "In how many years do you plan to have your next child?"). This question (naturally) showed a very high correlation with the target variable, so we concluded that features which perform well in predicting it should also perform well on the PreFer challenge. |
| chrisrellama | 5 | 0.48 | *Three gradient boosting classifiers are combined via a soft voting ensemble, averaging their predicted probabilities. Using XGBoost for feature importance selection.*<br>● plotted some features against 'age', noticed linearity, used linear regression, then used the intercept and 'age' to fill NaNs<br>● dropped some features that scored high in variance inflation factor<br>● found that the ensemble of XGBoost and LightGBM scored best in cross-validation<br>● utilised Optuna and StratifiedKFold for efficient hyperparameter optimisation |
| chrisrellama | 4 | 0.43 | *Single XGBClassifier, same approach to feature selection based on XGBoost feature importance.*<br>● drop some features based on variance inflation factor |
| | | | ● found that using XGBoost only gives the highest score in the train and validation data<br>● selected some features to use mean imputation *(for others, a mode was used*)<br>● isolated the rows with missing targets, fitted XGBoost to rows with complete targets, used the model to predict the missing targets, and then utilized the entire dataset for Optuna |
| chrisre llama | 3 | 0.71 | *Compared to the previous approach, more variables are included (no VIF-based column drops). A soft-voting ensemble of XGBoost, CatBoost, and LightGBM. Minimal preprocessing (no scaling).*<br>● adopted mode imputation to handle missing values across all data types<br>● expanded the model ensemble by incorporating CatBoost and LightGBM alongside XGBoost<br>● utilised Optuna for efficient hyperparameter optimisation<br>● leveraged VotingClassifier to combine the predictions from the three models (XGBoost, CatBoost, LightGBM) using soft voting (majority rule based on predicted probabilities) |
| chrisre llama | 2 | 0.69 | *Similar to the above, but mode imputation is used for categorical features and mean imputation for numerical features* |
| teamS GBD | 3 | 0.71 | Modelling choices:<br>We worked with a 50-50 training and validation split to get a sense of which models and features were working well and which were not. The splits were randomly generated at every iteration, allowing us to monitor variation in performance due to different training data. The splits were stratified on the outcome variable and made so that members of the same household would occur in the same split.<br><br>LGBM offers a fast and accurate model for classification, which is able to handle missing data in tree splitting. We tested other models, including Logistic Regression, Random Forests, Naive Bayes, and converged on LGBM for the final submission due to a robustly better performance.<br><br>We tweaked the decision boundary to accommodate the highly imbalanced problem and found that predicting someone will have a child if the model assigns a probability higher than 28%.<br><br>Missing data imputation: |

|  |  |  | We considered running MICE (while keeping track of which responses were not provided, after observing that the mere presence/absence of a response on our set of features was rather predictive, suggesting that non-responses were informative), but decided against it because it was not clear whether imputation on the test set could have relied on information in the training set. Moreover, since LGBM handles missing data without the need to impute, we decided against MI.<br><br>Feature selection:<br>We started from the full set of features in background and core studies, adopting a full bottom-up approach. For every individual, we added the last available value of the background characteristics and dropped variables with textual, date or time response types.<br><br>We then inspected confusion matrices and feature importance to gauge which waves and features provided the most information. This process highlighted the need to fine-tune the decision boundary to avoid a high number of false negatives. Moreover, we observed that important features consistently concerned waves from 2017 on. We thus discarded all previous waves to have a more manageable set of features. Feature importance confirmed that the model relied on features related to fertility, such as questions asking about fertility intentions.<br><br>Feature engineering:<br>From the variables that were consistently found important for different data splits, we derived new features. First, to reduce the missingness rate, we combined the most recent non-missing answers from the waves to questions relating to fertility intentions (cfxxx029, cfxxx128, and cfxxx130), as well as those about key family events, such as getting married (cfxxx031) and becoming parents (cfxxx456).<br><br>The background features were consistently considered important as well. Due to the low missingness of these variables, the prior strategy was unnecessary. Instead, we recorded the number of unique values across the years for variables related to housing situation (woning, sted, woonvorm, partner), education (oplzon, oplmet), and the variables burgstat and belbezig. The number of unique values over the years for a person can indicate changes in these variables. Another change in response values that we extracted was that of net household income (nettohh_f). For this variable, we calculated the slope for each individual by fitting a linear regression to the yearly observations.<br><br>Observations: |
|---|---|---|---|

| | | | We did observe that fertility intentions do predict fertility outcomes. Moreover, we observed that household income mattered more than individual income. Marital status also consistently showed up as an important feature. |
|---|---|---|---|
| teamS GBD | 2 | 0.42 | *LGBM with variables from the Background LISS survey. The most recent wave available for a person is used. No feature engineering; missing values are filled with a new category. The decision boundary is tuned (as in the final model).* |
| teamS GBD | 1 | 0.30 | *LightGBM trained on transformed data: all core studies (from all years, 2008-2020) are reshaped from wide to long format (one row per respondent per wave). Background data is merged per wave. The original outcome from 2021-2023 is used. At prediction time, the model produces one probability per wave per person, then aggregates these across waves into a single final prediction.* |
| Copen hagen Social Compl exity Lab | 3 | 0.39 | We proceed in three steps:<br>1. Gradient boosting algorithm (xgboost):<br>○ To establish a strong baseline.<br>○ To evaluate the predictive power of each question. Disadvantage: doesn't scale as well with data<br>2. Our best performing model is the ExcelFormer [4] + RNN:<br>○ (ExcelFormer) To create a deep representation of the (answered) questionnaires<br>○ (Recurrent Neural Network) To leverage temporal interactions between the questionnaires<br>3. Our work initially focused on the AutoEncoder model:<br>○ To capture complex, non-linear relationships in the data.<br>○ To explore new factors and interactions that traditional gradient boosting models might miss.<br>The last approaches allow us to model the data more similar to text (i.e. sequential data), enabling pretraining like seen in fields as NLP and CV.<br><br>For Xgboost, the most important parameters for the prediction task are: scale_pos_weight to account for the unbalanced outcome and reg_lambda for L2 regularization to avoid overfitting. Xgboost serves as a robust baseline with an F1-score = 0.71 on the hidden validation set of round 2. The most important features/questions are displayed in Fig. 1 [*see original repository in the replication package*]. The main characteristics of these questions are: cf code, indicating Family & Household type of questions; 2020 - most of the important questions are from the last year of the survey. The most important question is “Within how many years do you hope to have your [first/next] child?” (in 2020). This question remains significant across different survey years (Fig. 1). xgboost requires only the top ~200 questions to achieve the best predictive performance. Performance plateaus and even declines when |

additional features are included (Fig. 2) [*see original repository in the replication package*].

Final Model:
Our best performing model reuses the idea from the AutoEncoder pipeline. Here, we use ExcelFormer model to create dense representation of input, i.e. answers to a questionnaire, followed by the RNN model that captures temporal interactions between questionnaires of different years. Finally, we use a simple attention mechanism to aggregate the output of the RNN and create a person-embedding. We use these person embeddings to make the final predictions.

ExcelFormer
The *ExcelFormer* is a transformer-based model designed for the *tabular* data (in our work, we redesign certain aspects of the pipeline):

1. Each column (question) and possible answer (category) is embedded into a high-dimensional space - we have a separate embedding space for answers and columns (see Data Processing for more details),
2. *ExcelFormer* takes the sequence of the embedded questions-answers (that corresponds to a survey from a specific person p for a specific year y) and passes it throught encoder layers.
3. The *ExcelFormer* is trained directly on the downstream task.

Prior to passing questionnaire-sequence to the model, one need to sort the columns based on the importance to the target value (i.e. fertility). We take these importance scores from our experiments with the xgboost model.
The main reason is the mechanism behind the *ExcelFormer* self-attention: the more important columns are not allowed to incorporate information from the less important columns (while the less important columns can do so). We adapted the implementation presented in the PyTorch Frame package.
For each person p, we embed the corresponding surveys s for each year. If a person did not answer a survey for a specific year, we assume the embedding of the survey is a 0 vector.

RNN model (with attention mechanism)
A set of embedded surveys Sp=s0,s1,s2,..s13 of a person p is then passed to a RNN model in a chronological order. RNN returns a contextualized representation of a survey s at a given timepoint - hence RNN still returns a set of surveys.
We aggregate these contextualized representations, $h_t$, using a simple attention mechanism:

$$a_t = \frac{e^{h_t \cdot c}}{\sum e^{h_j \cdot c}}$$

Here, c is a learnable context vector. The final person embedding is

$$ \widehat{\mathbf{h}} = \sum a_t \cdot \mathbf{h}_t $$

The two-year bidirectional GRU [5] provided the best results.
To reduce overfitting on limited labelled data, we apply Exponential Moving Average weight averaging starting from epoch 3.

Performance:
Based on our bootstrap estimates (on the holdout dataset), the performance on the real unseen data is (with 95% confidence intervals):

1. F1-Score: 0.771 [0.667, 0.861]
2. MCC : 0.720 [0.600, 0.828]
3. Precision: 0.895 [0.786, 0.976]
4. Recall: 0.679 [0.545, 0.808]
5. mean Average Precision: 0.872 [0.789, 0.937]

After estimating the metric using our data splits, we retrained the model on the full dataset available.
The full training pipeline is in the train_finetune.ipynb notebook.

Data Processing:

We do a series of data processing steps to convert the tabular data into sequential data, to represent the temporal aspect of the surveys. We create a sequence for each year and tokenize each sequence. We decided to work only with categorical, numeric and date columns, as they are easy to process and represent most of the data. There also exists a file to handle free-text in data_processing/text2vec, but this requires an external model (LLM) to create the embeddings, which is why we have chosen not to include it.

Embedding and tokenisation:
In order for the deep learning models to understand the data, we must embed it. This requires us to put an integer value to all questions and answers.
For questions, it is easy: We just group questions asked in multiple surveys and assign each group an integer.
For the answers, we do this through tokenisation, where we convert each answer to a token.
For categorical (data_processing/categorical.py) answers, we assign each unique answer (accounting for the rephrasing) an integer. This creates the same mapping for an answer (e.g. "yes") that answers two different questions. This allows us to share the vocabulary, decreasing model size and increasing the semantic meaning of each answer.
For numeric or date or time (data_processing/numeric_and_date.py) we bin the values into percentiles (100 bins), that is again shared between questions to increase the semantic meaning.

| | | | |
|---|---|---|---|
| | | | If an individual has not answered a given question, we assign their answer to an unknown token [UNK].<br><br>Sequential input:<br>As our models require sequential data (so they better fit registry data in phase 2), we must convert the tabular format into sequences. We create a sequence for each year to reflect the temporal aspect of the data. The order of the sequence for each year is kept constant across years, as we have no predefined order. This is done using the codebook, where we allocate each question (e.g. cfxxx003) a specific spot in the sequence. This means the question "Gender respondent" will always be at the same index of the sequence across all years. For registry data, the sequence would be ordered in the order in which they appear throughout an individual's life.<br>If a question has not been asked for a given year, we still keep its spot in the sequence and assign an unknown token [UNK] as the answer for all individuals.<br>For the ExcelFormer, we order the sequence based on the feature importance from the XGBoost.<br>We also allow for subsetting the number of columns through the custom_pairs in data_processing/pipeline.py. |
| Copenhagen Social Complexity Lab | 2 | 0.71 | *Xgboost trained on all features, excluding string and datetime variables. Class imbalance is handled via scale_pos_weight, computed as the ratio of negative to positive cases in the training data. L2 and L1 regularisation to reduce overfitting, combined with row subsampling.* |
| GAS | 3 | 0.67 | [*See plots and tables included in the original description in the replication package or here [https://github.com/apiraccini/prefer_gas_official/blob/09f8d759871062b5e1f45d58830a3b2a9cddc12b/description.md](https://github.com/apiraccini/prefer_gas_official/blob/09f8d759871062b5e1f45d58830a3b2a9cddc12b/description.md)*]<br><br>Approaching this prediction task, the first thing that came to our mind was this basic principle: the panel wasn't conceived with the specific intent of predicting individual fertility. This led to three main hypotheses regarding the features.<br><br>• It is very likely that for predicting fertility during 2021-2023, the most important information will come from surveys conducted closer to 2021, hence it might be beneficial to primarily use the information from surveys conducted in recent years. |

- The most important information will probably lie in a few features; Therefore, it will be crucial to understand which features to select to best utilize the information in the important variables.
- We also expect that handling missing values will be relevant to our task: it's possible that the missing values are not caused by a purely random component but may, in some cases, contain information relevant to the study, e.g. due to the branching nature of questions during the survey completion process. Furthermore, considering that the data comes from a longitudinal survey with yearly surveys, it is possible that missing information for some variables in a given year can be correctly and consistently obtained from previous years (if it makes sense to do so).

Data exploration:
We removed the rows where the outcome is not available and the columns which had all values missing to obtain an initial train set with 987 rows and 25,868 columns.

In addition to the primary dataset, we also utilized the background dataset, which is structured in a longitudinal format. To align it with the main dataset, we filtered subjects with available outcomes, following the same approach used with the primary dataset. It's relevant to note that the background dataset contains information that we logically assumed to be the most important for predicting the likelihood of having a child (e.g. age, income, civil status).

We decided to focus on the train set first and only afterwards on the background survey. Our initial aim was to craft a procedure which could test our initial hypothesis. The idea was to develop a quick and automatic way to assess with sustainable precision the marginal predictive power of every feature available. We decided to evaluate the predictive performance on the task of predicting fertility of a univariate model, using a stratified cross-validation with 5 folds and iterating over every feature available. Our model chosen for this procedure was the decision tree, for several reasons: it can gracefully handle missing data, it's better suited with categorical/ordinal features stored as numbers that a linear method like logistic regression, and, last but not least, the optimized implementations available (together with the modest number of rows) allowed us to iterate across tens of thousands of features fairly quickly. The first plot shows the values of the F1 metric for each feature available, ordered by year of the survey. The performance metric decreases over time, thereby validating our initial hypothesis regarding the growing importance of surveys from recent years.
Encouraged by this confirmation, we further explored the variables from the last year, 2020, using the univariate tree method. As observed in the graph, except for the "Family & Household" survey, most surveys show only a few features with notable predictive power. This confirms our earlier assumption that amongst the many features available, only a few will be actually relevant for prediction. The most relevant variables seem to be those related to the desire to have a child, as well as income, age and civil status. It's interesting to note that the variable that indicates whether the person goes to the gynaecologist seems to be a good predictor of fertility: it might allow to identify pregnant women. Overall, these variables not only make logical sense but are also partially aligned with our initial assumptions.

Missing values:
Given that many features had a lot of missing values, we opted for a quick analysis and found out that, for the most important variables according to the procedure described above, the missing values present a correlation with the response. For example, for the variable "Within how many years do you hope to have your [first/next] child", it seems like subjects who don't respond are more likely not to have a child in the future. We developed an imputation strategy which is tailored to the longitudinal nature of survey variables. Beginning from 2020, for each variable containing missing values, we retrospectively examine earlier years for analogous variables asking the same question, imputing with the most recently observed value. Below is an illustrative example that demonstrates this method, displaying the resulting imputed values. To examine whether this approach was successful in improving the predictive power of our features, we conducted the tree selection procedure again on the newly imputed variables.
This approach enhances the predictive power of many features, as many variables with low scores in the previous tree selection plot appear to be relevant now. Despite that, we note that the variable with the highest score in the previous examination (cf20m130, "Within how many years do you hope to have your [first/next] child") has a much lower score now: this confirms our consideration that missing values contain relevant information and might actually be useful for prediction.

Background data:
We decided to leverage this background data to calculate aggregated measures over the recent years (2016-2020) for those key features that were repeatedly asked or recorded in both the training data and background data, but were found to be cleaner and more reliable in the latter.

In particular, in addition to extracting information on gender, age, family, domestic, work situation and current marital status, we focused on engineering new features that could summarise changes over time. Specifically, we created a flag to indicate if the subject got married during the 2016-2020 period and introduced indices to measure the median and standard deviation of income across those years. At the end of this process, we obtained a background dataset with 986 observations and 19 variables. We then applied the previously described tree methods to evaluate the predictive capacity of these variables. The results showed that the variables with good performance included those we created regarding income and marital status changes, as well as age.

Submission History:
The explorations reported above were part of an iterative process spanning the entire duration of the competition. These determined the preprocessing procedure utilized for crafting the final model, but were not taken into consideration for the early submissions. We provide a few details in the first two submissions before diving deep into the details of our final solution.

Submission 1

- Initially, we built a baseline logistic model using a few variables as suggested by the competition guides. We then upgraded to a more powerful classifier, a random forest, utilizing the same features, and observed an improvement in performance, likely due to the model's greater flexibility.
- Next, we focused on the background dataset and developed our preprocessing strategy, creating new features that moved us beyond the baseline according to our internal evaluation. To further enhance the predictive capabilities, we explored the use of a stochastic gradient boosting classifier (implemented using the CatBoost library).
- To incorporate information from the questionnaires not directly related to an individual's background, we investigated questions about children in general. By selecting a sample of these questions, we observed a significant improvement in internal predictive performance. This led to our first submission, which performed well on the test dataset, achieving an F1 score of 0.69 on the leaderboard.

Submission 2:

- For the second submission, we began by developing our tree selection procedure to understand the predictive power of each feature. By analyzing the results, we identified several relevant questions that could enhance our model's performance.
- Next, we compared different models using this new dataset, including combinations such as voting and stacking. Based on our internal evaluations, we chose a stacking combination of CatBoost and random forest for the second submission. This model achieved an F1 score of 0.70 on the leaderboard test dataset.

| | | | The results from the second submission were worse than anticipated. Up to that point, we had internally evaluated the models using a 5-fold stratified cross-validation, which showed to be a bit too optimistic. We then decided to use a stratified shuffle split procedure with a validation set proportion of 30% and ten repetitions, finding that the internal scores aligned more closely with those on the test set.<br><br>Final preprocessing:<br>Based on our exploratory analysis of the features' predictive power and handling of missing data, our final cleaned dataset comprises selected features from the year 2020, presented in two forms: raw and imputed. However, not all variables underwent imputation. For example, questions such as "Do you visit the gynecologist?" lose their contextual relevance when imputed, transforming into a generic "Have you ever visited the gynecologist?" rather than addressing current pregnancy-related visits. Therefore, for such questions, only the raw data from 2020 was used in the model.<br><br>Our technique for imputing missing values involves searching previous years for a response from the same subject to the specific question. For instance, if a user answered the question in 2017, that response was used to fill the missing value for 2020.<br><br>We introduced a slight variation in the imputation approach for the variable cf20m130, which asks "In how many years do you want to have your first/next child?". This feature is the strongest predictor in our analysis, and the standard imputation method would introduce bias: for example, if in 2017 the response to "In how many years do you want to have a child?" was 3 years, directly imputing this value for 2020 fails to account for the elapsed time, leading to inaccuracies. To address this, we subtracted k from a response given k years before 2020. For instance, if the answer in 2015 was "10 years," the imputed value for 2020 would be "5 years" (10 - (2020 - 2015)). This approach better reflects the passage of time, although it can result in negative numbers in some situations. We used two versions of this newly imputed feature: one retaining negative values and the other setting negative values to zero.<br><br>We also engineered new features using the data present in the training dataset. Specifically, utilizing "year of birth of the i-th child", we created additional variables related to respondents' family demographics, such as the number of children and the birth year of the youngest child. Additionally, we created the variable "same_sex", identifying same-sex relationships (of the 32 instances found, only one individual reported a child born between 2021 and 2023). |
|---|---|---|---|
| | | | The final group of variables introduced were derived from personality surveys. The questions from this survey were leveraged through factor analysis to derive scores for each individual across the five components of personality according to the Big Five model (refer to the appendix for a detailed explanation of the methodology followed).<br>Topics of included features: demographics, fertility intentions, relationships, socioeconomic status, family of origin and distance from parents, household and living situation, health, political views, attitudes and values, religiosity, personality [*see full list in the original repository*]<br><br>Modeling:<br>After finalizing our preprocessed training set, we compared several modeling approaches to identify the optimal model.<br><br>For our internal evaluation, we employed a stratified shuffle split method. This approach involves splitting the data into training and validation sets, repeating the process k times (with k=10 and a validation proportion of 0.3). This validation size was chosen to address the observed discrepancy between internal results and external test performance, which was partly due to the small number of validation samples used initially.<br><br>We explored various models in our comparison, favoring tree-based approaches due to their capacity to handle complex relationships and correlated features. Based on past submissions, the most reliable models were random forest and stochastic gradient boosting. We also evaluated the random forest in its extra trees variant, which uses random splits. For gradient boosting, we used three different implementations: CatBoost, LightGBM, and XGBoost. Although these implementations are based on the same underlying model, their different estimation methods can produce notably different results.<br><br>For these top candidates (forests and boosting), we designed a hyperparameter tuning procedure using a Bayesian approach. After initial runs, we also considered combining the best models, specifically CatBoost, LightGBM, and extra trees, using both stacking and voting methods. For the voting combination, we found that giving a higher weight (0.5) to CatBoost, since it was the best-performing model, provided the best results.<br><br>In conclusion, the final model proposed is a weighted voting combination of CatBoost, LightGBM, and extra trees, with CatBoost given a higher weight (0.5) due to its superior performance. Notably, the best standalone model, CatBoost, performs closely to the combined model, emphasizing its reliability. |

| GAS | 2 | 0.70 | *A stacking combination of CatBoost and random forest; trained on a subsample of handpicked features + identified via a data-driven procedure (see above). Topics include fertility intentions, family and household, relationships, family background, childcare, distribution of child care responsibilities, assets. Handling of missing values is explained above.* |
|---|---|---|---|
| GAS | 1 | 0.69 | *Catboost trained on a small subsample of handpicked features about family and household, including fertility intentions. A distinct category label is created for missing values.* |
| Scratch cooking | 2 | 0 | *The prediction function assigns the same class (0) to all observations.* |
| Scratch cooking | 1 | 0.68 | *RandomForest trained on 12 features about fertility intentions, employment, and partnership status. Missing values are mean-imputed.* |
| rishabind | 4 | 0.65 | *HistGradientBoostingClassifier trained on features on demographics, partnership status, socioeconomic background, and fertility intentions, supplemented by two engineered variables capturing wave-to-wave variability in fertility intentions. To address class imbalance, the minority class (having a child) is upsampled to 60% of the majority class size before training. Missing values are handled natively by the gradient boosting algorithm* |
| Bernoullistorksquad | 3 | 0.64 | *XGBoost trained on features on demographics, fertility intentions, partnerships, assets, income, employment, health, religion, and attitudes. Several groups of variables are averaged into scales: attitudes toward children, gendered religiosity, traditional fertility/motherhood/fatherhood/marriage attitudes, working mother attitudes, sexism, anxiety, long-standing illness, and religiosity. Missingness is handled with KNN imputation.* |
| Bernoullistorksquad | 2 | 0.41 | *AdaBoost trained mostly on variables from the Background survey (on partnership status, employment, income, demographics, education, housing, and a few variables from the Family and household survey (e.g. fertility intentions); no scales were constructed. All features are treated as categorical and encoded with CatBoostEncoder. Missingness is handled with KNN imputation.* |
| ConceptionCoders | 3 | 0.64 | *Model is a hard-voting ensemble of five classifiers: random forest, HistGradientBoosting, logistic regression, adaboost, and svm. The ensemble weights are determined empirically via training each classifier separately on a 70/30 split of the balanced data and uses their individual F1 scores as weights in the final vote. Class imbalance is addressed via oversampling of the minority class (those who had a child) to match the majority class size. Feature selection: variables from 2007–2020 covering fertility intentions, health, religiosity, number of children, and mortality beliefs, plus a set of 2020 Background survey variables on income, education, civil status, housing, urbanicity. For several variables, means and standard deviations are calculated across waves to capture average levels and individual variability over time. Missing values are imputed with a new category.* |
| danielvwijk | 3 | 0.64 | The goal of my submission is to provide a theoretically-informed model that predicts fertility outcomes well while also being intuitive and easy to interpret. Therefore, the model includes just a few variables. The choice of these variables is guided by theoretical considerations (i.e. what type of considerations drive the decision to have children) and by results of prior work in demography and sociology.<br>As argued by the Theory of Planned Behavior, fertility intentions are likely (one of) the most important predictor of fertility behavior. Therefore, fertility intentions were the starting point of my modelling exercise, and I include a detailed measure of these intentions in the model.<br>In addition, I include age (using a quadratic specification to capture the age pattern in fertility) and partnership status (distinguishing single, LAT, cohabiting, and married respondents) in the model. These are standard demographic characteristics that have been shown over and over again to have strong relationships with fertility.<br>Finally, I add a variable that measures religious attendance to take into account the higher fertility of more religious individuals.<br>In previous modelling attempts, I also included other characteristics that have been linked to fertility, namely parity, gender, educational attainment, employment status, and income. Adding these variables did not improve model fit (indicated by the AIC) and/or results were not as expected. This is most likely because fertility intentions and partnership status already capture much of the variation in fertility outcomes, which informed my decision to keep the model simple by focusing on these strong predictors of fertility. |
| BearlyPredictable | 3 | 0.64 | Summary:<br>Our team used the tidymodels package in R paired with the targets pipeline software to automate our model selection procedure. We tested several different modeling approaches (random forest, BART, XGBoost, MARS, logistic regression, etc.) and a variety of feature combinations. We preprocessed the predictor variables using the "recipes" framework within |
| | | | tidymodels. To tune model hyperparameters, we used 10 fold cross-validation with grid search, repeated 5 times. We then ranked the performance of each model using the average F1 score of the cross-validation folds and evaluated models with the highest scores on our held-out test set. The model we chose to submit is one of our best-performing models, a random forest with hand-selected and carefully prepared variables.<br><br>Variables:<br>One of the benefits of using a tidymodels pipeline for model testing is that we were able to evaluate model performance using a variety of feature combinations. Two of our best-performing sets of variables were 1) a large set of indiscriminate features chosen with minimal deliberation (around 250) and 2) a smaller set of theory-based, hand-selected variables (around 60). The hand-crafted set included questions about prior fertility and future intentions, education, partnership, housing, income, migration, gender, and age and mostly takes values from the most recent survey wave. These features were selected on the basis of their established relevance to fertility as documented in the demography literature.<br>Survey questions about fertility intentions (when planning to have more kids, how many more children wanted, etc.) tended to dominate feature importance plots for both sets of variables. Other influential variables in the hand-crafted set included age, income, domestic situation, number of children, and housing.<br><br>Preprocessing Steps:<br>We prepared variables using tidymodels "recipes", which build a series of preprocessing steps that make training data on many different models easier. For the larger set of variables, we mostly treated all features the same way; we removed all zero variance predictors, imputed missing values using k-nearest neighbor imputation, converted all categorical features to factor variables, and centered and scaled most numeric predictors. In the hand-crafted recipe on the other hand, we were more intentional which each variable and started by taking data from the most recent year only. We then adjusted and cleaned features to ensure they contained as much useful information as possible. These cleaning steps included converting many categorical features to binary dummy variables, creating new variables to represent information from multiple questions more succinctly, and ensuring that all variables will be processed correctly by the model. Finally, we either created a missing category for dummy variables or imputed missing values using k-nearest neighbors imputation.<br>Model:<br>We fit and tuned five types of models--random forest, BART, XGBoost, MARS, logistic regression--using grid search cross-validation to both select the best hyperparameters for each model and identify top performing models. |
| | | | Model performance was evaluated using all metrics, though we ultimately used the F1 score to select our final models. The winning model in terms of F1 score alone was a random forest using the indiscriminate variables. However, though this model performed best on our cross validation folds and test set, we found that it didn't generalize nearly as well to the competition holdout set (F1 score of ~0.93 on our test set vs. F1 score of ~0.61 on the holdout set). We thought using a model with our hand-crafted variables might generalize better. We decided to submit our second best performing model for the final Phase 1 submission instead, a random forest model using the hand-selected features. The test set F1 score for our hand-crafted model was also only around 0.002 lower than the indiscriminate model, inspiring confidence that it will likely perform at least as well as our last submission (and hopefully better!). |
| Bearly Predictable | 1 | 0.13 | *Xgboost model trained on the Background survey variables.* |
| matei4501 | 3 | 0.63 | After testing several approaches (random effects regression, deep learning, machine learning algorithms for time-series data) the best performing one was XGBoost. This could be due to several factors: Firstly, when using XGBoost we can keep all features, even ones with a high percentage of missing data, because of XGBoost's ability to handle missing values by learning branch directions for missing values. Therefore, we do not lose any information. As a side, various imputation methods were tried (interpolation + ffill +bfill, knn, multiple imputation), but that only degraded the results. Secondly, drawbacks of other methods. The regression methods we tried could not capture non-linear relationships. There were not enough observations for a deep learning approach to be able to shine.<br><br>Due to the lack of preprocessing required for an XGBoost model we kept preprocessing simple. We only kept observations where the outcome was known and we only kept columns that were either numeric or categorical.<br><br>There are several advantages to our final result. The model is fast and can take advantage of categorical features, it requires minimal preprocessing, and it is interpretable due to the access to feature importances. |
| matei4501 | 1 | 0.21 | *HistGradientBoosting trained only on age.* |
| RosemberGuerra | 3 | 0.58 | One significant challenge encountered was the extensive amount of missing data within the provided datasets. Instead of employing imputation techniques, which could potentially introduce biases and inflate performance |

metrics, we opted for a more conservative approach by selecting variables with less than 20% missing data. This strategy prevents the model from learning from artificially created data, thereby maintaining the integrity and reliability of our predictive analysis.
We employed the LightGBM algorithm for both variable selection and prediction. LightGBM is renowned for its ability to handle datasets with minor missing values directly within the algorithm, eliminating the need for preliminary data imputation. This feature, along with its robustness and accuracy across various conditions and time frames, makes LightGBM an ideal choice for our objectives.

During our first submission, we observed that the model performed well, indicating promising predictive accuracy. However, we recognized the potential for selection bias, where variables might have been chosen more by chance than for their actual predictive power. In subsequent iterations, we refined our approach to mitigate this bias, resulting in a more robust and reliable model.

Variable selection:

- Adding Noise Variables: To benchmark the importance of each feature, two noise variables were introduced. noise1 follows a normal distribution, and noise2 follows a uniform distribution.
- Model-Based Feature Importance: A LightGBM classifier was utilized across multiple parameter settings to evaluate the importance of each feature relative to the noise variables. This was achieved through a simulation with a predefined number of iterations, ensuring robustness in the variable importance estimation.
- Importance Calculation: For each simulation, the model was trained, and the feature importances were recorded. Features were deemed significant if their importance exceeded that of both noise variables. This importance was then scaled relative to the sum of all importances to normalize across different model runs.
- Frequency and Relevance of Selection: Each feature's selection frequency and scaled importance were calculated across all simulations. Features frequently identified as important were retained for further analysis.

Rationale: the selection of variables through model-based importance ensures that only the most predictive features are retained, reducing model complexity and potential overfitting. By comparing feature importance against noise variables, we can objectively assess whether a feature provides meaningful information or merely adds noise. This method is particularly effective in large datasets where distinguishing signal from noise is crucial. Additionally, using a grid of hyperparameters for LightGBM allows for a

| | | | |
|---|---|---|---|
| | | | comprehensive evaluation under various modeling conditions, further validating the robustness of the selected features.<br>Variables that present a higher importance come from the Family & Household, Economic Situation Assets, and Health surveys in that order [see the table with feature importance in the original repository].<br><br>Model Estimation<br><br>● Data Splitting: The data was split into training and testing sets, with 30% of the data reserved for testing. This split helps validate the model on unseen data, ensuring that our model generalizes well beyond the training data.<br>● Model Definition: LightGBM, a gradient boosting framework that uses tree-based learning algorithms, was chosen for its effectiveness and efficiency with large data sets.<br>● Hyperparameter Tuning:<br>○ A grid search was conducted over a specified parameter grid including num_leaves, max_depth, learning_rate, and n_estimators to find the best model configuration.<br>○ Cross-validation with five folds was used during the grid search to ensure that the model's performance was robust across different subsets of the data.<br>○ The grid search was configured to refit the best model based on the F1 score, balancing the precision and recall of the model.<br>● Model Evaluation: The best model from the grid search was evaluated on the test set. Metrics such as accuracy, precision, recall, and F1 score were computed to assess the model's performance comprehensively.<br>● Rationale: the use of grid search and cross-validation ensures that the model is not only tuned to perform optimally in terms of prediction accuracy but is also stable and reliable across different data splits. This rigorous approach to model tuning and evaluation is crucial for developing a high-performing model that can be confidently used for making predictions. |
| RosembergGuerra | 1 | 0.62 | *Similar to above, but the background survey is not used, there are fewer variables from the core studies, and default hyperparameters are used.* |
| Floratility | 3 | 0.62 | *Logistic regression with four variables: age, fertility intentions, education, and income. Missing values are imputed with the mean.* |
| OxfordDemSci | 3 | 062 | Model Overview<br>We have chosen to utilize a binary GradientBoostingClassifier for our analysis. This method was selected due to its robustness in handling various |
| | | | types of data and its effectiveness in binary classification tasks. Gradient boosting is particularly adept at improving predictions by minimizing errors sequentially using decision trees, making it well-suited for complex datasets with intricate patterns.<br><br>Feature Selection<br>Our model incorporates 75 manually selected features. These features were carefully chosen based on established fertility theories to ensure that each contributes meaningfully to the understanding and prediction of fertility behaviours. The selection process involved rigorous analysis of literature and existing studies, ensuring that the variables used are not only theoretically sound but also empirically validated. There are 38 features that are dummy variables, which were created from 4 categorical variables in the dataset. These dummy variables were included to capture the nuances of the original categorical variables and provide a more detailed representation of the data.<br><br>Data Preprocessing<br>One of the significant challenges we encountered during the data preprocessing stage was the prevalence of missing values within the dataset. Missing data can obscure the true relationships between variables and potentially bias the model's outcomes. To address this, we implemented data imputation techniques aimed at preserving the underlying distribution and relationships of the data as much as possible, thereby allowing for a more accurate and reliable model. |
| Oxford DemSci | 2 | 0.45 | Binary GradientBoostingClassifier with eight manually selected features, including: gender, age, partnership status, domestic situation, household net income, most recent fertility intentions, education level and the number of children in the household. |
| NextFinBaby | 5 | 0.59 | We are a group of fertility experts who wanted to provide a model with predictors driven by theoretical relevance. We included predictors that reflect demographic, economic, social, and psychological factors shown to be important for childbearing in the literature, particularly in the near future. We use a logistic regression model that includes interactions between gender and all other predictors in the model. These interactions were motivated because a) theoretically, many variables tend to have different relationships with fertility for women and men, and b) it improved the predictive accuracy of the model. We manually selected these predictors: gender, age, time since first birth, migration background, education level and field, employment, income, financial satisfaction, home ownership, urban character of place of residence, partner/marriage info and duration, relationship satisfaction, help from own parents, religiousness, personality, subjective health, fertility |

| | | | intentions, distance from parents, cohabitation type with partner, difference of opinion about household work. |
|---|---|---|---|
| NextFinBaby | 3 | 0.61 | *Same approach but without a few variables (distance from parents, currently has a partner, cohabitation type with partner, difference of opinion about household work).* |
| PRC Team | 3 | 0.60 | Random forest with oversampling, based on theoretically-selected features. |
| PRC Team | 2 | 0.61 | Random forest with oversampling, based on theoretically-selected features. |
| christinapao | 2 | 0.61 | Simple decision tree model using scikit-learn. Cleaning process removes all columns in the training data that are missing over half of the entries or non-numeric, and then for the remaining empty entries, replaces with -1, which is not used as a value. |
| UU Human data science | 1 | 0.60 | We kept it simple for this first submission: we selected variables we thought were most relevant, then we performed imputation using a truncated singular value decomposition, and then we performed a default-style random forest using the ranger R package. |
| KinderComputer | 1 | 0.59 | The methods employed here are a mixed approach. We developed custom feature encoding based on domain knowledge, including several non-intuitive variables, and used these variables plus core fertility predictors to train a cross-validated recursively partitioned regression model - also known as a Classification and Regression Tree. |
| BimBO | 3 | 0.59 | The preprocessing steps primarily involve recoding some categorical variables, removing variables with a high proportion of missing values in the training set, and adding some transformed numeric variables.<br>Developing a model: We rely on the Sparse Wrapper AlGorithm (SWAG), employing random forest as base model and considering the implementation proposed by Wright et a.. The SWAG explores the low-dimensional attribute space to identify a set of learners that use a minimal number of attributes yet possess high predictive power. The algorithm follows a forward-step approach, beginning with a few attributes and progressively adding more in each step. At each fixed number of attributes, it tests various randomly selected learners and selects those with the best out-of-sample performance. The algorithm uses the information coming from the best learners at the previous step to build and test learners in the following step. We then construct a set of best performing models based on the estimated out-of-sample performance, estimated by a 10-fold cross-validation and considering the F1 score. |
| | | | Prediction: To predict on the test set with missing data, we use the following strategy. First, we create a copy of the test set where missing data is imputed using the method proposed by Stekhoven et al. For each row in the test set, we identify all applicable models from the set of best-performing models. A model is applicable if it uses only the variables observed in that row. We then predict the outcome for each row using all applicable models and select the most frequent class among these predictions as the final prediction for that row. If no top-performing models can be applied to a row due to missing variables, we use the imputed copy of the row. Predictions are then made using all best-performing models, and the most frequent class among these predictions is taken as the final prediction. |
| tredmill | 3 | 0.59 | We wanted to see how a theory-based approach would fare against 'pure' data-driven techniques.<br>(1) We select a set of indicators based on the literature on fertility transitions and the expert knowledge of team members.<br>(2) Imputation with Mice.<br>(3) Predicted outcomes with a basic neural network (linear probability, 5 hidden layers, rprop+, no cross-validation). |
| Bob & Bobette | 3 | 0.57 | We proceeded in two steps:<br>1. We estimate the distribution of the occurrence of first/second/third child using the full information about respondents and their potential children from 2007 to 2020. For this purpose, we used a recursive neural network (RNN) with gated recurrent units (GRU) and an attention mechanism to fit a time-to-event model for the probability to get a first/second/third child at each age from 18 to 45. This model is trained on the train data alone and does not use information from 2020-2023, which removes the risk of overfitting. This took us a lot of time to implement this model; so far we have done no parameter tuning and we still need to implement some changes, which should increase the interpretability of the model. However, the results with default parameters are encouraging: we obtain a good accuracy for predicting respondents' age at first child, which decreases for the second and third child. From this model, we extract a probability to get a child in the next three years after 2020 for each respondent.<br>2. We use the probability estimated in the neural network with other covariates to predict the outcome in an XGBoost model. The F1 score on the training set is encouraging (F1=0.71, Accuracy=0.87), knowing that so far we have done no parameter tuning and we rely on a first rough selection of variables for both models. The probability estimated from step 1 has the highest explanatory power in the XGBoost model. |

| Bob & Bobette | 1 | 0.37 | *Similar to the above, but a different architecture. The prediction task is treated as a survival analysis problem. Data (demographic, socioeconomic, and health variables) is reshaped into a longitudinal time series per person. An RNN is used to estimate when (at what age) someone is likely to have a child, then extracts whether that falls within the 2021-2023 window. There is no ensemble step as above (Xgboost); the RNN produces the final prediction directly.* |
|---|---|---|---|
| JustThe2OfUs | 3 | 0.58 | Binary logistic regression with eight variables (2020) - age, gender, number of children, fertility intentions, having a partner, educational level, migration background, household net income - selected manually.<br>*Rows with missing values in these variables are removed.* |
| wrjv | 3 | 0.56 | Logistic regression with manually selected variables based on existing literature, removed when there was little to no effect.<br><br>Variables included: age (+ squared & cubic term), cohabitation duration (+ squared & cubic term), having a partner, fertility intentions, being in church on a weekly basis, being unable to make ends meet.<br><br>I have also looked at other variables, including whether a person already had a child, owned a house, was in good health, wasn't obese, was not in a relationship in which the female partner worked, lived in a non-polluted area and did no care work for other family members, but these factors only seemed to increase the complexity of the model.<br><br>I have looked into machine learning techniques, but decided against using those, as I feel the added complexity won't add much here, and I do value the model to be well interpretable as well. |
| JMGN Princeton | 1 | 0.54 | Logistic regression model using these predictors: age, several fertility intentions questions and year of marriage. These variables were decided upon using random forest and selecting the top contributing variables. |
| OvenBakedPeanuts | 3 | 0.50 | We used the logit model to predict whether the respondent had a child or not. In the first round, we added variables from 2020, including gender, age, migration background, education, living arrangements, house ownership, and household income.<br>In the second round, we did not use the variable of living arrangements. We added two other more detailed variables (had a partner in 2020 and the number of children in 2020) to replace the variable of living arrangement.<br>We also added two attitudinal variables measured in 2020: whether the respondents agreed that people who want to have children should get married and whether the respondents thought they would have [more] children in the future. |

| | | | We also tried to add variables, including job status, health conditions, cohabitation, and gender attitudes. But they did not improve the model significantly. |
|---|---|---|---|
| OvenBakedPeanuts | 1 | 0.34 | Logit model; the predictors are variables from 2020 including gender, age, migration background, education, living arrangements, house ownership, and household income. |
| SudaMax | 3 | 0.39 | *The submission uses a full encoder-decoder Transformer. First, tabular data is turned into text (by concatenation). These variables are used: age, gender, number of children, partnership status, housing, employment, education, migration background, urbanicity, income quintile, and household income quintile. This text is then processed through the TransformerEncoder to generate vector representations (embeddings), which are then used for prediction.* |
| SudaMax | 2 | 0.39 | *A similar approach, but fewer variables are used (hence shorter sequences).* |
| Coding babies | 3 | 0.37 | *Logistic regression with five variables: partnership status, fertility intentions, satisfaction with current relationships, and satisfaction with financial situation. Rows with missing values in these variables are removed.* |
| ttliu00 | 3 | 0.29 | *Logistic regression with age, partnership status, religiousness, household income, and urbanicity. Oversampling the minority class to create a balanced training set. Decision threshold 0.25 instead of the default 0.5 to transform probabilities into predicted classes to increase recall for the positive class.* |
| Kawaii-Killer | 1 | 0.19 | Binary logistic regression with 20 imputed binary variables - selected manually |
| SOC-555-CS | 1 | 0 | *A neural network with a wide range of variables and KNN missing values imputation. All columns are normalised; the ID column got normalised too, which led to a zero score.* |
| YueW66 | 1 | 0 | personal_finance_satisfaction as the feature, and a logistic regression algorithm for prediction. |
| Baby Bettors | 2 | 0 | *Logistic regression trained on a broad selection of variables from 9 out of 10 LISS modules across 2017–2020, supplemented by background variables. Missing values are imputed using mean imputation.* |

| aphillips | 1 | 0 | Binary logistic regression using background variables selected manually to give insight into the relative class of an individual (education, living situation, etc.). |
|---|---|---|---|
| data storks | 3 | 0 | *Logistic regression based only on age.* |
| PhDecline | 3 | 0 | *Logistic regression with four variables: age, gender, migration background, and education. Rows with missing values in these variables are removed.* |

## 4. Additional funding

Lucas Sage gratefully acknowledges funding from the French Agence Nationale de la Recherche (under the Investissement d'Avenir programme, ANR-17-EURE-0010).

Julia Hellstrand was supported by the Strategic Research Council (SRC) of the Academy of Finland, FLUX consortium (Family Formation in Flux – Causes, Consequences, and Possible Futures), decision numbers 364374 and 364375.

Kelsey Wright, Julia Hellstrand, and Ziwei Rao were supported by the Strategic Research Council (SRC) in Finland through the FLUX consortium, decision numbers 345130 and 345131

Felix Tropf is funded by the UK Research and Innovation Grant FINDME (EP/Y023080/1).

Jessica Nisén was supported by the Strategic Research Council (FLUX, nr. 364374) and by the Academy of Finland (INVEST, nr. 320162).

Chiara Ludovica Comolli received funding from the Italian Ministry of Universities and Research pursuant to Decree No. 1236 of 01/08/2023 – Italian Science Fund – FIS 2 Call (Funding Admission Decree No. 23178 of 10/12/2024) for the project SO-UNFER (MUR code: FIS-2023-03148; CUP: J53C24004290001).

Javier Garcia-Bernardo acknowledges support from the Dutch Research Council (NWO grant VI.Veni.231S.148)

Team SBU EUI was partially supported by NSF grants IIS1926781, IIS1927227, OAC-1919752, a Fulbright Scholarship, and funding from the French Agence Nationale de la Recherche under the Investissement d'Avenir programme (ANR-17-EURE-0010).

The Princeton team was supported by grants from Princeton Precision Health, the Princeton AI Lab, and the Princeton Catalysis Initiative.